\documentclass[10pt]{article} 
\usepackage[preprint]{tmlr}

\usepackage{amsmath,amsfonts,bm}

\def\eqref#1{equation~\ref{#1}}

\def\1{\bm{1}}

\DeclareMathAlphabet{\mathsfit}{\encodingdefault}{\sfdefault}{m}{sl}
\SetMathAlphabet{\mathsfit}{bold}{\encodingdefault}{\sfdefault}{bx}{n}

\usepackage{hyperref}
\usepackage{url, amssymb}
\usepackage{amsfonts, amsmath, subcaption} 
\usepackage{longfigure}
\usepackage{graphicx, booktabs}
\usepackage{subcaption, float, wrapfig2}
\usepackage{longtable}

\usepackage{array}
\newcolumntype{P}[1]{>{\centering\arraybackslash}p{#1}}
\newcolumntype{M}[1]{>{\centering\arraybackslash}m{#1}}

\newcommand{\tr}{\operatorname{Tr}}

\title{Heavy-Tailed Class-Conditional Priors for Long-Tailed \\ Generative Modeling}

\author{%
	\name Aymene Mohammed Bouayed \email aymene.bouayed@ens.fr\\
	\addr DIÉNS, ÉNS, CNRS, PSL University, Paris, France\\
	Be-Ys Research, France
	\AND
	\name Samuel Deslauriers-Gauthier \email samuel.deslauriers-gauthier@inria.fr\\
	\addr Centre Inria d'Université Côte d'Azur, Nice, France
	\AND
	\name Adrian Iaccovelli \email adrian.iaccovelli@be-ys.com\\
	\addr Be-Ys Research, France
	\AND
	\name David Naccache \email david.naccache@gmail.com \\
	\addr DIÉNS, ÉNS, CNRS, PSL University, Paris, France
}

\def\month{MM}  
\def\year{YYYY} 
\def\openreview{\url{https://openreview.net/forum?id=XXXX}} 

\begin{document}

	\maketitle
	
	\begin{abstract}
		Variational Autoencoders (VAEs) with global priors trained under an imbalanced empirical class distribution can lead to underrepresentation of tail classes in the latent space. While $t^3$VAE improves robustness via heavy-tailed Student's $t$-distribution priors, its single global prior still allocates mass proportionally to class frequency.
		We address this latent geometric bias by introducing C-$t^3$VAE, which assigns a per-class Student's $t$ joint prior over latent and output variables. This design promotes uniform prior mass across class-conditioned components. To optimize our model we derive a closed-form objective from the $\gamma$-power divergence, and we introduce an equal-weight latent mixture for class-balanced generation. On \mbox{SVHN-LT}, CIFAR100-LT, and CelebA datasets, C-$t^3$VAE consistently attains lower FID scores than $t^3$VAE and Gaussian-based VAE baselines under severe class imbalance while remaining competitive in balanced or mildly imbalanced settings. In per-class F1 evaluations, our model outperforms the conditional Gaussian VAE across highly imbalanced settings. Moreover, we identify the mild imbalance threshold $\rho < 5$, for which Gaussian-based models remain competitive. However, for $\rho \geq 5$ our approach yields improved class-balanced generation and mode coverage.
	\end{abstract}
	
	\section{Introduction}
	Class imbalance and long-tail distributions are common in real-world datasets, yet generative models often fail to represent rare classes accurately. Under skewed training data, models tend to overfit dominant modes and underrepresent minority ones in latent and output spaces, resulting in biased generations. This issue is particularly consequential in applications such as facial synthesis \citep{ldm-face-2} and medical imaging \citep{ldm-brain}, where such biases can exacerbate social and diagnostic disparities \citep{bias}. Addressing class imbalance to ensure balanced representational capacity across categories remains a major challenge for all generative models.
	
	For Generative Adversarial Networks (GANs)~\citep{gan}, their inherently unstable training dynamics, exacerbated by data imbalance, often lead to biased and mode-collapsed generations. Several works have sought to mitigate these effects. The Wasserstein GAN~\citep{wgan} replaces the Jensen--Shannon divergence with the Wasserstein distance to stabilize optimization and improve sample diversity. PacGAN~\citep{pacgan} enhances robustness by packing multiple samples into the discriminator, reducing mode collapse and improving diversity under implicit imbalances. Similarly,~\citep{teaching_gan_what_not_to_learn} introduces negative data augmentation to prevent under-representation of minority classes, enabling more class-balanced supervised GAN training. More recently, RareGAN~\citep{raregan} addresses unlabeled long-tailed data through a weighted loss and adaptive labeling budgets, improving both balance and generative diversity. Despite these advances, GANs remain difficult to train reliably, especially under strong imbalance.
	
	Diffusion models have emerged as a more stable alternative to GANs, achieving superior image quality and convergence behavior~\citep{diffusion_beats_gans}. Several diffusion-based methods explicitly tackle data imbalance. Class-Balancing Diffusion Models~\citep{class-balancing-diffusion-model} modify the denoising process to be class-invariant, while~\citep{longtailed_diffusion_calibration} employs weighted score matching with Bayesian calibration to transfer knowledge from majority to minority classes, improving diversity and fidelity on long-tailed datasets. Heavy-Tailed Diffusion Models~\citep{heavytailed} take a different approach by replacing the Gaussian noise assumption with a Student’s $t$ formulation, offering a more robust fit for imbalanced distributions. Nevertheless, effectively addressing class imbalance within diffusion frameworks remains an open and active research area.
	
	Variational Autoencoders~(VAEs)~\citep{vae} are valuable models when interpretable, structured latent spaces and efficient single-step sampling are needed. VAEs provide a principled probabilistic foundation, stable training dynamics, and compatibility with latent-variable pipelines such as latent diffusion \citep{ldm}. However, in class imbalanced setting, standard VAEs using isotropic Gaussian priors struggle to capture heavy-tailed or rare structures, creating class-coverage bottlenecks~\citep{dist_proof}. Therefore, prior work explores non-Gaussian priors, especially Student’s $t$-distributions~\citep{student-t-1, student-t-2, gamma-divergence, t3vae}, to improve robustness to outliers and class imbalance. 
	
	In this work, we view class imbalance not only as a sampling issue but as a geometric one: optimizing under the empirical distribution tends to allocate latent probability mass in proportion to class frequency. Consequently, majority classes occupy larger effective regions of latent space, while minority classes are confined to smaller regions with reduced representational capacity. This interpretation is particularly transparent in latent-variable models with an explicit global prior, such as VAEs, where representational allocation is directly governed by prior density. While imbalance also affects GANs and diffusion models, in those frameworks it is intertwined with adversarial training dynamics or iterative score-based denoising. In contrast, VAEs allow us to study how prior specification alone shapes latent geometry. We therefore focus on VAEs to isolate the role of prior-induced frequency bias in a controlled and analytically tractable setting. 
	
	Current VAE based approaches retain a single global prior \citep{vae, student-t-1, student-t-3, student-t-2}. Hence through our geometric view the latent remains frequency-aligned where class regions scale with empirical probability and heavy tails alone do not eliminate this frequency-induced geometric bias. Indeed, optimizing a VAE under $p_{data}(x) = \sum_y p(y)p(x|y)$ tends to allocate latent regions whose effective volume correlates with class frequency $p(y)$.	We address this issue by introducing the Conditional-$t^3$VAE \mbox{(C-$t^3$VAE)}, which imposes a per-class Student's $t$-distribution prior over the joint latent-output space. The design promotes uniform prior mass across class-conditioned components under the sampling distribution, mitigating majority-class dominance while the heavy tails capture intra-class variation. For class-balanced generation, we derive an equal-weight latent mixture of Student's $t$-distributions with analytically derived component variances. Importantly, C-$t^3$VAE class-specific heavy-tailed priors induce a structured latent mixture whose geometry differs fundamentally from both global heavy-tailed priors and Gaussian conditional VAEs. We summarize our main contributions in the following points~:
	
	
	\begin{itemize}
		\item We propose the C-$t^3$VAE model with a training objective based on the $\gamma$-power divergence.
		\item We develop an equal-weight latent mixture sampling scheme with analytically derived optimal variance scaling for each component.
		\item We outperform relevant baselines in FID on SVHN-LT \citep{svhn}, \mbox{CIFAR100-LT} \citep{cifar100lt}, and CelebA \citep{celeba} under severe imbalance, and show via per-class evaluation that \mbox{C-$t^3$VAE} better avoids mode collapse, exceeding a Gaussian conditional VAE in per class Recall and F1 while remaining competitive on Precision.
		\item We empirically observe a transition regime around $\rho \approx 5$, beyond which Gaussian priors become suboptimal, providing guidance for model selection on skewed datasets.
	\end{itemize}
	
	\section{Related works}
	Since the introduction of Variational Autoencoders (VAEs)~\citep{vae}, numerous extensions have aimed to enhance latent representation expressiveness by replacing the standard Gaussian prior with more flexible formulations. Notable examples include Gaussian mixtures~\citep{gmvae, shape-your-space}, hyperspherical priors~\citep{s-vae}, normalizing flows~\citep{noramalizing-flow-imbalance}, Riemannian manifolds~\citep{gemetric-vae}, and implicit distributions~\citep{implicit-prior}. Most of these retain the Evidence Lower Bound (ELBO) optimization framework, while others introduce alternative divergence measures for greater modeling flexibility~\citep{aae}. Hierarchical VAEs~\citep{nvae} and vector-quantized VAEs~\citep{vqvae} have also been proposed to improve disentanglement and mitigate posterior collapse.
	
	Student's $t$-distributions have been widely studied for their heavy-tailed robustness to outliers~\citep{dist_proof}. Early works~\citep{DE-VAE, student-t-2} incorporated $t$-distributed priors into the VAE framework using KL-divergence-based ELBO objectives to promote robust latent encodings. However, because the KL divergence between $t$-distributions lacks a closed-form solution, such approaches depend on numerical approximations, increasing computational cost. \citep{student-t-1} proposes a Student's $t$ decoder to improve generative performance, but the heavy-tailed assumption is confined to the decoder, leaving the latent space Gaussian and limiting validation to tabular data. The $t^3$VAE~\citep{t3vae} further advances this line of work by jointly modeling latent and output distributions under Student's $t$-assumptions for the encoder, decoder, and prior, replacing the KL divergence with a closed-form $\gamma$-divergence~\citep{gamma-divergence}. 
	
	While several works improve prior expressiveness or robustness, they do not explicitly analyze how global prior formulations induce frequency-aligned allocation of latent probability mass under imbalanced data. In particular, existing approaches focus on better density modeling or optimization stability, rather than controlling how representational capacity is distributed across classes. As a result, even expressive or heavy-tailed priors may continue to reflect empirical class proportions in latent space. 
	
	A complementary line of research addresses imbalance through conditional modeling. Conditional VAEs (CVAEs)~\citep{cvae, cvae-2} condition latent encodings on class labels, enabling targeted generation of minority categories. However, their Gaussian priors poorly approximate heavy-tailed or rare data structures. Our C-$t^3$VAE bridges two threads: it inherits the heavy-tailed robustness of $t^3$VAE while adopting the class-conditional structure of CVAEs, and further introduces a theoretically grounded equal-weight mixture sampling scheme that explicitly counteracts majority-class dominance at generation time. Hence, our contribution is not merely heavy tails nor conditioning, but decoupling representational capacity from empirical frequency through prior mass allocation.

	\section{Background}
	This section introduces the theoretical background and baseline models relevant to our work. We assume access to a labeled, imbalanced dataset \mbox{$\mathcal{D}= \{(x_i, y_i)\}_{i=1}^N$}, where $x_i \in \mathbb{R}^n$ is a data sample of dimension~$n$, $y_i \in \{1, \dots, K\}$ its class label with $K$ being the number of classes and $m$ being the latent space dimension.
	
	\subsection{VAEs and Conditional VAEs}
	VAEs \citep{vae} are generative models trained via variational inference by maximizing the Evidence Lower Bound (ELBO) of the log-likelihood. The standard objective of this model is
	\begin{equation}
		\mathcal{L}_{\theta, \phi}:= \mathbb{E}_{z\sim q_\phi(\cdot \vert x)}[\log p_\theta(x\vert z)] - \mathcal{D}_{KL}(q_\phi(z\vert x) \Vert p(z)),
		\label{ELBO:VAE}
	\end{equation}
	where the first term is the reconstruction loss with $p_\theta(x\vert z)$ being the decoder model. The second term is the Kullback–Leibler (KL) divergence between the approximate posterior $q_\phi(z\vert x)$ and the prior $p(z)$. The $\beta$-VAE is a weighted variant of the VAE model which introduces a $\beta$ scaling term for the KL divergence \citep{beta-vae}:
	\begin{equation}
		\mathcal{L}_{\theta, \phi}:= \mathbb{E}_{z\sim q_\phi(\cdot \vert x)}[\log p_\theta(x\vert z)] - \beta \mathcal{D}_{KL}(q_\phi(z\vert x) \Vert p(z)),
		\label{ELBO:beta-VAE}
	\end{equation}
	with $p(z) \sim \mathcal{N}_m(0,I)$, $q_\phi(\cdot \vert x) \sim \mathcal{N}_m(\mu_\phi(x), \Sigma_\phi(x))$, and $p_\theta(x\vert z) \sim \mathcal{N}_m(\mu_\theta(z), \sigma^2I)$. $\mu_\phi(\cdot)$ and $\Sigma_\phi(\cdot)$ are the mean and covariance matrices of the approximate posterior. They are inferred by a neural network with parameters~$\phi$ given the input $x$. Moreover, $\mu_\theta(\cdot)$ is the decoder neural network with parameter~$\theta$ and $\sigma$ is a parameter controlling the decoder's output covariance. This variant of the VAE model allows to place more weight on disentangling the latent space or on the reconstruction of the data points. To generate samples from the VAE or the $\beta$-VAE model, we sample a latent vector $z \sim  \mathcal{N}_m(0,I)$. Then, the generated data point would be $\hat{x} \sim  \mathcal{N}_m(\mu_\theta(z), \sigma^2I)$.
	
	Nevertheless, since Eq.~(\ref{ELBO:VAE}) and Eq.~(\ref{ELBO:beta-VAE}) optimize the ELBO over the data distribution \(p_{data}(x)\), which can be decomposed as $p_{\mathrm{data}}(x) = \sum_{y_i} p(y_i) \, p_{\mathrm{data}}(x \mid y_i)$. In the context of imbalanced data, this optimization inherently biases the model toward head classes with larger \(p(y_i)\). As a result, most generated samples come from overrepresented classes, while tail classes' samples are underrepresented and of lower quality, a phenomenon commonly referred to as \textsl{mode collapse}.
	Therefore, when labels are available, it is preferable to define class-conditional posterior and prior distributions: $q_\phi(z \vert x, y)$ and $p(z \vert y)$. This yields the Conditional-VAE (CVAE) model trained using the objective \citep{cvae}:
	\begin{equation}
		\sum_y \ \mathbb{E}_{z\sim q_\phi(\cdot \vert x, y)}[\log p_\theta(x\vert z, y)] - \beta \mathcal{D}_{KL}(q_\phi(z\vert x,y) \Vert p(z\vert y)).
		\label{ELBO:CVAE}
	\end{equation}
	In practice, in Eq. (\ref{ELBO:CVAE}) we optimize the class-conditional objective with uniform weighting across labels, removing explicit frequency-dependent scaling from the loss. Also, we define $p(z \vert y) \sim \mathcal{N}_m(\mu_y,I)$ with learnable class-wise means $\mu_y$. To generate a data point $\hat{x}_y$ from class $y$, we sample $z_y \sim \mathcal{N}_m(\mu_y,I)$, then we get $\hat{x}_y \sim p_\theta(x\vert z_y, y)$.
	Nevertheless, despite conditioning, this formulation remains Gaussian. Unlike Student's $t$-distributions, Gaussian priors poorly approximate heavy-tailed data distributions \citep{dist_proof}.
	
	\subsection{Multivariate Student's $t$-Distribution}
	A $d$-dimensional Student's $t$-distribution with mean $\mu \in \mathbb{R}^d$, covariance $\Sigma \in \mathbb{R}^{d \times d}$, and degrees of freedom~\mbox{$\nu > 2$} is a heavy-tail, super-Gaussian distribution defined as 
	\begin{equation}
		t_d(x) = C_{\nu, d} \vert\Sigma\vert^{-\frac{1}{2}} \left(1+ \frac{(x-\mu)^\top \Sigma^{-1}(x-\mu)}{\nu}\right)^{-\frac{\nu+d}{2}}, \qquad
		C_{\nu, d} = \frac{\Gamma\left(\frac{\nu+d}{2}\right)}{\Gamma\left(\frac{\nu}{2}\right) \left(\nu\pi\right)^{\frac{d}{2}}}.
		\label{def student t}
	\end{equation}
	The power form of this distribution prevents a closed-form KL divergence between two Student's \mbox{$t$-distributions}. Instead, the $\gamma$-power divergence $\mathcal{D}_\gamma(q \Vert p)$ is used \citep{gamma-divergence, t3vae}. This divergence is defined for $q \sim t_d(\mu_0;\Sigma_0; \nu)$, \mbox{$p \sim t_d(\mu_1;\Sigma_1; \nu)$} as 
	\begin{equation}
		\mathcal{D}_\gamma(q \Vert p) := \frac{\mathcal{C}_\gamma(q,p) - \mathcal{H}_\gamma(p)}{\gamma}
		\label{def gamma divergence}
	\end{equation}
	with $\gamma = -\frac{2}{\nu + d}$, the $\gamma$-entropy $\mathcal{H}_\gamma(p)$ and $\gamma$-cross-entropy $\mathcal{C}_\gamma(q,p)$ being 
	\begin{equation*}
		\mathcal{H}_\gamma(p) := -\Vert p \Vert_{1+\gamma} = -\left(\int p(x)^{1+\gamma} dx \right)^{\frac{1}{1+\gamma}}, \qquad
		\mathcal{C}_\gamma(q,p) := -\int q(x) \left(\frac{p(x)}{\Vert p \Vert_{1+\gamma}}\right)^{\gamma} dx.
	\end{equation*}
	Then, substituting the definition of a Student's $t$-distribution from Eq.~(\ref{def student t}) into Eq.~(\ref{def gamma divergence}), the following closed-form formula for the $\gamma$-power divergence can be derived (Derivation of Proposition 3 in \citep{t3vae})
	\begin{align}
		\mathcal{D}_\gamma(q \Vert p) &= - \frac{C_{\nu, d}^{\frac{\gamma}{1+\gamma}}}{\gamma} \left(1+\frac{d}{\nu-2}\right)^{-\frac{\gamma}{1+\gamma}} \Bigg[-\vert \Sigma_0 \vert^{-\frac{\gamma}{2(1+\gamma)}}  
		\quad  \left(1+\frac{d}{\nu-2}\right)   \nonumber \\
		&\quad + \vert \Sigma_1 \vert^{-\frac{\gamma}{2(1+\gamma)}} \Bigg( 1 + \frac{\tr\left(\Sigma_1^{-1}\Sigma_0\right)}{\nu -2} + \frac{(\mu_0 - \mu_1)^\top \Sigma_1^{-1}(\mu_0 - \mu_1)}{\nu} \Bigg)
		\Bigg].
		\label{gamma divergence}
	\end{align}
	
	\subsection{$t^3$-Variational Autoencoder}
	\subsubsection{Definition}
	The $t^3$VAE model \citep{t3vae} is a non-ELBO-based autoencoder which models the joint prior distribution $p_\theta(x,z)$ using multivariate Student's $t$-distribution
	\begin{equation*}
		p_\theta(x,z) = \sigma^{-n} C_{\nu, m+n} \left[ 1 + \frac{1}{\nu} \left(\Vert z \Vert^2 + \frac{\Vert x - \mu_\theta(z) \Vert^2}{\sigma^2} \right)\right]^{-\frac{\nu+m+n}{2}}.
	\end{equation*}
	From this joint distribution, the marginal latent prior $p(z)$ and decoder distribution $p_{\theta}(x \vert z)$ can be defined as follows
	\begin{equation*}
		p(z) = t_m(z \vert 0, I, \nu), \quad
		p_\theta(x\vert z) = t_n \left( x \middle\vert \mu_\theta(z), \frac{1 + \nu^{-1} \Vert z \Vert^2}{1+\nu^{-1}m} \sigma^2I, \nu+m \right).
	\end{equation*}
	Furthermore, the posterior distribution is defined as :
	\begin{equation*}
		q_\phi(z\vert x) = t_m \left( x \middle\vert \mu_\phi(x), \frac{\Sigma_\phi(x)}{1+\nu^{-1}n}, \nu+n \right).
	\end{equation*}
	Hence, the data distribution would be $q_\phi(x,z)=p_{\text{data}}(x)q_\phi(z\vert x)$. As a result, relying on the $\gamma$-divergence in Eq. (\ref{gamma divergence}) applied to the $p_\theta(x,z)$ and $q_\phi(x,z)$ distributions, the following loss function is derived to optimize the $t^3$VAE's parameters :
	\begin{equation}
		\mathcal{L}_\gamma = \mathbb{E}_{x}\Bigg[ \frac{\mathbb{E}_{z}\left[\Vert x-\mu_\theta(z) \Vert^2 \right]}{\sigma^2} + \Vert \mu_\phi(x) \Vert^2 + \frac{\nu \tr\left(\Sigma_\phi(x) \right)}{\nu+n-2} 
		- \frac{\nu C_1}{C_2} \vert \Sigma_\phi(x) \vert^{-\frac{\gamma}{2(1+\gamma)}} 
		\Bigg],
		\label{t3VAE loss}
	\end{equation}
	with $\gamma = -\frac{2}{\nu+n+m}$ and $C_1$ and $C_2$ being constants theoretically derived in \citep{t3vae}. We note that the first term in this loss function represents the standard reconstruction term in VAE models and the rest of the terms are regularization terms over the latent space. To sample from the latent space of the $t^3$VAE, \citep{t3vae} propose the $p_\nu^\star(z) = t_m(0, \tau^2 I, \nu+n)$ distribution with
	\begin{equation}
		\tau^2 = \left(1+\nu^{-1}n\right)^{-1} \left( \sigma^n C^{-1}_{\nu, n} \quad \frac{\nu+n-2}{\nu-2}\right)^{-\frac{2}{\nu+n-2}}.
		\label{kim tau}
	\end{equation}
	
	Moreover, sampling from a multi-variate Student's $t$-distribution \mbox{$T \sim t_d(\mu, \Sigma, \nu)$} both in the learning (Eq.~(\ref{t3VAE loss})) and sampling (Eq. (\ref{kim tau})) phases is performed through the standard reparameteration trick for Student's \mbox{$t$-distributions} $T := \mu + Z \sqrt{\nu V^{-1}}$ where $Z\sim \mathcal{N}(0, \Sigma)$ and $V \sim \mathcal{X}^2(\nu)$.
	
	\subsubsection{$\beta$-$t^3$VAE extension}
	From Eq. (\ref{t3VAE loss}) we can also define a $\beta$-$t^3$VAE model by multiplying all the regularization terms by a $\beta$ factor. Similarly to $\beta$-VAE models, this improves the versatility of the model and allows either a focus on generation or disentangling.
	
	In summary, although the $t^3$VAE effectively models heavy-tailed distributions through Student's \mbox{$t$-distributions} and $\gamma$-power divergence, it does not explicitly address class imbalance in the latent space as it does not allocate equal prior mass across class-conditioned components. In the next section, we introduce a class-conditional variant of the $t^3$VAE, designed to enable class-balanced generation across all classes.
	
	\section{Conditional $t^3$-Variational Autoencoder}
	In this section, we propose the Conditional $t^3$-Variational Autoencoder~\mbox{(C-$t^3$VAE)}, present its formulation, training objective, and sampling strategy. C-$t^3$VAE models the latent space as a mixture of Student’s $t$-distributions, one per class, inducing uniform prior mass across class-conditioned components under the sampling distribution. Intra-class variability is captured through the heavy-tailed nature of the Student’s $t$ prior.
	
	\subsection{Model definition}
	The C-$t^3$VAE we propose is based on the following class conditional joint prior distribution
	\begin{equation*}
		p_\theta(x,z \vert y) = \frac{C_{\nu, m+n}}{\vert \Sigma_x \vert^{\frac{1}{2}} \vert \Sigma_y \vert^{\frac{1}{2}} }  
		\Bigg[ 1 + \frac{(z-\mu_y)^\top \Sigma_y^{-1}(z-\mu_y)	+ (x-\mu_\theta(z))^\top \Sigma_x^{-1}(x-\mu_\theta(z))}{\nu} 
		\Bigg]^{-\frac{\nu+m+n}{2}},
	\end{equation*}
	with $\nu$, $n$ and $m$ being the degrees of freedom of the Student's $t$-distribution, the dimension of the input data and the dimension of the latent space respectively. $\mu_y \in \mathbb{R}^m$ is a learnable mean vector representing class centers in latent space of dimension $m$. Moreover, $\Sigma_x$ and $\Sigma_y$ are the covariance matrices of the prior distributions over the latent and output variables.
	
	From this joint distribution, we can derive the conditional latent prior $p(z\vert y) $ and decoder distribution $p_\theta(x\vert z,y)$ (See Appendix \ref{prior derivation}) 
	\begin{equation*}
		p(z\vert y) = t_m(z \vert \mu_y, \Sigma_y, \nu), \quad 
		p_\theta(x\vert z,y) = t_n \Bigg( x \Bigg\vert \mu_\theta(z), \frac{ \left(1 + \nu^{-1} (z-\mu_y)^\top \Sigma_y^{-1}(z-\mu_y)\right)}{(1+\nu^{-1}m)} 
		\Sigma_x, \nu+m \Bigg).
	\end{equation*}
	By defining class-specific priors, the model ensures that each class occupies a proportionally equal latent region, countering the frequency-aligned bias of a global prior. Furthermore, as in $t^3$VAE, we define the posterior $q_\phi(z \vert x)$ as a multivariate Student's $t$-distribution capturing heavy-tailed structure in the latent space :
	\begin{equation*}
		q_\phi(z\vert x) = t_m \left( z \middle\vert \mu_\phi(x), \frac{\Sigma_\phi(x)}{1+\nu^{-1}n}, \nu+n \right).
	\end{equation*}
	Although the posterior is defined as $q_\phi(z|x)$ without explicit class conditioning, the class-specific prior and objective derived in the following section enforce class-discriminative latent encodings.
	
	\subsection{Objective function}
	Harnessing Eq. (\ref{def gamma divergence}) and the defined prior and posterior distributions of the proposed C-$t^3$VAE, we derive in Appendix \ref{loss derivation} the following class-wise objective
	\begin{align*}
		\mathcal{L}(\gamma,y) = \mathbb{E}_{x} &\Bigg[ \mathbb{E}_{z}\Big[(x-\mu_\theta(z))^\top \Sigma_x^{-1}(x-\mu_\theta(z)) \Big] 
		+ (\mu_\phi(x)-\mu_y)^\top\Sigma_y^{-1}(\mu_\phi(x)-\mu_y) \notag\\
		&\quad + \frac{\nu \tr\left( \Sigma_y^{-1} \Sigma_\phi(x) \right)}{\nu+n-2} -\frac{\nu C_1}{C_2} \vert \Sigma_\phi(x) \vert^{-\frac{\gamma}{2(1+\gamma)}}
		\Bigg],
	\end{align*}
	with $C_1 = \left(C^{\gamma}_{\nu+n, m} \left( 1 + \frac{n}{\nu} \right)^{\frac{\gamma m}{2}}  \frac{\nu+n+m-2}{\nu+n-2} \right)^{\frac{1}{1+\gamma}}$ and $C_2 = \left( \frac{C_{\nu, m+n}^{\gamma}}{\vert \Sigma_x \vert^{\frac{\gamma}{2}} \vert \Sigma_y \vert^{\frac{2\gamma +1}{2}}} \left(1 + \frac{m+n}{\nu-2}\right)^{-\gamma} \right)^{\frac{1}{1+\gamma}}$.
	
	\noindent By taking $\Sigma_x = \sigma^2I$ and $\Sigma_y = I$, $\mathcal{L}(\gamma,y)$ objective function simplifies to : 
	\begin{equation}
		\mathcal{L}(\gamma,y) = \mathbb{E}_{x}\Bigg[ \frac{\mathbb{E}_{z}\left[\Vert x-\mu_\theta(z) \Vert^2 \right]}{\sigma^2} + \Vert \mu_\phi(x)-\mu_y \Vert^2 
		+ \frac{\nu \tr\left(\Sigma_\phi(x) \right)}{\nu+n-2}  -\frac{\nu C_1}{C_2} \vert \Sigma_\phi(x) \vert^{-\frac{\gamma}{2(1+\gamma)}} 
		\Bigg].
		\label{objective function}
	\end{equation}
	From Eq. (\ref{objective function}), we observe that the first term aims to minimize the input reconstruction error, ensuring data fidelity, while the second term promotes alignment between the input class mean and the inferred posterior mean. The third and fourth terms jointly regularize the posterior covariance $\Sigma_\phi(x)$; specifically, the trace penalty restricts the total variance to encourage compactness, whereas the determinant term acts as a barrier function to prevent the covariance matrix from collapsing to a singularity.
	
	Finally, we express the final loss function $\mathcal{L}(\gamma)$ over the whole dataset as :
	$$
	\mathcal{L}(\gamma) = \sum_y \mathcal{L}(\gamma,y).
	$$
	As in Eq. (\ref{ELBO:CVAE}), here too we use uniform class weighting so the loss does not scale with empirical frequency and the generative step targets all classes uniformly.
	
	\subsection{Sampling distribution}
	To sample from the latent space of the C-$t^3$VAE, we define the following sampling distribution :
	\begin{equation}
		p_\nu^\star(z) = \sum_{y=1}^K \alpha_y \cdot p_{\nu,y}^\star(z) = \sum_{y=1}^K \alpha_y \cdot t_m(\mu_y, \tau^2 I, \nu+n), \quad \forall y, \;\; \alpha_y= \frac{1}{K}.
		\label{sampling distribution}
	\end{equation}
	For the the variance parameter $\tau^2$ in $t_m(\mu_y, \tau^2 I, \nu+n)$, we derive it theoretically. In this derivation we first derive the $\gamma$-power divergence between $p_{\nu,y}^\star(z) = t_m(\mu_y, \tau^2 I, \nu+n)$ and $q_\phi(z\vert x)$ for every $y$. Then, we compare the obtained result to the corresponding regularization terms in $\mathcal{L}(\gamma,y)$ Eq.~(\ref{objective function}) (See Appendix \ref{sampling derivation} for details). The obtained form is similar to the form expressed in Eq.~(\ref{kim tau}).%
	
	The mixture-based sampling distribution we define in Eq.~(\ref{sampling distribution}) with equal $\alpha_y$ is not a post-hoc balancing heuristic. Rather, it is the generative counterpart of promoting uniform prior mass across class-conditioned latent components. Sampling proportionally to empirical frequency would partially reintroduce frequency-aligned bias at generation time, attenuating the structural change imposed during training. Our framework also allows prioritization by modifying the mixture weights $\alpha_y$ when targeted generation is desired; however, in this work we focus on the balanced setting.
	
	\subsubsection{$\beta$-C-$t^3$VAE extension}
	As with $t^3$VAE, the class-wise objective defined in Eq. (\ref{objective function}) can be split into a reconstruction and regularization terms. By preceding the regularization term with a $\beta$ scalar, we can define a $\beta$-C-$t^3$VAE model thereby improving the domain of applicability of the model.
	
	\section{Experiments}
	This section presents quantitative and qualitative results for the C-$t^3$VAE model and closely related baselines (VAE, C-VAE, and $t^3$VAE, with their $\beta$ variants). This controlled comparison isolates the contributions of our design choices:
	\begin{itemize}
		\item \textbf{VAE :} ELBO trained standard Gaussian-based VAEs.
		\item \textbf{C-VAE :} VAE supplemented by conditional Gaussian priors to assess the class conditioning effect without changing the prior family.
		\item \textbf{$t^3$VAE :} Student's $t$-distribution latent prior and $\gamma$-power divergence objective; it does not use class-conditional priors and does not allow class-conditional generation. Comparing to this model isolates the effect of conditional modeling with heavy-tailed priors.
	\end{itemize} 
	We also evaluate $\beta$-VAE, $\beta$-C-VAE, and $\beta$-$t^3$VAE variants to assess the effect of tuning regularization on latent disentanglement and generative performance.
	
	Besides, we note that we do not aim to compete with state-of-the-art diffusion or adversarial models in raw sample quality. Instead, we isolate the effect of latent prior geometry under class imbalance within a controlled variational framework. Comparing across fundamentally different architectural families would confound inductive bias, training stability, and objectives with the geometric contribution we study. Hence, our evaluation is restricted to VAE-family models and we do not benchmark against diffusion-based generators.
	
	In the following, we first present the evaluation metrics and protocol, then analyze the latent sampling standard deviation~$\tau$ used in $t^3$VAE and \mbox{C-$t^3$VAE} by varying $\tau$ to assess alignment between empirical and theoretical values. We then report FID comparisons across baselines with optimized hyper-parameters (the tuning of $\beta$, $\nu$, and $\tau$ is reported in Appendix~\ref{hyper param tunning}), followed by per-class generative evaluation.

	\subsection{Evaluation metrics}
	\label{pre:evaluation_metrics}
	In this section, we present the evaluation metrics used throughout this work. We report FID and generative Precision, Recall and F1 because they are the most commonly used metrics in the VAE and long-tailed generation literature, enabling direct comparison with prior work. Also, these metrics allow us to assess whether the class-conditional heavy-tailed prior in C-$t^3$VAE improves minority-class generation relative to baselines.
	\subsubsection{Fréchet Inception Distance}
	\label{fid_description}
	The Fréchet Inception Distance (FID) \citep{fid} is a standard metric for evaluating the quality of synthetic images. It measures the similarity between the distributions of real and generated images, with a score of 0 indicating identical distributions.
	
	To compute the FID, feature encodings of real and generated images are extracted using the InceptionV3 network \citep{inception_v3}, excluding the classification layer. Assuming these encodings follow a multivariate Gaussian distribution, the mean vectors $\mu_r$ and $\mu_s$ and covariance matrices $\Sigma_r$ and $\Sigma_s$ are estimated for the real and synthetic image sets, respectively. The FID is then calculated as:
	\begin{equation*}
		FID = \Vert \mu_r - \mu_s \Vert_2^{2} + \tr \left( \Sigma_r + \Sigma_s - 2 \cdot \left( \Sigma_r \Sigma_s \right)^{\frac{1}{2}} \right).
	\end{equation*}
	
	\subsubsection{Generative Precision, Recall, and F1}
	\label{precision_recall_F1_description}
	While FID provides a holistic measure of distributional similarity, it does not explicitly assess the quality and diversity of generated samples. To address this, generative Precision and Recall metrics have been proposed \citep{precision_recall}. These metrics evaluate respectively the sharpness and mode coverage of generated samples relative to the target distribution. Their harmonic mean, the F1 score, offers a balanced assessment.
	
	Given sets of real and generated samples $X_r$ and $X_g$, feature representations are extracted using a classifier network, yielding vectors $\delta_r$ and $\delta_g$. The complete sets of feature vectors are denoted $\Delta_r$ and $\Delta_g$, with $\Delta \in \{\Delta_r, \Delta_g\}$. A binary function is defined as:
	\begin{equation*}
		f(\delta, \Delta) = 
		\begin{cases}
			1 & \text{if } \Vert \delta - \delta^\prime \Vert_2 \leq \Vert \delta^\prime - \text{NN}_k(\delta^\prime, \Delta) \Vert_2 \text{ for at least one } \delta^\prime \in \Delta, \\
			0 & \text{otherwise,}
		\end{cases}
	\end{equation*}
	where $\text{NN}_k(\delta^\prime, \Delta)$ denotes the $k$-th nearest neighbor of $\delta^\prime$ in $\Delta$ (we use $k=3$ in all experiments). The function $f(\delta_g, \Delta_r)$ determines whether a generated sample appears realistic, while $f(\delta_r, \Delta_g)$ assesses whether a real sample could be reproduced by the generator. Precision and Recall are defined as:
	\begin{align*}
		\text{Precision}(\Delta_r,\Delta_g) &= \frac{1}{\vert \Delta_g \vert} \sum_{\delta_g \in \Delta_g} f(\delta_g, \Delta_r), \\
		\text{Recall}(\Delta_r,\Delta_g) &= \frac{1}{\vert \Delta_r \vert} \sum_{\delta_r \in \Delta_r} f(\delta_r, \Delta_g).
	\end{align*}
	Finally, the F1 score is computed as:
	\begin{equation*}
		F1 = 2 \cdot \frac{\text{Precision} \cdot \text{Recall}}{\text{Precision} + \text{Recall}}.
	\end{equation*}
	We note that Precision, Recall, and F1 range from 0 to 1, with 1 indicating optimal performance. 
	
	\subsection{Evaluation procedure}
	We train all models on imbalanced datasets, where the class frequency decreases as a function of the class index, and evaluate them on balanced test sets. This protocol measures robustness by assessing a model's ability to generate high-quality samples across all classes, regardless of their frequency during training.
	
	To evaluate performance across varying complexities, we utilize SVHN~\citep{svhn}, CIFAR100~\citep{cifar100,cifar100lt}, and CelebA~\citep{celeba} (see Appendix~\ref{datasets-mt3vae}). SVHN serves as a tractable baseline—simple enough to ensure convergence, yet rich enough to highlight generative discrepancies. CIFAR100 provides a more challenging benchmark with high semantic diversity, particularly testing performance in low-data regimes. Finally, CelebA enables a detailed analysis of generative quality and class balance in the presence of natural attributes.
	
	We construct Long-Tail (LT) variants of SVHN and CIFAR100 to explicitly control the training imbalance. Specifically, we induce imbalance via an exponential decay in sample counts after equalizing the initial class sizes. The imbalance ratio $\rho$ defines the disparity between the most and least frequent classes, with the sample count $M_{y_i}$ for class $y_i$ given by:
	\begin{equation*}
		M_{y_i} = M \cdot \rho^{-\frac{y_i - 1}{K - 1}},
	\end{equation*}
	where $M$ denotes the original sample count per class and $y_i \in \{1, \dots, K\}$ represents the class index.
	
	For CelebA, we exploit the inherent imbalance of facial attributes. We select Mustache and Young to represent strong ($\rho \approx 25$) and moderate ($\rho \approx 3.5$) imbalance, respectively, alongside the balanced Male and Smiling attributes. We deliberately refrain from defining classes via attribute composition; such an approach would lead to a combinatorial expansion of the label space and result in prohibitive sample sparsity per class.
	
	We employ the FID~\citep{fid} as the primary global metric to quantify the distribution shift between generated and real samples. However, as FID can be biased in low-sample regimes, we complement it with Precision, Recall, and F1-score to provide a granular, per-class evaluation of the proposed model. We detail our experimental settings in Appendix~\ref{exp_setup_mt3vae}.
	
	\subsection{$\tau$ parameter study}
\begin{figure}[ht]
		\centering
		\begin{subfigure}[b]{0.325\columnwidth}
			\centering
			\includegraphics[width=\linewidth]{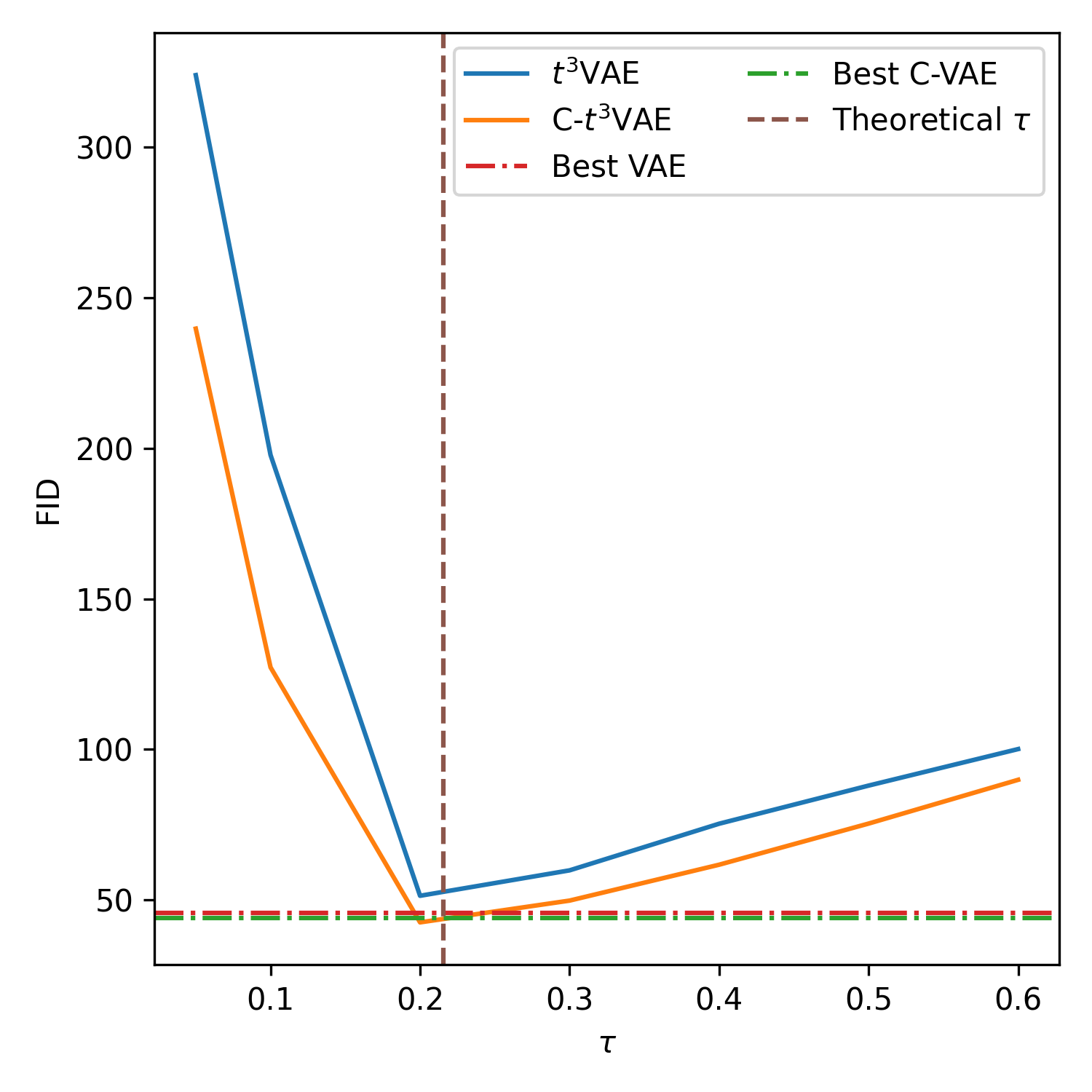}
			\caption{SVHN-LT}
		\end{subfigure}
		\hfill
		\begin{subfigure}[b]{0.325\columnwidth}
			\centering
			\includegraphics[width=\linewidth]{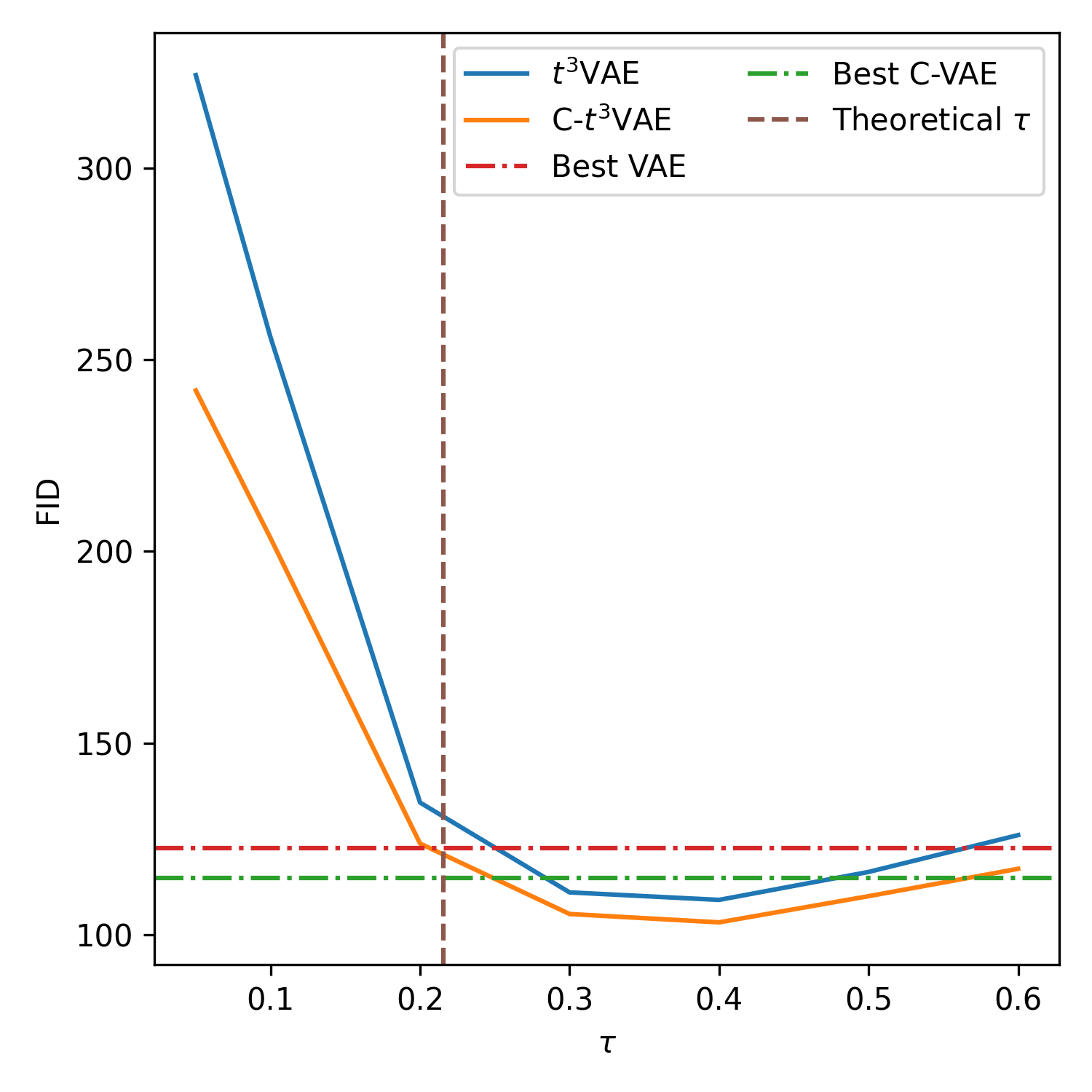}
			\caption{CIFAR100-LT}
		\end{subfigure}
		\hfill
		\begin{subfigure}[b]{0.325\columnwidth}
			\centering
			\includegraphics[width=\linewidth]{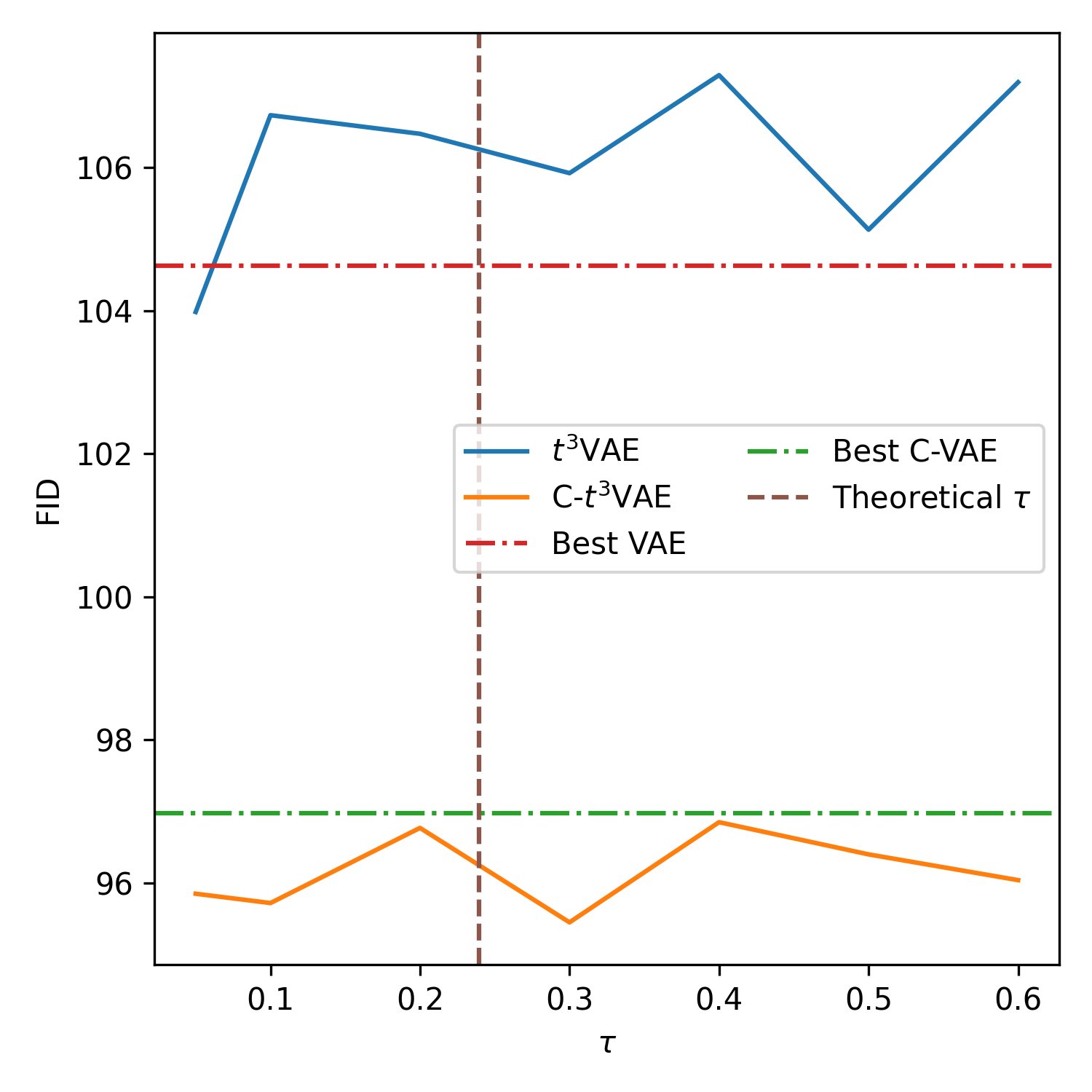}
			\caption{CelebA}
		\end{subfigure}
		
		\caption{FID score as a function of $\tau$ for the $t^3$VAE and C-$t^3$VAE models. Results are for the imbalance ratio $\rho=100$ for the SVHN-LT and CIFAR100-LT, and the Mustache attribute \mbox{($\rho=25$)} for CelebA. Other imbalance ratios' results paint a similar picture and are provided in Appendix (\ref{tau tunning}). The horizontal dashed lines is the FID value of the best performing VAE and C-VAE on each dataset and the vertical dashed line is the value of $\tau$ as derived in Eq. (\ref{kim tau}). We note that the used models in these figures have optimized $\beta$ and~$\nu$ hyper-parameters.}
		\label{tau study}
	\end{figure}

	Figure \ref{tau study} illustrates the impact of the sampling standard deviation $\tau$ on FID. Models based on the Student's $t$-distribution benefit from higher standard deviation on \mbox{CIFAR100-LT} compared to SVHN-LT. Notably, C-$t^3$VAE outperforms C-VAE for \mbox{$\tau \in [0.25; 0.55]$} on CIFAR100-LT, \mbox{$\tau \in [0.19; 0.28]$} on SVHN-LT, and for all $\tau$ values on CelebA. Additionally, it surpasses the $t^3$VAE FID for all $\tau$ values and across all datasets, underlining the importance of equal per-class prior mass. 
	
	Moreover, for both Student's $t$-distribution based models, the optimal FID score for SVHN-LT occurs near the theoretically derived $\tau$ value. However, for the more complex \mbox{CIFAR100-LT} dataset, the optimal $\tau$ is higher than the theoretical value $\tau=0.4$. The analytically derived $\tau$ provides a principled initialization grounded in divergence geometry. Dataset-specific deviations reflect encoder–decoder capacity limits rather than theoretical inconsistency, highlighting interactions between prior heaviness and representation complexity. For CelebA, $\tau$ has minimal impact on performance, likely due to the lower variability in the dataset's images (All faces are visually similar, reducing the impact of latent scale). Developing tighter theoretical characterizations of $\tau$ that account for dataset complexity and model capacity remains an interesting direction for future work.
	
	\subsection{Optimized Model Results Discussion}
	After optimizing the hyperparameters of the various models tested in this work, we present their generation FID scores in Table~\ref{quantitaive svhn cifar}. We provide results for optimized $\beta$ models and non-$\beta$ models ($\beta=1$) to underscore the importance of this parameter, which was not explored in the original $t^3$VAE work \citep{t3vae}.
	
	\begin{table}[htb]
		\centering
		\caption{Generation FID results on the SVHN-LT, CIFAR100-LT and CelebA datasets. For the SVHN-LT and CIFAR100-LT datasets we use different imbalance ratios $\rho \in \{100, 50, 10, 1\}$. However, for the CelebA dataset we use the Mustache, Young, Male and Smiling attributes which have imbalance ratios of  $25$, $3.5$, $1.4$ and $1$ respectively. The $\beta$ models undertook an optimization of the $\beta$ hyper-parameter while non-$\beta$ models have $\beta=1$. All models have optimized $\nu$ and $\tau$ hyper-parameters. The attributes for the CelebA dataset column indicate which attribute is used to condition the conditional models and balance the test set.}
		\label{quantitaive svhn cifar}
		\resizebox{\textwidth}{!}{%
			\begin{tabular}{l|cccc||cccc||cccc}
				\toprule
				& \multicolumn{4}{c||}{SVHN-LT} & \multicolumn{4}{c||}{CIFAR100-LT} & \multicolumn{4}{c}{CelebA} \\
				Models & $\rho=$100 & 50 & 10 & 1 & $\rho=$100 & 50 & 10 & 1 & Mustache & Young & Male & Smiling  \\\midrule
				VAE & 93.89 & 91.91 & 91.66 & 92.16 & 163.66 & 162.91 & 165.47 & 166.46 & 110.58 & 92.01 & 110.58  & 82.05 \\
				$\beta$-VAE & 47.11 & 49.81 & 45.70 & 43.48 & 122.62 & 123.07 & 123.72 & 124.43 & 104.63 & 92.87 & \phantom{0}87.96 & 83.15 \\
				C-VAE & 74.75 & 70.40 & 72.30 & 74.16 & 157.90 & 163.67 & 162.09 & 163.24 & \phantom{0}96.98 & 89.17 & \phantom{0}86.17 & \textbf{78.35} \\
				$\beta$-C-VAE & 48.39 & 46.39 & 43.97 & \textbf{43.87} & 114.88 & 118.89 & 114.89 & 118.21 & \phantom{0}98.35 & 85.53 & \phantom{0}\textbf{79.76} & 78.46 \\\midrule
				$t^3$VAE & 57.07 & 54.30 & 52.10 & 51.52 & 136.63 & 137.24 & 138.92& 135.23 & 105.80 & 88.07 & \phantom{0}83.62 & 78.90\\
				$\beta$-$t^3$VAE & 51.62 & 49.55& 48.93 & 45.37 & 109.11 & 107.93 & 108.97 & \textbf{111.00} & 105.86 & 88.21 & \phantom{0}83.83 & 78.89 \\
				C-$t^3$VAE & 47.09 & 46.29 & 47.43 & 51.32 & 125.48 & 127.96 & 130.28& 129.40 & 101.18 & 87.07 & \phantom{0}81.92 & 80.97 \\
				$\beta$-C-$t^3$VAE & \textbf{44.02} & \textbf{42.60} & \textbf{42.01} & 44.49 & \textbf{103.25} & \textbf{102.99} & \textbf{105.92} & 112.37 & \phantom{0}\textbf{95.82} & \textbf{82.61} & \phantom{0}81.65 & 80.08 \\\bottomrule
			\end{tabular}
		}	
	\end{table}
	
\begin{figure}[ht]
		\centering
		\includegraphics[width=.5\textwidth]{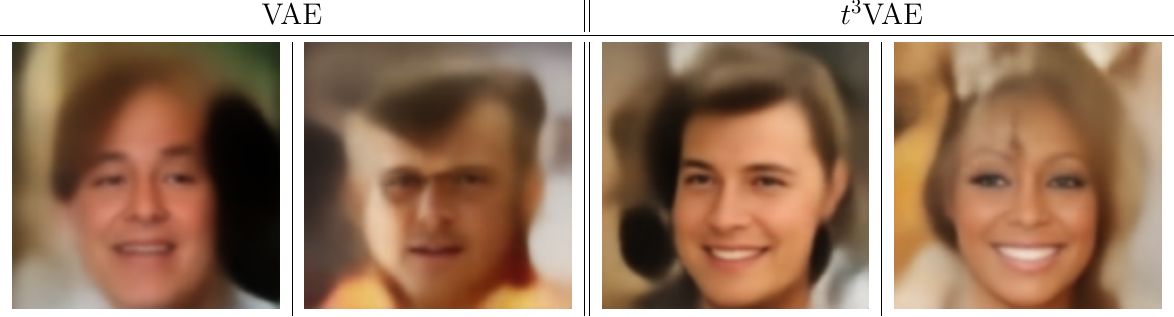}
		\caption{Sample synthetic images from the optimized VAE and $t^3$VAE models trained on the CelebA dataset. No class conditioning is possible for these models.}
		\label{samples vae t3vae}
	\end{figure}

	From Table~\ref{quantitaive svhn cifar}, $t^3$VAE improves over VAE, highlighting the generative advantage of the \mbox{Student's $t$-distribution} prior over a Gaussian one, in addition to the reconstruction advantage noted in \citep{t3vae}. Optimizing $\beta$ improves $t^3$VAE FID over VAE on CIFAR100-LT and CelebA, while remaining competitive on SVHN-LT. Qualitative results in Figure~\ref{samples vae t3vae} show that $t^3$VAE produces sharper synthetic images on CelebA than VAE.
	
\begin{figure}[htb]
		\centering
		\includegraphics[width=\textwidth]{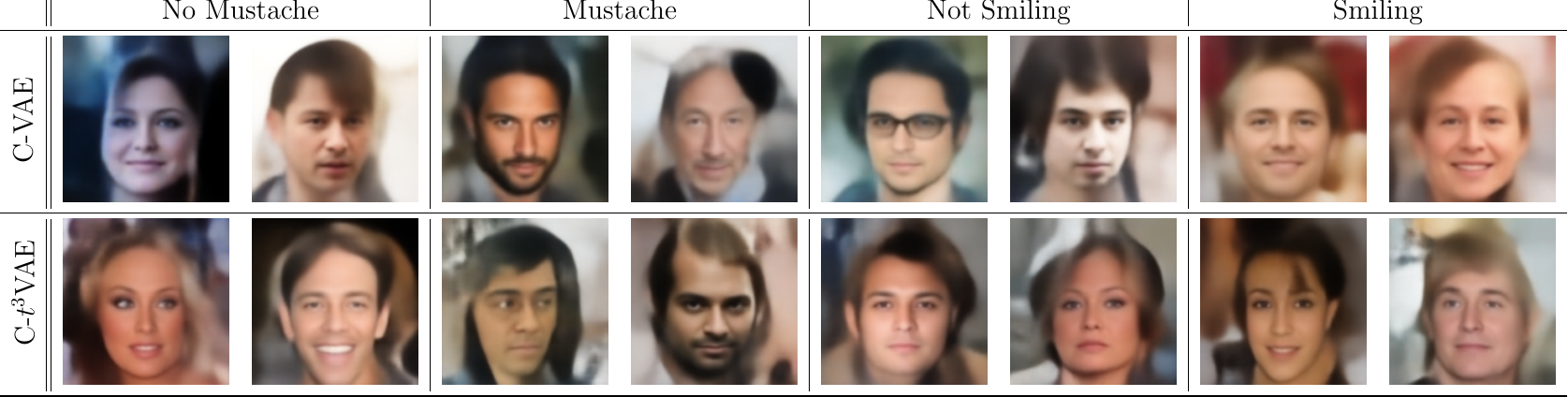}
		\caption{Sample synthetic images for the optimized C-VAE and \mbox{C-$t^3$VAE} trained on specific attributes of the CelebA dataset.}
		\label{samples gmvae}
	\end{figure}

	For class-conditional models, optimized \mbox{C-$t^3$VAE} yields strong FID improvements across imbalanced settings on SVHN-LT and CIFAR100-LT, and on heavily imbalanced CelebA attributes. As shown in Table~\ref{quantitaive svhn cifar}, it achieves gains of up to 4, 5, and 10 FID points over \mbox{$\beta$-$t^3$VAE} on imbalanced settings of SVHN-LT, \mbox{CIFAR100-LT}, and CelebA, respectively. This supports the role of uniform prior mass allocation in high-imbalance regimes. Moreover, \mbox{$\beta$-C-$t^3$VAE} reduces FID by up to 4 and 15 points over C-VAE on SVHN-LT and CIFAR100-LT, respectively. For CelebA, \mbox{$\beta$-C-$t^3$VAE} achieves the best results on heavily imbalanced attributes like Mustache, indicating improved generation for underrepresented classes. The gain over C-VAE follows from the Student's $t$-distribution latent prior and its ability to better capture intra-class long-tail structure. Qualitative samples of conditional optimized models (Figure~\ref{samples gmvae}) show sharper facial features for \mbox{C-$t^3$VAE} than C-VAE, notably on the imbalanced Mustache attribute. Overall, C-$t^3$VAE exhibits the most consistent performance in high-imbalance regimes within the VAE family.
	
	\subsection{Per-class evaluation}
	In this section, we evaluate the conditional models on a per-class basis. Since FID can be biased on small datasets and offers limited insight as a single scalar metric, we rely on Precision, Recall, and F1 metrics \citep{precision_recall}. Our results on CelebA are shown in Figure~\ref{precision recall f1 celeba}, with additional results for SVHN-LT and CIFAR100-LT included in Appendix~\ref{per class eval}.
	
	\begin{figure}[h]
		\centering
		\begin{subfigure}[b]{0.325\columnwidth}
			\centering
			\includegraphics[width=\linewidth]{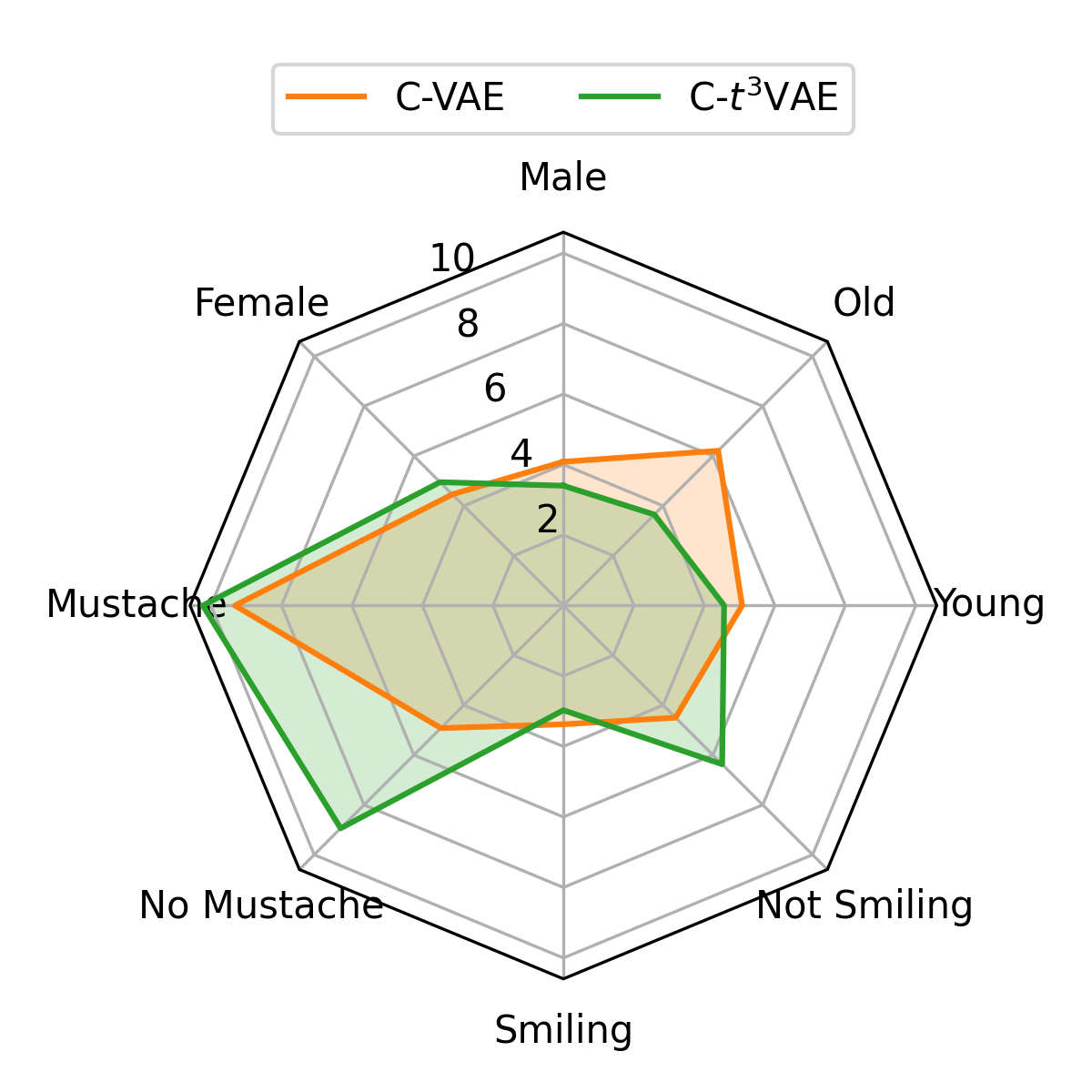}
			\caption{Recall}
		\end{subfigure}
		\hfill
		\begin{subfigure}[b]{0.325\columnwidth}
			\centering
			\includegraphics[width=\linewidth]{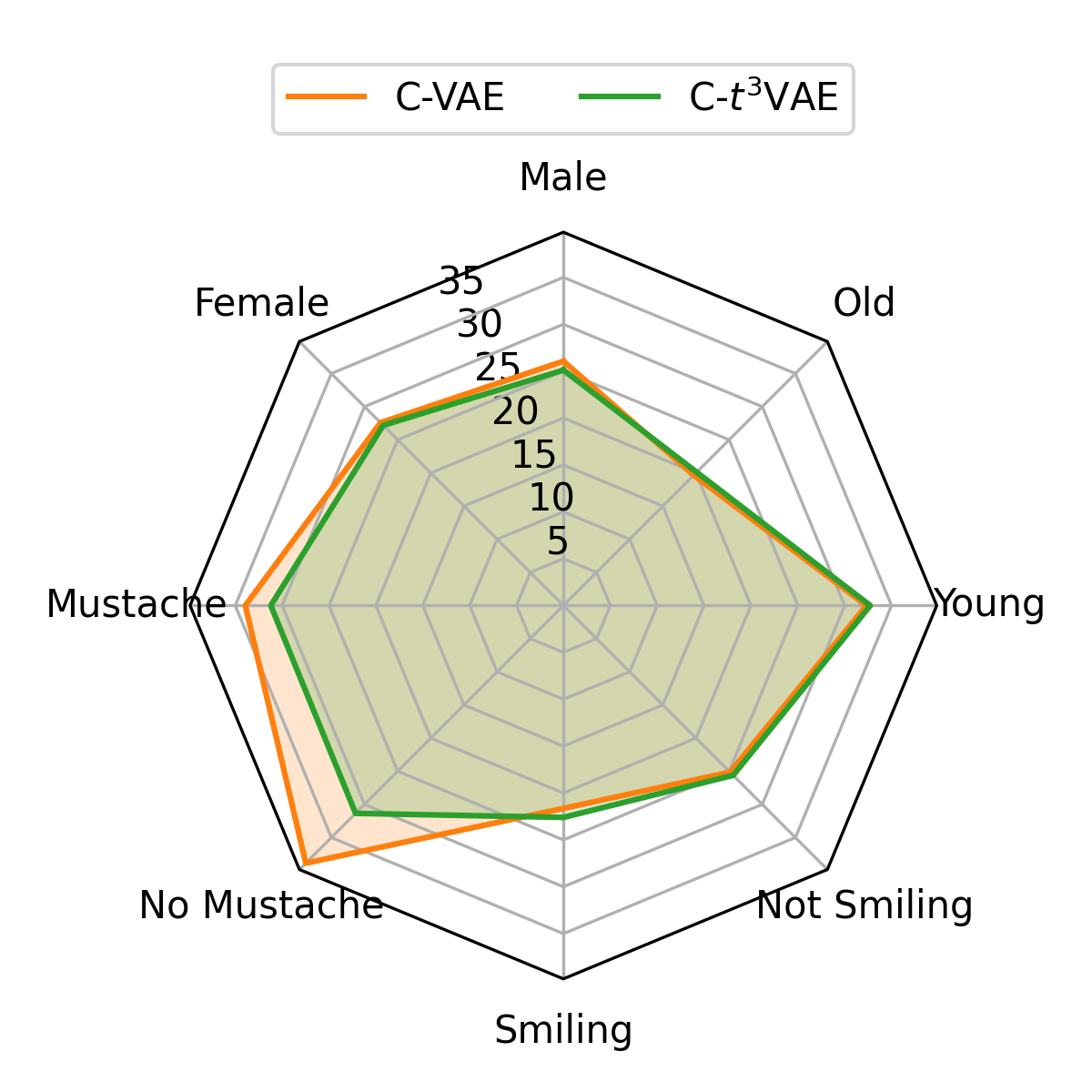}
			\caption{Precision}
		\end{subfigure}
		\hfill
		\begin{subfigure}[b]{0.325\columnwidth}
			\centering
			\includegraphics[width=\linewidth]{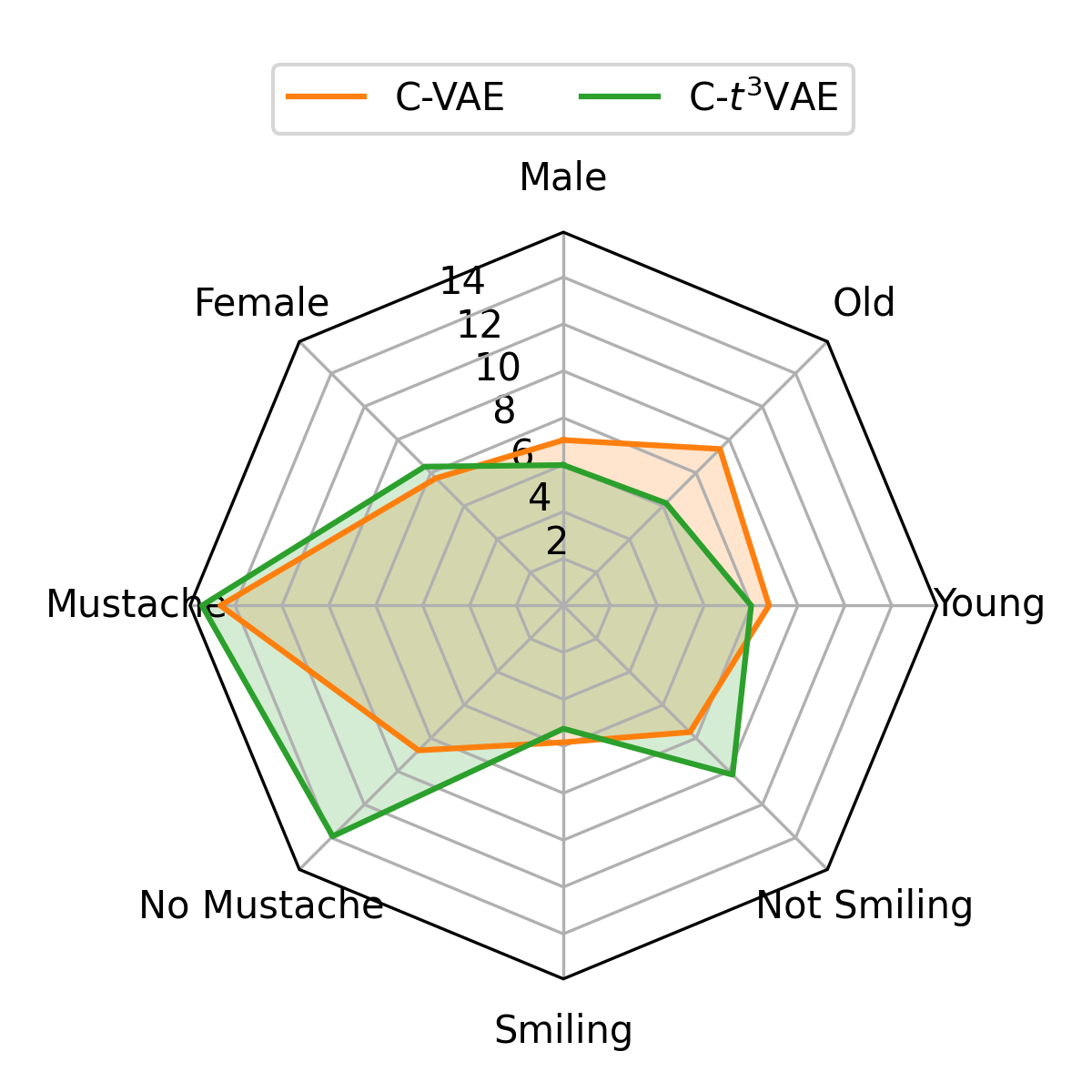}
			\caption{F1 score}
		\end{subfigure}
		
		\caption{Per-class generative metrics on CelebA after optimization of all hyper-parameters notably $\beta$, $\nu$ and $\tau$. We note that the imbalance ratio of the Mustache, Young, Male and Smiling factors $\rho$ are $25$, $3.5$, $1.4$ and $1$ respectively.}
		\label{precision recall f1 celeba}
	\end{figure}
	
\begin{wrapfigure}{r}{.55\textwidth}
		\centering
		\vspace{-.3cm}
		\includegraphics[width=.5\columnwidth]{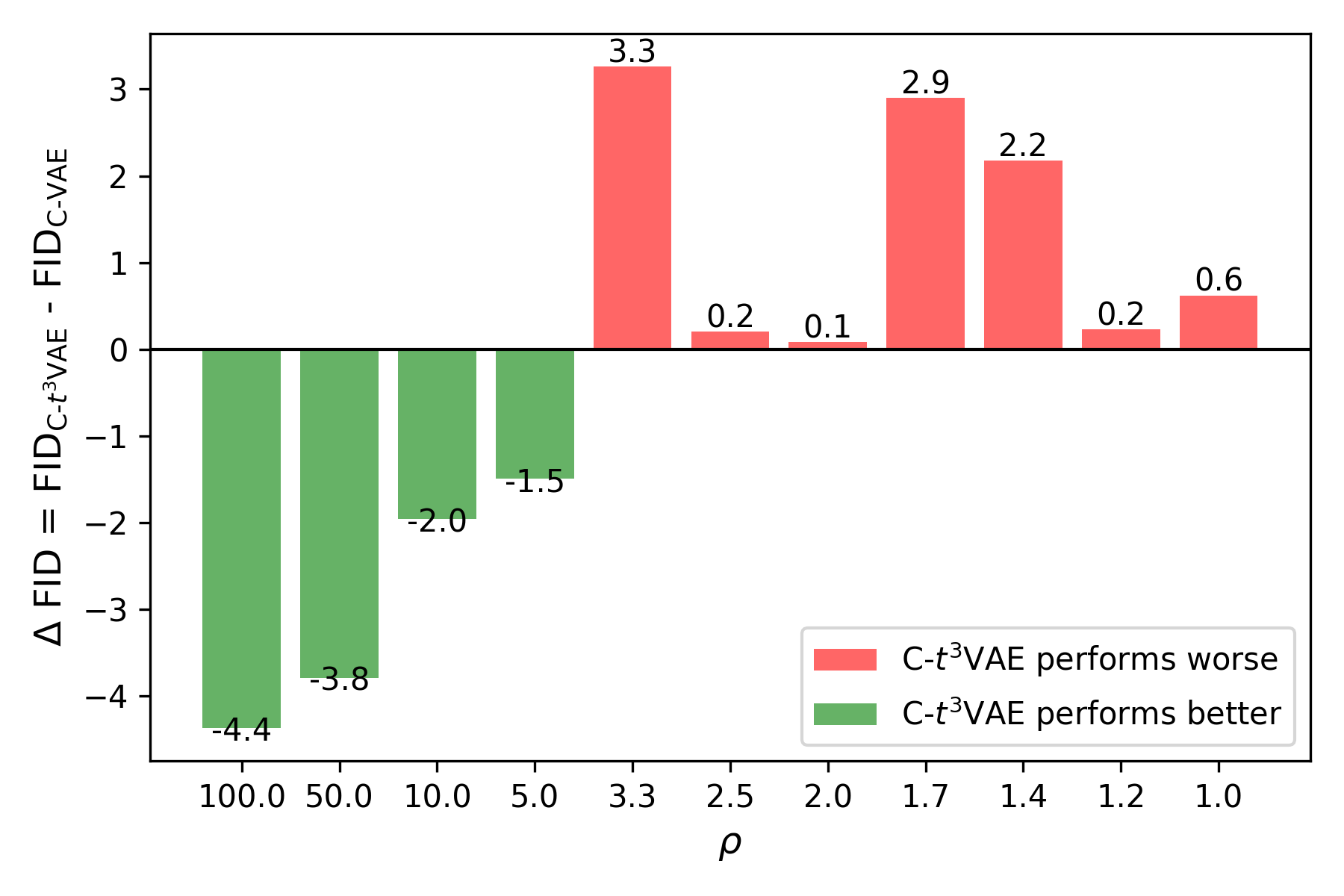}
		\caption{C-$t^3$VAE versus C-VAE models under varying imbalance ratios on the SVHN-LT dataset.}
		\label{fine grained evaluation svhn}
	\end{wrapfigure}

	On CelebA (Figure~\ref{precision recall f1 celeba}), C-$t^3$VAE improves Recall and F1 on the most imbalanced attribute (Mustache), but not on more balanced ones (Male, Smiling), which is expected. When class imbalance is moderate (e.g., Young attribute with $\rho = 3.5$), C-VAE slightly outperforms C-$t^3$VAE, indicating that Gaussian priors suffice when encoder variance and decoder flexibility can compensate for small frequency skew. This suggests the presence of a regime where Gaussian priors may suffice. To explore this, we vary the imbalance ratio on SVHN-LT from $100$ to $1$ and plot the results in Figure~\ref{fine grained evaluation svhn}. This figure shows a threshold around $\rho \approx 5$: below it C-VAE can be better, while above it \mbox{C-$t^3$VAE} has the advantage. As the imbalance ratio increases, the FID gap grows in favor of C-$t^3$VAE. For moderate imbalance ($\rho < 5$), encoder variance and decoder flexibility can compensate for frequency skew; beyond this regime, posterior--prior mismatch grows nonlinearly, and Gaussian priors increasingly compress minority-class latent regions. Heavy-tailed class-conditional priors mitigate this compression by permitting larger dispersion without penalizing minority likelihood, explaining the observed transition.
	
	For SVHN-LT (Figures in Appendix~\ref{per class eval}), \mbox{C-$t^3$VAE} achieves higher Recall and competitive Precision compared to C-VAE, indicating better mode coverage while maintaining image quality, which yields higher F1 scores, especially on tail classes. For CIFAR100-LT (Figures in Appendix~\ref{per class eval}), C-VAE often attains high Precision but near-zero Recall, indicating mode collapse. In contrast, C-$t^3$VAE preserves Recall at the cost of slightly lower Precision, resulting in improved F1 scores. Consequently, on SVHN-LT, CIFAR100-LT, and CelebA, C-$t^3$VAE is the most reliable method in our study for high-imbalance settings within the VAE family.

	\section{Conclusion}
	In this work, we introduced C-$t^3$VAE. This class-conditional generative model mitigates frequency-aligned latent allocation effects observed under imbalance. It uses per-class Student’s $t$-distributions in the latent space, paired with a theoretically derived, equal-weight sampling scheme. This design allocates uniform prior mass across all classes while capturing heavy-tailed intra-class variations. Furthermore, these structural improvements do not sacrifice efficiency. The computational complexity of C-$t^3$VAE matches that of standard conditional VAEs.
	
	We evaluated our model on SVHN-LT, CIFAR100-LT, and CelebA. After optimizing the hyperparameters $\beta$, $\nu$, and $\tau$, C-$t^3$VAE consistently outperforms baseline models in high-imbalance regimes. It achieves up to a 15-point FID improvement over $t^3$VAE and conditional VAEs. Additionally, per-class Precision, Recall, and F1 metrics confirm superior mode coverage for tail classes. We also identified a critical imbalance ratio threshold at $\rho \approx 5$. For milder imbalances ($\rho \lesssim 5$), Gaussian-based models remain competitive. However, as imbalance grows beyond this threshold, C-$t^3$VAE demonstrates a significant and widening advantage.
	
	Future research should explore extending the C-$t^3$VAE framework to multi-label settings. The MultiFacet VAE model \citep{multifacet} provides a promising starting point for handling complex, overlapping annotations. Furthermore, our experiments show that the sampling variance $\tau$ heavily impacts image quality. This parameter currently requires dataset-specific tuning. Therefore, developing tighter theoretical bounds and adaptive latent-space sampling methods remains a crucial next step for high-fidelity VAE generation.
	
	\section*{Acknowledgment}
	This work received access to the High-Performance Computing (HPC) resources of MesoPSL, financed by the Region Île-de-France and the Equip@Meso project (reference ANR-10-EQPX-29-01) of the Investissements d’avenir program supervised by the Agence nationale pour la recherche.
	

	\bibliographystyle{tmlr}
	
	\newpage
	\begin{center}
		{\Large \textbf{Supplementary Material}}
	\end{center}
	
	\appendix
	
	\section{Priors derivations}
	\label{prior derivation}
	In this section, we present our derivations of the different prior distributions defining our proposed C-$t^3$-VAE model. Starting from the proposed joint distribution :
	\begin{equation*}
		p_\theta(x,z\vert y)= \frac{C_{\nu, m+n}}{\vert \Sigma_x \vert^{\frac{1}{2}} \vert \Sigma_y \vert^{\frac{1}{2}}}  \left[ 1 + \frac{(z-\mu_y)^\top \Sigma_y^{-1}(z-\mu_y) + (x-\mu_\theta(z))^\top \Sigma_x^{-1}(x-\mu_\theta(z))}{\nu} \right]^{-\frac{\nu+m+n}{2}}.
	\end{equation*}
	To calculate the prior distribution on the latent space we marginalize out $x$ as follows :
	\begin{align*}
		p(z\vert y) &= \int p_\theta(x,z\vert y) dx \\
		&= \int C_{\nu, m+n} \vert \Sigma_x \vert^{-\frac{1}{2}} \vert \Sigma_y \vert^{-\frac{1}{2}} \Bigg[ 1 + \frac{(z-\mu_y)^\top \Sigma_y^{-1}(z-\mu_y)}{\nu} + \frac{(x-\mu_\theta(z))^\top \Sigma_x^{-1}(x-\mu_\theta(z))}{\nu} \Bigg]^{-\frac{\nu+m+n}{2}} dx \\
		&= C_{\nu, m+n} \vert \Sigma_x \vert^{-\frac{1}{2}} \vert \Sigma_y \vert^{-\frac{1}{2}} \left[ 1 + \frac{1}{\nu} (z-\mu_y)^\top \Sigma_y^{-1}(z-\mu_y) \right]^{-\frac{\nu+m+n}{2}} \nonumber \\
		& \quad \times \int \left( 1 + \frac{(1+\nu^{-1}m)(x-\mu_\theta(z))^\top \Sigma_x^{-1}(x-\mu_\theta(z))}{ \left(1 + \nu^{-1} (z-\mu_y)^\top \Sigma_y^{-1}(z-\mu_y)\right) (\nu+m)}  \right)^{-\frac{\nu+m+n}{2}} dx .
	\end{align*}
	Given that :
	\begin{align*}
		\int C_{\nu+m, n} \vert \Sigma \vert^{-\frac{1}{2}} \left( 1+ \frac{(x-\mu)^\top \Sigma^{-1}(x-\mu)}{\nu+m} \right)^{-\frac{\nu+m+n}{2}} dx &= 1 \\
		\Rightarrow \int \left( 1+ \frac{(x-\mu)^\top \Sigma^{-1}(x-\mu)}{\nu+m} \right)^{-\frac{\nu+m+n}{2}} dx &= C_{\nu+m, n}^{-1} \vert \Sigma \vert^{\frac{1}{2}},
	\end{align*}
	and when setting : 
	$$
	\Sigma^{-1} =  \frac{(1+\nu^{-1}m) \Sigma_x^{-1}}{1 + \nu^{-1} (z-\mu_y)^\top \Sigma_y^{-1}(z-\mu_y)},
	$$
	We get :
	\begin{align*}
		&\int \left( 1 + \frac{(1+\nu^{-1}m)(x-\mu_\theta(z))^\top \Sigma_x^{-1}(x-\mu_\theta(z))}{ \left(1 + \nu^{-1} (z-\mu_y)^\top \Sigma_y^{-1}(z-\mu_y)\right) (\nu+m)}  \right)^{-\frac{\nu+m+n}{2}} dx \\
		&\qquad \qquad \qquad \qquad \qquad \qquad= C_{\nu+m, n}^{-1} \left\vert \left( \frac{(1+\nu^{-1}m) \Sigma_x^{-1}}{1 + \nu^{-1} (z-\mu_y)^\top \Sigma_y^{-1}(z-\mu_y)} \right)^{-1} \right\vert^{\frac{1}{2}} \\
		&\qquad \qquad \qquad \qquad \qquad \qquad= C_{\nu+m, n}^{-1} \left\vert  \frac{1 + \nu^{-1} (z-\mu_y)^\top \Sigma_y^{-1}(z-\mu_y)}{(1+\nu^{-1}m)}  \Sigma_x \right\vert^{\frac{1}{2}} \\
		&\qquad \qquad \qquad \qquad \qquad \qquad= C_{\nu+m, n}^{-1} \left(  \frac{1 + \nu^{-1} (z-\mu_y)^\top \Sigma_y^{-1}(z-\mu_y)}{1+\nu^{-1}m} \right) ^{\frac{n}{2}}  \left\vert \Sigma_x \right\vert^{\frac{1}{2}}.
	\end{align*}
	Therefore, $p(z\vert y)$ simplifies to :
	\begin{align*}
		p(z\vert y) &= C_{\nu, m+n} \vert \Sigma_x \vert^{-\frac{1}{2}} \vert \Sigma_y \vert^{-\frac{1}{2}} \left[ 1 + \frac{1}{\nu} (z-\mu_y)^\top \Sigma_y^{-1}(z-\mu_y) \right]^{-\frac{\nu+m+n}{2}} C_{\nu+m, n}^{-1} \left\vert \Sigma_x \right\vert^{\frac{1}{2}} \\ 
		&\quad \times \left(  \frac{1 + \nu^{-1} (z-\mu_y)^\top \Sigma_y^{-1}(z-\mu_y)}{1+\nu^{-1}m} \right) ^{\frac{n}{2}}   \\
		&= C_{\nu, m+n} C_{\nu+m, n}^{-1} \left(1 + \frac{m}{\nu} \right) ^{-\frac{n}{2}} \vert \Sigma_y \vert^{-\frac{1}{2}} \left[ 1 + \frac{1}{\nu} (z-\mu_y)^\top \Sigma_y^{-1}(z-\mu_y) \right]^{-\frac{\nu+m}{2}}    \\
		&= C_{\nu, m} \vert \Sigma_y \vert^{-\frac{1}{2}} \left[ 1 + \frac{1}{\nu} (z-\mu_y)^\top \Sigma_y^{-1}(z-\mu_y) \right]^{-\frac{\nu+m}{2}} \\
		&= t_m(z \vert \mu_y, \Sigma_y, \nu).
	\end{align*}
	Here and in the following, we use the fact
	$$
	C_{\nu, m+n} = C_{\nu+m, n} C_{\nu, m} \left(1+\frac{m}{\nu}\right)^{\frac{n}{2}}.
	$$
	Besides, the prior distribution over the output of the decoder model $p(x\vert z,y)$ can be derived as follows :
	\begin{align*}
		p_\theta(x\vert z,y) &= \frac{p_\theta(x,z\vert y)}{p(z\vert y)} \\
		&= \frac{C_{\nu, m+n}}{\vert \Sigma_x \vert^{\frac{1}{2}} \vert \Sigma_y \vert^{\frac{1}{2}}}  \left[ 1 + \frac{(z-\mu_y)^\top \Sigma_y^{-1}(z-\mu_y) + (x-\mu_\theta(z))^\top \Sigma_x^{-1}(x-\mu_\theta(z))}{\nu} \right]^{-\frac{\nu+m+n}{2}} \nonumber \\
		&\quad \times C_{\nu, m}^{-1} \vert \Sigma_y \vert^{\frac{1}{2}} \left[ 1 + \frac{1}{\nu} (z-\mu_y)^\top \Sigma_y^{-1}(z-\mu_y) \right]^{\frac{\nu+m}{2}} \\
		&= C_{\nu+m, n}  \vert \Sigma_x \vert^{-\frac{1}{2}} \left(1 + \frac{m}{\nu} \right) ^{\frac{n}{2}} \left[ 1 + \frac{1}{\nu} (z-\mu_y)^\top \Sigma_y^{-1}(z-\mu_y) \right]^{-\frac{n}{2}} \nonumber \\ 
		&\quad \times\left( 1 + \frac{(1+\nu^{-1}m)(x-\mu_\theta(z))^\top \Sigma_x^{-1}(x-\mu_\theta(z))}{ \left(1 + \nu^{-1} (z-\mu_y)^\top \Sigma_y^{-1}(z-\mu_y)\right) (\nu+m)}  \right)^{-\frac{\nu+m+n}{2}} \\
		&= t_n \left( x \middle\vert \mu_\theta(z), \frac{ \left(1 + \nu^{-1} (z-\mu_y)^\top \Sigma_y^{-1}(z-\mu_y)\right)}{(1+\nu^{-1}m)} \Sigma_x, \nu+m \right).
	\end{align*}
	
	\newpage
	\section[Loss derivation]{Loss function derivation}
	\label{loss derivation}
	In this section, we derive the loss function of C-$t^3$-VAE. We start by calculating the different double integrals $\iint p_\theta(x,z\vert y)^{1+\gamma} dx dz$, $\iint q_\phi(x,z\vert y) p_\theta(x,z\vert y)^{\gamma} dx dz$, and $\iint q_\phi(x,z \vert y)^{1+\gamma} dx dz$.
	
	Firstly,
	\begin{align*}
		\iint p_\theta(x,z\vert y)^{1+\gamma} dx dz &= \mathbb{E}_{z \sim p(z\vert y)} \mathbb{E}_{x \sim p_\theta(x\vert z, y)} \left[ p_\theta(x,z\vert y)^\gamma \right] \\
		&= C_{\nu, m+n}^{\gamma} \vert \Sigma_x \vert^{-\frac{\gamma}{2}} \vert \Sigma_y \vert^{-\frac{\gamma}{2}} \mathbb{E}_{z} \mathbb{E}_{x} \Bigg[ 1 + \frac{(z-\mu_y)^\top \Sigma_y^{-1}(z-\mu_y)}{\nu} \\
		&\quad + \frac{(x-\mu_\theta(z))^\top \Sigma_x^{-1}(x-\mu_\theta(z))}{\nu} \Bigg] \\
		&= C_{\nu, m+n}^{\gamma} \vert \Sigma_x \vert^{-\frac{\gamma}{2}} \vert \Sigma_y \vert^{-\frac{\gamma}{2}} \mathbb{E}_{z} \bigg[ 1 + \nu^{-1}(z-\mu_y)^\top \Sigma_y^{-1}(z-\mu_y) \\
		&\quad + \nu^{-1}\mathbb{E}_{x} \left[\tr(\Sigma_x^{-1}(x-\mu_\theta(z))(x-\mu_\theta(z))^\top) \right] \bigg] \\
		&= C_{\nu, m+n}^{\gamma} \vert \Sigma_x \vert^{-\frac{\gamma}{2}} \vert \Sigma_y \vert^{-\frac{\gamma}{2}} \mathbb{E}_{z} \Bigg[ 1 + \nu^{-1}(z-\mu_y)^\top \Sigma_y^{-1}(z-\mu_y) \\
		&\quad  + \nu^{-1} \tr\left( \Sigma_x^{-1} \Sigma_x \frac{\nu+m}{\nu+m-2} \frac{ \left(1 + \nu^{-1} (z-\mu_y)^\top \Sigma_y^{-1}(z-\mu_y)\right)}{(1+\nu^{-1}m)}  \right) \Bigg]
	\end{align*}
	Here, we use the following identities
	$$
	(k-p)^\top H^{-1}(k-p) = \tr \left(H^{-1}(k-p)(k-p)^\top\right); \qquad \mathbb{E}[\tr(\cdot)] = \tr(\mathbb{E}[\cdot])
	$$
	and the covariance of a multivariate Student's t distribution $p \sim t(\mu; \Sigma;\nu)$ is $\frac{\nu}{\nu-2}\Sigma$. Consequently, and after a few simplifications we get
	
	\begin{align*}
		\iint p_\theta(x,z\vert y)^{1+\gamma} dx dz&= C_{\nu, m+n}^{\gamma} \vert \Sigma_x \vert^{-\frac{\gamma}{2}} \vert \Sigma_y \vert^{-\frac{\gamma}{2}} \mathbb{E}_{z} \bigg[ 1 + \nu^{-1}(z-\mu_y)^\top \Sigma_y^{-1}(z-\mu_y) \\
		&\quad + \frac{n}{\nu+m-2} \left(1 + \nu^{-1} (z-\mu_y)^\top \Sigma_y^{-1}(z-\mu_y)\right) \bigg] \\
		&= C_{\nu, m+n}^{\gamma} \vert \Sigma_x \vert^{-\frac{\gamma}{2}} \vert \Sigma_y \vert^{-\frac{\gamma}{2}} \mathbb{E}_{z} \Bigg[ \left( 1 + \frac{n}{\nu+m-2} \right)\\
		&\quad \times \Big(1 + \nu^{-1} (z-\mu_y)^\top \Sigma_y^{-1}(z-\mu_y)\Big) \Bigg] \\
		&= C_{\nu, m+n}^{\gamma} \vert \Sigma_x \vert^{-\frac{\gamma}{2}} \vert \Sigma_y \vert^{-\frac{\gamma}{2}} \left( 1 + \frac{n}{\nu+m-2} \right) \\
		&\quad \times \Big(1 + \nu^{-1} \mathbb{E}_{z} \left[ (z-\mu_y)^\top \Sigma_y^{-1}(z-\mu_y)\right]\Big) \\
		&= C_{\nu, m+n}^{\gamma} \vert \Sigma_x \vert^{-\frac{\gamma}{2}} \vert \Sigma_y \vert^{-\frac{\gamma}{2}} \left( 1 + \frac{n}{\nu+m-2} \right) \\
		&\quad \times \Big(1 + \nu^{-1} \mathbb{E}_{z}\left[\tr(\Sigma_y^{-1}(z-\mu_y)(z-\mu_y)^\top\right]\Big) \\
		&= C_{\nu, m+n}^{\gamma} \vert \Sigma_x \vert^{-\frac{\gamma}{2}} \vert \Sigma_y \vert^{-\frac{\gamma}{2}} \left( 1 + \frac{n}{\nu+m-2} \right) \left(1 + \frac{m}{\nu-2}\right) \\
		&= C_{\nu, m+n}^{\gamma} \vert \Sigma_x \vert^{-\frac{\gamma}{2}} \vert \Sigma_y \vert^{-\frac{\gamma}{2}} \left(1 + \frac{m+n}{\nu-2}\right).
	\end{align*}
	
	Secondly,
	\begin{align*}
		\iint q_\phi(x,z\vert y) p_\theta(x,z\vert y)^{\gamma} dx dz &= \mathbb{E}_{x \sim p_{data}} \mathbb{E}_{z \sim q(z\vert x)} \left[ p_\theta(x,z\vert y)^{\gamma} \right] \\
		&= C_{\nu, m+n}^{\gamma} \vert \Sigma_x \vert^{-\frac{\gamma}{2}} \vert \Sigma_y \vert^{-\frac{\gamma}{2}} \mathbb{E}_{x} \mathbb{E}_{z} \Bigg[ 1 + \frac{(z-\mu_y)^\top \Sigma_y^{-1}(z-\mu_y)}{\nu} \\
		&\quad + \frac{(x-\mu_\theta(z))^\top \Sigma_x^{-1}(x-\mu_\theta(z))}{\nu}\Bigg] \\
		&= C_{\nu, m+n}^{\gamma} \vert \Sigma_x \vert^{-\frac{\gamma}{2}} \vert \Sigma_y \vert^{-\frac{\gamma}{2}} \mathbb{E}_{x}\Bigg[ 1 + \frac{1}{\nu} \mathbb{E}_{z}\left[(z-\mu_y)^\top \Sigma_y^{-1}(z-\mu_y) \right] \\
		&\quad  + \frac{1}{\nu} \mathbb{E}_{z}\left[(x-\mu_\theta(z))^\top \Sigma_x^{-1}(x-\mu_\theta(z)) \right]\Bigg] .
	\end{align*}
	
	\noindent Simplifying $\mathbb{E}_{z}\left[(z-\mu_y)^\top \Sigma_y^{-1}(z-\mu_y) \right]$ :
	\begin{align*}
		\mathbb{E}_{z}\left[(z-\mu_y)^\top \Sigma_y^{-1}(z-\mu_y) \right] &= \mathbb{E}_{z}\left[\tr\left( \Sigma_y^{-1}(z-\mu_y)(z-\mu_y)^\top \right) \right] \\
		&= \mathbb{E}_{z}\left[\tr\left( \Sigma_y^{-1}(z-\mu(x)+\mu(x)-\mu_y)(z-\mu(x)+\mu(x)-\mu_y)^\top \right) \right] \\
		&= \mathbb{E}_{z}[\tr (\Sigma_y^{-1} ( (z-\mu(x))(z-\mu(x))^\top + (z-\mu(x))(\mu(x)-\mu_y)^\top \nonumber \\
		&+ (\mu(x)-\mu_y)(z-\mu(x))^\top +(\mu(x)-\mu_y)(\mu(x)-\mu_y)^\top ) )] \\
		&= \frac{\nu}{\nu+n-2} \tr\left( \Sigma_y^{-1} \Sigma_\phi(x) \right) +(\mu(x)-\mu_y)^\top\Sigma_y^{-1}(\mu(x)-\mu_y).
	\end{align*}
	
	Then, $\iint q(x,z\vert y) p_\theta(x,z\vert y)^{\gamma} dx dz$ simplifies to :
	\begin{align*}
		\iint q_\phi(x,z\vert y) p_\theta(x,z\vert y)^{\gamma} dx dz &= C_{\nu, m+n}^{\gamma} \vert \Sigma_x \vert^{-\frac{\gamma}{2}} \vert \Sigma_y \vert^{-\frac{\gamma}{2}} \mathbb{E}_{x}\Bigg[ 1 + \frac{1}{\nu} \mathbb{E}_{z}\left[(z-\mu_y)^\top \Sigma_y^{-1}(z-\mu_y) \right] \\
		&\quad + \frac{1}{\nu} \mathbb{E}_{z}\left[(x-\mu_\theta(z))^\top \Sigma_x^{-1}(x-\mu_\theta(z)) \right]\Bigg] 
	\end{align*}
	\begin{align*}
		\iint q_\phi(x,z\vert y) p_\theta(x,z\vert y)^{\gamma} dx dz &= C_{\nu, m+n}^{\gamma} \vert \Sigma_x \vert^{-\frac{\gamma}{2}} \vert \Sigma_y \vert^{-\frac{\gamma}{2}} \, \vert \Sigma_y \vert^{-\frac{1}{2}} \, \mathbb{E}_{x}\Bigg[ 
		1 + \frac{1}{\nu} \frac{\nu \tr\left( \Sigma_y^{-1} \Sigma_\phi(x) \right)}{\nu+n-2} \notag\\
		&\quad + \frac{(\mu(x)-\mu_y)^\top\Sigma_y^{-1}(\mu(x)-\mu_y) }{\nu} + \frac{\mathbb{E}_{z}\left[(x-\mu_\theta(z))^\top \Sigma_x^{-1}(x-\mu_\theta(z))\right]}{\nu} 
		\Bigg].
	\end{align*}
	
	Finally, the third term $\iint q_(x,z \vert y)^{1+\gamma} dx dz$ is 
	\begin{equation*}
		\iint q_\phi(x,z \vert y)^{1+\gamma} dx dz = C^{\gamma}_{\nu+n, m} \left( 1 + \frac{n}{\nu} \right)^{\frac{\gamma m}{2}} \left( 1 + \frac{m}{\nu+n-2} \right) \int \vert \Sigma_\phi(x) \vert^{-\frac{\gamma}{2}} p_{data}(x)^{1+\gamma} dx,
	\end{equation*}
	where this last double integral is equal to the one computed for the $t^3$-VAE. 
	
	Equipped with these formulas we can calculate the entropy $\mathcal{H}_\gamma$, cross-entropy $\mathcal{C}_\gamma$ and the $\gamma$-divergence $\mathcal{D}(q\vert \vert p)$ of our model. Firstly,
	\begin{align*}
		\mathcal{H}_\gamma &= - \left( \iint q(x,z)^{1+\gamma} dx dz\right)^{\frac{1}{1+\gamma}} \\
		&= - C^{\frac{\gamma}{1+\gamma}}_{\nu+n, m} \left( 1 + \frac{n}{\nu} \right)^{\frac{\gamma m}{2(1+\gamma)}} \left( 1 + \frac{m}{\nu+n-2} \right)^{\frac{1}{1+\gamma}} \left( \int \vert \Sigma_\phi(x) \vert^{-\frac{\gamma}{2}} p_{data}(x)^{1+\gamma} dx \right)^{\frac{1}{1+\gamma}},
	\end{align*}
	Which is similar to the one calculated in the $t^3$VAE model.

	\noindent Secondly,
	\begin{align*}
		\mathcal{C}_\gamma &= - \left( \iint q(x,z \vert y) p_\theta(x,z\vert y)^\gamma dx dz \right) \left( \iint p_\theta(x,z\vert y)^{1+\gamma} \right)^{-\frac{\gamma}{1+\gamma}} \\
		&= - C_{\nu, m+n}^{\gamma} \vert \Sigma_x \vert^{-\frac{\gamma}{2}} \vert \Sigma_y \vert^{-\frac{\gamma}{2}-\frac{1}{2}} \mathbb{E}_{x}\Bigg[ 
		1 + \frac{1}{\nu} \frac{\nu \tr\left( \Sigma_y^{-1} \Sigma_\phi(x) \right)}{\nu+n-2} + \frac{(\mu(x)-\mu_y)^\top\Sigma_y^{-1}(\mu(x)-\mu_y)}{\nu} \notag\\
		&\quad + \frac{1}{\nu} \mathbb{E}_{z}\left[(x-\mu_\theta(z))^\top \Sigma_x^{-1}(x-\mu_\theta(z)) \right]
		\Bigg] \left( C_{\nu, m+n}^{\gamma} \vert \Sigma_x \vert^{-\frac{\gamma}{2}} \vert \Sigma_y \vert^{-\frac{\gamma}{2}} \left(1 + \frac{m+n}{\nu-2}\right) \right)^{-\frac{\gamma}{1+\gamma}} \\
		&= - \left( C_{\nu, m+n}^{\gamma} \vert \Sigma_x \vert^{-\frac{\gamma}{2}} \vert \Sigma_y \vert^{-\frac{\gamma}{2}} \right)^{\frac{1}{1+\gamma}} \vert \Sigma_y \vert^{-\frac{1}{2}} \left(1 + \frac{m+n}{\nu-2}\right)^{-\frac{\gamma}{1+\gamma}}  \, \mathbb{E}_{x}\Bigg[ 1 + \frac{1}{\nu} \Bigg( \frac{\nu \tr\left( \Sigma_y^{-1} \Sigma_\phi(x) \right)}{\nu+n-2} \notag\\
		&\quad + (\mu(x)-\mu_y)^\top\Sigma_y^{-1}(\mu(x)-\mu_y) + \mathbb{E}_{z}\left[(x-\mu_\theta(z))^\top \Sigma_x^{-1}(x-\mu_\theta(z)) \right] \Bigg)
		\Bigg].
	\end{align*}
	
	\noindent Hence, we can define our divergence as :
	\begin{align*}
		\mathcal{D}_\gamma(q\vert \vert p) &= \frac{C_1}{\gamma} \left( \int \vert \Sigma_\phi(x) \vert^{-\frac{\gamma}{2}} p_{data}(x\vert y)^{1+\gamma} dx \right)^{\frac{1}{1+\gamma}} -\frac{C_2}{\gamma} \mathbb{E}_{x}\Bigg[ 1 + \frac{1}{\nu} \Bigg( \frac{\nu \tr\left( \Sigma_y^{-1} \Sigma_\phi(x) \right)}{\nu+n-2} \\
		&\quad + (\mu(x)-\mu_y)(\mu(x)-\mu_y)^\top + \mathbb{E}_{z}\left[(x-\mu_\theta(z))^\top \Sigma_x^{-1}(x-\mu_\theta(z)) \right] \Bigg)
		\Bigg] \\
		&= \mathbb{E}_{x}\Bigg[\frac{C_1}{\gamma} \vert \Sigma_\phi(x) \vert^{-\frac{\gamma}{2(1+\gamma)}} -\frac{C_2}{\gamma} \Bigg( 1 + \frac{1}{\nu} \Bigg( \frac{\nu}{\nu+n-2} \tr\left( \Sigma_y^{-1} \Sigma_\phi(x) \right) \notag\\
		&\quad + (\mu(x)-\mu_y)^\top\Sigma_y^{-1}(\mu(x)-\mu_y) + \mathbb{E}_{z}\left[(x-\mu_\theta(z))^\top \Sigma_x^{-1}(x-\mu_\theta(z)) \right] \Bigg)\Bigg),
		\Bigg]
	\end{align*}
	
	\noindent with  $C_1$ and $C_2$ being :
	\begin{align*}
		C_1 &= C^{\frac{\gamma}{1+\gamma}}_{\nu+n, m} \left( 1 + \frac{n}{\nu} \right)^{\frac{\gamma m}{2(1+\gamma)}} \left( 1 + \frac{m}{\nu+n-2} \right)^{\frac{1}{1+\gamma}} \\
		C_2 &= \left( C_{\nu, m+n}^{\gamma} \vert \Sigma_x \vert^{-\frac{\gamma}{2}} \vert \Sigma_y \vert^{-\frac{2\gamma +1}{2}} \right)^{\frac{1}{1+\gamma}} \left(1 + \frac{m+n}{\nu-2}\right)^{-\frac{\gamma}{1+\gamma}}.
	\end{align*}
	
	\noindent On that account, the loss function for a class $y$ is :
	\begin{align*}
		\mathcal{L}(\gamma,y) &= -\frac{\nu \gamma}{C_2} \cdot \mathcal{D}_\gamma(q\vert \vert p)\\
		&= \mathbb{E}_{x}\Bigg[ \mathbb{E}_{z}\left[(x-\mu_\theta(z))^\top \Sigma_x^{-1}(x-\mu_\theta(z)) \right] + (\mu(x)-\mu_y)^\top\Sigma_y^{-1}(\mu(x)-\mu_y) \notag\\
		&\quad  + \frac{\nu}{\nu+n-2} \tr\left( \Sigma_y^{-1} \Sigma_\phi(x) \right) -\frac{\nu C_1}{C_2} \vert \Sigma_\phi(x) \vert^{-\frac{\gamma}{2(1+\gamma)}} 
		\Bigg],
	\end{align*}
	
	\noindent and by taking $\Sigma_x = \sigma^2I$ and $\Sigma_y = I$, we obtain :
	\begin{equation*}
		\mathcal{L}_(\gamma,y) = \mathbb{E}_{x}\Bigg[ \frac{\mathbb{E}_{z}\left[\vert \vert x-\mu_\theta(z) \vert \vert^2 \right]}{\sigma^2} + \vert \vert \mu(x)-\mu_y) \vert \vert^2 + \frac{\nu \tr\left(\Sigma_\phi(x) \right)}{\nu+n-2} -\frac{\nu C_1}{C_2} \vert \Sigma_\phi(x) \vert^{-\frac{\gamma}{2(1+\gamma)}} 
		\Bigg].
	\end{equation*}
	
	\section[Variance derivation]{Sampling distribution variance derivation}
	\label{sampling derivation}
	In this section, we present the derivation of $\tau^2$ used in the sampling of $t^3$VAE and C-$t^3$VAE model. We present only the derivation for the C-$t^3$-VAE and it is identical to the one for the $t^3$-VAE since the former model is a generalization of the later.
	
	First, we simplify the divergence $\mathcal{D}(q \Vert p^\star)$ :
	
	\begin{align*}
		\mathcal{D}(q \Vert p^\star) &= -\frac{C_{\nu+n, m}^{\frac{\gamma}{1+\gamma}}}{\gamma}  \left(1+\frac{m}{\nu+n-2}\right)^{-\frac{\gamma}{1+\gamma}} \Bigg[- \left(1+\nu^{-1}n\right)^{\frac{\gamma m}{2(1+\gamma)}} \vert \Sigma_\phi(x)\vert^{-\frac{\gamma}{2(1+\gamma)}} \left(1+\frac{m}{\nu+n-2}\right)  \\
		&\quad +\vert \tau^2 I \vert^{-\frac{\gamma}{2(1+\gamma)}} \times \left(1+\frac{\tr\left(\tau^{-2} \left(1+\nu^{-1}n\right)^{-1} \Sigma_\phi(x)\right)}{\nu+n-2} + \frac{\tau^{-2}}{\nu+n} \Vert \mu(x)-\mu_y) \Vert^2 \right)
		\Bigg]
	\end{align*}	
	Here, we use the fact that $\vert \alpha A \vert^{\delta} = \alpha^{\delta n} \vert A\vert^\delta$ where $n$ is the dimension of the square $A$ matrix. Also, we use $\tr(\alpha A)=\alpha\tr(A)$. After simplification and rearranging we get :
	\begin{align*}
		\mathcal{D}(q \Vert p^\star)&=-\frac{1}{\gamma} C_{\nu+n, m}^{\frac{\gamma}{1+\gamma}} \left(1+\frac{m}{\nu+n-2}\right)^{-\frac{\gamma}{1+\gamma}} \Bigg[- \left(1+\nu^{-1}n\right)^{\frac{\gamma m}{2(1+\gamma)}} \vert \Sigma_\phi(x)\vert^{-\frac{\gamma}{2(1+\gamma)}} \left(1+\frac{m}{\nu+n-2}\right) \\
		&\quad +\tau^{\frac{-\gamma m}{1+\gamma}} \left(1+\frac{\tau^{-2} \left(1+\nu^{-1}n\right)^{-1} }{\nu+n-2} \tr\left(\Sigma_\phi(x)\right) + \frac{\tau^{-2}}{\nu+n} \Vert \mu(x)-\mu_y) \Vert^2 \right)
		\Bigg]\\
		&=-\frac{1}{\gamma} C_{\nu+n, m}^{\frac{\gamma}{1+\gamma}} \left(1+\frac{m}{\nu+n-2}\right)^{-\frac{\gamma}{1+\gamma}} \Bigg[- \left(1+\nu^{-1}n\right)^{\frac{\gamma m}{2(1+\gamma)}} \vert \Sigma_\phi(x)\vert^{-\frac{\gamma}{2(1+\gamma)}} \left(1+\frac{m}{\nu+n-2}\right) \\
		&\quad + \frac{1}{\nu + n} \tau^{-2-\frac{\gamma m}{1+\gamma}} \left(\kappa+\frac{\nu }{\nu+n-2} \tr\left(\Sigma_\phi(x)\right) +  \Vert \mu(x)-\mu_y) \Vert^2 \right)
		\Bigg]\\
		&=-\frac{1}{\gamma} C_{\nu+n, m}^{\frac{\gamma}{1+\gamma}} \left(1+\frac{m}{\nu+n-2}\right)^{-\frac{\gamma}{1+\gamma}}  \frac{1}{\nu + n} \tau^{-2-\frac{\gamma m}{1+\gamma}} \Bigg[-(\nu + n) \tau^{2+\frac{\gamma m}{1+\gamma}}  \left(1+\nu^{-1}n\right)^{\frac{\gamma m}{2(1+\gamma)}} \\
		&\quad \vert \Sigma_\phi(x)\vert^{-\frac{\gamma}{2(1+\gamma)}} \left(1+\frac{m}{\nu+n-2}\right) + \kappa+\frac{\nu }{\nu+n-2} \tr\left(\Sigma_\phi(x)\right) +  \Vert \mu(x)-\mu_y) \Vert^2
		\Bigg],
	\end{align*}
	with:
	$$
	\kappa = \tau^{2}(\nu+n) .
	$$
	Then, we match the result to the loss function in Eq. (\ref{objective function}) to get :
	\begin{equation*}
		\tau^{2+\frac{\gamma m}{1+\gamma}} \left(1+\nu^{-1}n\right)^{\frac{\gamma m}{2(1+\gamma)}+1} \left(1+\frac{m}{\nu+n-2}\right) = \frac{C_1}{C_2}.
	\end{equation*}
	Moreover, we have :
	\begin{align*}
		\frac{C_1}{C_2} &=  C^{\frac{\gamma}{1+\gamma}}_{\nu+n, m} \left( 1 + \frac{n}{\nu} \right)^{\frac{\gamma m}{2(1+\gamma)}} \left( 1 + \frac{m}{\nu+n-2} \right)^{\frac{1}{1+\gamma}} C_{\nu, m+n}^{-\frac{\gamma}{1+\gamma}} \sigma^{\frac{n\gamma}{1+\gamma}} \left(1 + \frac{m+n}{\nu-2}\right)^{\frac{\gamma}{1+\gamma}} \\
		&= \sigma^{\frac{n\gamma}{1+\gamma}} C^{\frac{\gamma}{1+\gamma}}_{\nu+n, m}  C_{\nu, m+n}^{-\frac{\gamma}{1+\gamma}} \left( 1 + \frac{n}{\nu} \right)^{\frac{\gamma m}{2(1+\gamma)}} \left( 1 + \frac{m}{\nu+n-2} \right)^{\frac{1}{1+\gamma}} \left(1 + \frac{m+n}{\nu-2}\right)^{\frac{\gamma}{1+\gamma}} \\
		&= \sigma^{\frac{n\gamma}{1+\gamma}} C^{\frac{-\gamma}{1+\gamma}}_{\nu, n} \left( 1 + \frac{m}{\nu+n-2} \right)^{\frac{1}{1+\gamma}} \left(1 + \frac{m+n}{\nu-2}\right)^{\frac{\gamma}{1+\gamma}}.
	\end{align*}
	Consequently we obtain :
	\begin{equation*}
		\tau^{2+\frac{\gamma m}{1+\gamma}} \left(1+\nu^{-1}n\right)^{\frac{\gamma m}{2(1+\gamma)}+1} \left(1+\frac{m}{\nu+n-2}\right) =\sigma^{\frac{n\gamma}{1+\gamma}} C^{\frac{-\gamma}{1+\gamma}}_{\nu, n} \left( 1 + \frac{m}{\nu+n-2} \right)^{\frac{1}{1+\gamma}} \left(1 + \frac{m+n}{\nu-2}\right)^{\frac{\gamma}{1+\gamma}} 
	\end{equation*}	
	\begin{align*}
		\tau^{2+\frac{\gamma m}{1+\gamma}}&= \sigma^{\frac{n\gamma}{1+\gamma}} C^{\frac{-\gamma}{1+\gamma}}_{\nu, n}  \left(1+\nu^{-1}n\right)^{-\frac{\gamma m}{2(1+\gamma)}-1} \left( 1 + \frac{m}{\nu+n-2} \right)^{-\frac{\gamma}{1+\gamma}} \left(1 + \frac{m+n}{\nu-2}\right)^{\frac{\gamma}{1+\gamma}} \\
		&= \sigma^{\frac{n\gamma}{1+\gamma}} C^{\frac{-\gamma}{1+\gamma}}_{\nu, n}  \left(1+\nu^{-1}n\right)^{-\frac{\gamma m}{2(1+\gamma)}-1} \left(\frac{\nu+n-2}{\nu-2}\right)^{\frac{\gamma}{1+\gamma}} \\
		&= \left(1+\nu^{-1}n\right)^{-\frac{\gamma m}{2(1+\gamma)}-1} \left( \sigma^n C^{-1}_{\nu, n} \quad \frac{\nu+n-2}{\nu-2}\right)^{\frac{\gamma}{1+\gamma}} .
	\end{align*}
	Hence, we get :
	\begin{equation*}
		\tau^2 = \left(1+\nu^{-1}n\right)^{-1} \left( \sigma^n C^{-1}_{\nu, n} \quad \frac{\nu+n-2}{\nu-2}\right)^{-\frac{2}{\nu+n-2}}.
	\end{equation*}
	which is the form of $\tau^2$ we report in Eq. (\ref{kim tau}).
	
	\section[Experimental setup]{Experimental setup}
	\label{exp_setup_mt3vae}
	\subsection{Datasets}
	\label{datasets-mt3vae}
	We conduct experiments on three datasets notably SVHN-LT \citep{svhn}, \mbox{CIFAR100-LT}~\citep{cifar100, cifar100lt} and CelebA \citep{celeba} each chosen to highlight different challenges related to generative modeling under class imbalance and varying visual complexity.
	\begin{itemize}
		\item \textbf{SVHN-LT :} The Street View House Numbers (SVHN) dataset \citep{svhn} is composed of real-world digit images collected from Google Street View. It contains more than $600{,}000$ labeled digits ($0$–$9$) with size $32 \times 32$, complex backgrounds, and diverse illumination conditions. The digits in SVHN are not centered or uniformly scaled, which makes the dataset considerably more challenging. Figure~\ref{fig:svhn_samples} shows sample images of this dataset.
		
		\begin{figure}[H]
			\centering
			\begin{minipage}{0.1\textwidth}
				\includegraphics[width=\columnwidth]{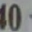}
			\end{minipage}
			\begin{minipage}{0.1\textwidth}
				\includegraphics[width=\columnwidth]{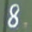}
			\end{minipage}
			\begin{minipage}{0.1\textwidth}
				\includegraphics[width=\columnwidth]{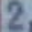}
			\end{minipage}
			\begin{minipage}{0.1\textwidth}
				\includegraphics[width=\columnwidth]{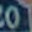}
			\end{minipage}
			\begin{minipage}{0.1\textwidth}
				\includegraphics[width=\columnwidth]{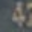}
			\end{minipage}
			\begin{minipage}{0.1\textwidth}
				\includegraphics[width=\columnwidth]{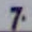}
			\end{minipage}
			\begin{minipage}{0.1\textwidth}
				\includegraphics[width=\columnwidth]{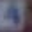}
			\end{minipage}
			\begin{minipage}{0.1\textwidth}
				\includegraphics[width=\columnwidth]{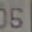}
			\end{minipage}
			\caption{Sample images from the SVHN dataset \citep{svhn}.}
			\label{fig:svhn_samples}
			\vspace{-.5cm}
		\end{figure}
		
		However, as this dataset is naturally imbalanced, we balance the number of images across classes to have full control over the imposed imbalance ratio. In Table \ref{svhn imabalnced} we provide the number of images present in the dataset before balancing.
		\begin{table}[H]
			\centering
				\begin{tabular}{c|cccccccccc}
					\toprule
					Class & 0 & 1 & 2 & 3 & 4 & 5 & 6 & 7 & 8 & 9 \\\midrule
					Train set & 4948 & 13861 & 10585 & 8497 & 7458 & 6882 & 5727 & 5595 & 5045 & \textbf{4656} \\
					Test set & 1744 & \phantom{0}5099 & \phantom{0}4149 & 2882 & 2523 & 2384 & 1977 & 2019 & 1660 & \textbf{1595} \\
					\bottomrule
				\end{tabular}
			\caption{The Number of images in the SVHN dataset for the train and test sets before balancing. The value in bold is the one used to balance the dataset.}
			\label{svhn imabalnced}
			\vspace{-.4cm}
		\end{table}
		For both training and testing, we crop each class to the minimum number of samples available across all classes. The only data augmentation applied is a random horizontal flip with 50\% probability.
		
		\item \textbf{CIFAR100-LT :} The CIFAR100 dataset \citep{cifar100} consists of $60{,}000$ color images of size $32 \times 32$ pixels, evenly distributed across $100$ object categories. Each category contains $600$ images, split into $500$ training samples and $100$ test samples. Figure~\ref{fig:cifar_samples} shows sample images after preprocessing.
		
		\begin{figure}[H]
			\centering
			\begin{minipage}{0.1\textwidth}
				\includegraphics[width=\columnwidth]{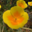}
			\end{minipage}
			\begin{minipage}{0.1\textwidth}
				\includegraphics[width=\columnwidth]{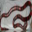}
			\end{minipage}
			\begin{minipage}{0.1\textwidth}
				\includegraphics[width=\columnwidth]{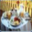}
			\end{minipage}
			\begin{minipage}{0.1\textwidth}
				\includegraphics[width=\columnwidth]{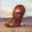}
			\end{minipage}
			\begin{minipage}{0.1\textwidth}
				\includegraphics[width=\columnwidth]{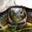}
			\end{minipage}
			\begin{minipage}{0.1\textwidth}
				\includegraphics[width=\columnwidth]{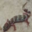}
			\end{minipage}
			\begin{minipage}{0.1\textwidth}
				\includegraphics[width=\columnwidth]{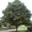}
			\end{minipage}
			\begin{minipage}{0.1\textwidth}
				\includegraphics[width=\columnwidth]{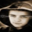}
			\end{minipage}
			\caption{Sample images from the CIFAR100 dataset \citep{cifar100}.}
			\label{fig:cifar_samples}
			\vspace{-.3cm}
		\end{figure}
		We use the dataset in its entirety without class filtering. As with SVHN-LT, we apply a random horizontal flip with 50\% probability for data augmentation.
		
		\item \textbf{CelebA :} The CelebFaces Attributes Dataset (CelebA) \citep{celeba} contains 202,599 color images of celebrity faces at a resolution of $178 \times 218$. Each image is annotated with 40 binary facial attributes, such as Mustache, Smiling, and Young. This dataset exhibits significant variability in pose, expression, and illumination while providing high resolution images. We preprocess the images of this dataset by cropping the central region to $160 \times 160$ and resizing to $128 \times 128$ using bilinear interpolation. Figure \ref{fig:celeba_samples} illustrates sample images after preprocessing. Also, we adhere to the original training, validation, and test splits provided by the dataset.
		\begin{figure}[ht]
			\centering
			\begin{minipage}{0.1\textwidth}
				\includegraphics[width=\columnwidth]{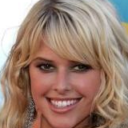}
			\end{minipage}
			\begin{minipage}{0.1\textwidth}
				\includegraphics[width=\columnwidth]{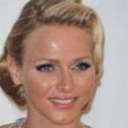}
			\end{minipage}
			\begin{minipage}{0.1\textwidth}
				\includegraphics[width=\columnwidth]{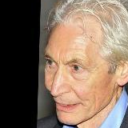}
			\end{minipage}
			\begin{minipage}{0.1\textwidth}
				\includegraphics[width=\columnwidth]{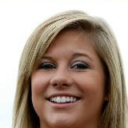}
			\end{minipage}
			\begin{minipage}{0.1\textwidth}
				\includegraphics[width=\columnwidth]{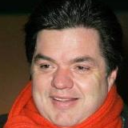}
			\end{minipage}
			\begin{minipage}{0.1\textwidth}
				\includegraphics[width=\columnwidth]{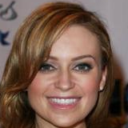}
			\end{minipage}
			\begin{minipage}{0.1\textwidth}
				\includegraphics[width=\columnwidth]{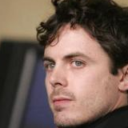}
			\end{minipage}
			\begin{minipage}{0.1\textwidth}
				\includegraphics[width=\columnwidth]{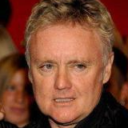}
			\end{minipage}
			\caption{Sample images from the CelebA dataset after preprocessing \citep{celeba}.}
			\label{fig:celeba_samples}
		\end{figure}

		\begin{table}[ht]
			\centering
			\resizebox{0.9\columnwidth}{!}{%
				\begin{tabular}{c|cccccccc}
					\toprule
					& Mustache & No Mustache & Young & Old & Male & Female & Smiling & Not Smiling \\\midrule
					Train set & 6642 & 156128 & 126788 & 35982 & 68261 & 94509 & 78080 & 84690 \\
					Test set & \phantom{0}722 & \phantom{0}19190 & \phantom{0}15114 & \phantom{0}4848 & \phantom{0}7715 & 12247 & \phantom{0}9987 & \phantom{0}9975 \\
					\bottomrule
				\end{tabular}
			}
			\caption{Number of images in the CelebA dataset for the train and test sets for the Mustache, Young, Male and Smiling attributes.}
			\label{celeba imabalnced}
		\end{table}
		
		Given CelebA’s multi-attribute structure, we select four binary attributes—Mustache, Young, Male, and Smiling—for training and evaluation, treating each attribute and its negation as distinct classes. These attributes were chosen due to their varying imbalance ratios, defined as the frequency of the majority class divided by the minority. The resulting ratios are 25 for Mustache, 3.5 for Young, 1.4 for Male, and 1 for Smiling. This enables the study of imbalance effects in generative modeling. We also balance the test sets by downsampling the larger class for each attribute. In Table \ref{celeba imabalnced} we report the number of images per selected attribute in the dataset.
	\end{itemize}
	
	\subsection{Model Architecture}
	Our encoder-decoder models follow a modular block design. Each encoder block consists of a convolutional layer, followed by 2D batch normalization and ReLU activation. Decoder blocks mirror this structure but replace convolutional layers with transposed convolutions.
	
	\begin{itemize}
		\item \textbf{SVHN-LT and CIFAR100-LT :}	Encoders consist of four convolutional blocks with channels \{64, 128, 256, 512\}, followed by two linear layers for estimating mean and covariance. The decoder uses three transposed convolutional blocks with channel sizes \{128, 64, 32\}, ending with a three-channel convolution and Sigmoid activation.
		\item \textbf{CelebA :} The CelebA encoder includes six convolutional blocks with channels \{64, 128, 256, 512, 512, 512\}, ending with two linear layers. The decoder has six transposed convolutional layers with channels \{512, 512, 256, 128, 64, 32\}, followed by a final convolutional layer and Sigmoid activation.
		
	\end{itemize}
	
	\subsection{Training Details}
	All models are trained using the AdamW optimizer with a learning rate of $10^{-3}$ for 150 epochs. We use a batch size of 64 for SVHN-LT and CIFAR100-LT, and 128 for CelebA.
	
	\section{Hyperparameter Tuning}
	\label{hyper param tunning}
	We present the hyperparameter tuning process used across all evaluated models. We first optimize $\beta$, then~$\nu$, and finally $\tau$, yielding the models' results reported in Table~\ref{quantitaive svhn cifar}.
	
	\subsection{$\beta$ Optimization}
	We perform a hyperparameter study over $\beta$ for all tested models. Unless otherwise noted, we use the theoretically derived $\tau^2$ and set $\nu = 10$.
	
	\subsubsection{On the SVHN-LT dataset}
\begin{figure}[H]
		\centering
		\begin{subfigure}[b]{0.44\columnwidth}
			\centering
			\includegraphics[width=\linewidth]{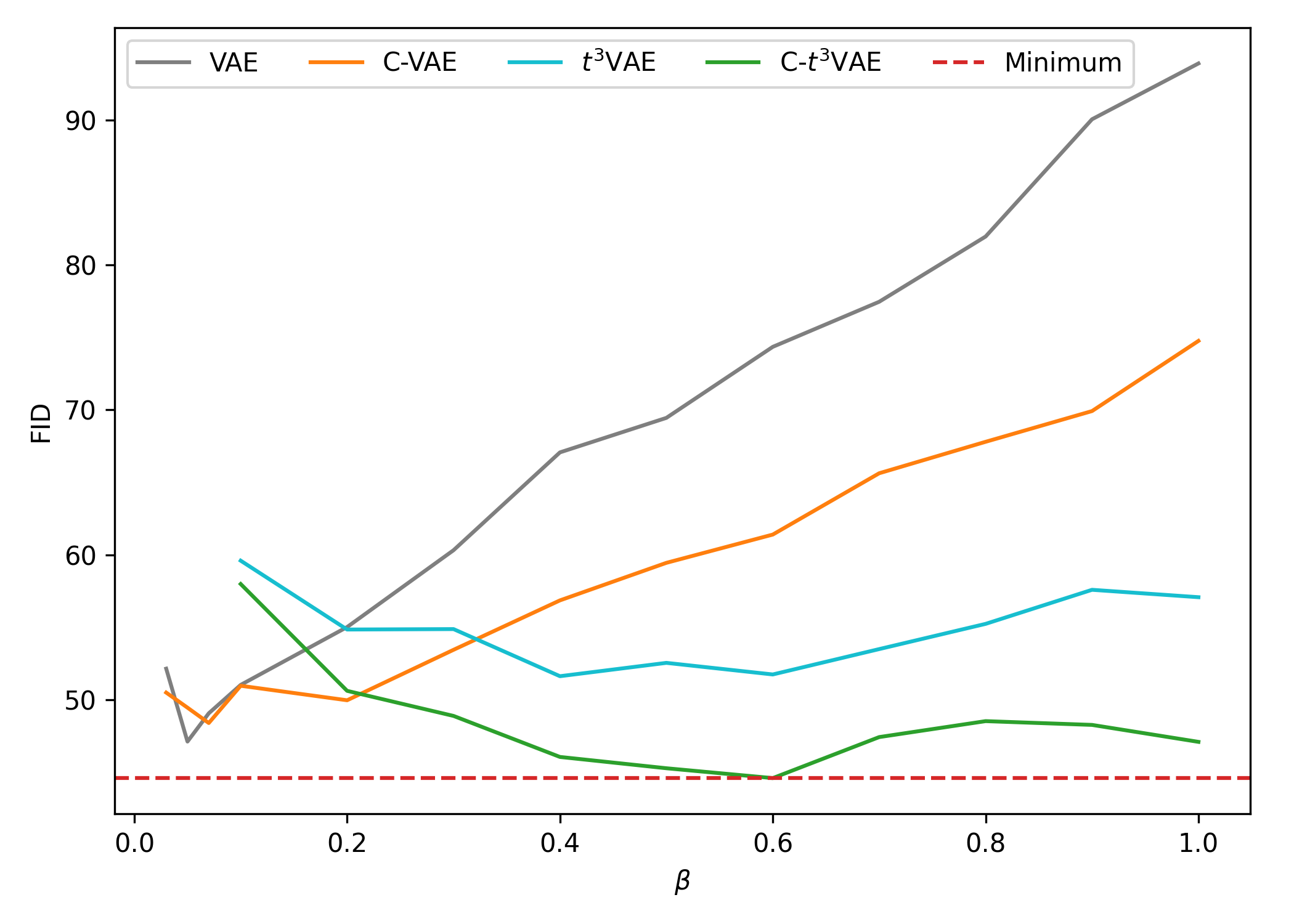}
			\caption{SVHN-LT $\rho$=100}
		\end{subfigure}
		\quad 
		\begin{subfigure}[b]{0.44\columnwidth}
			\centering
			\includegraphics[width=\linewidth]{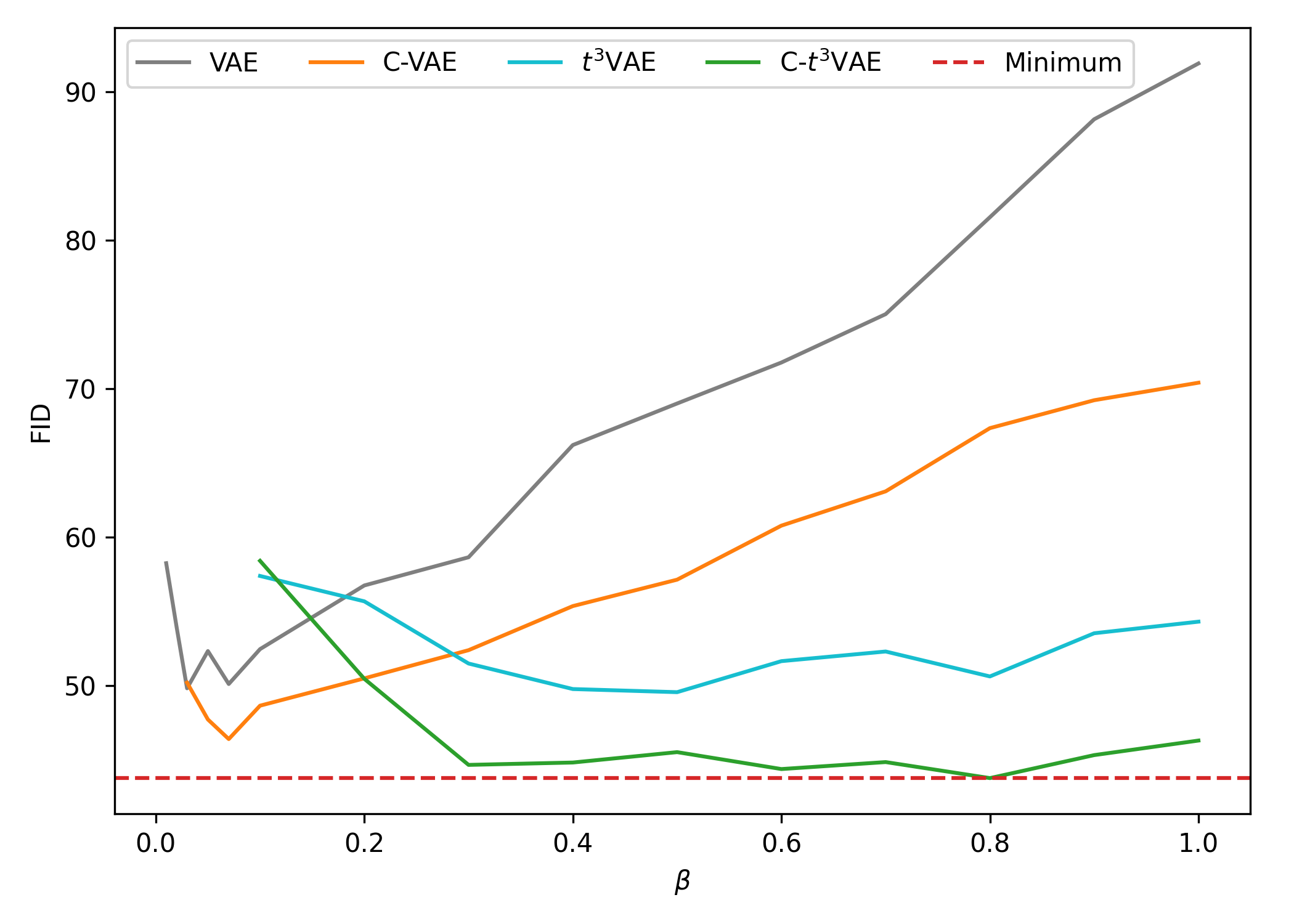}
			\caption{SVHN-LT $\rho$=50}
		\end{subfigure}
		%
		\centering
		\begin{subfigure}[b]{0.44\columnwidth}
			\centering
			\includegraphics[width=\linewidth]{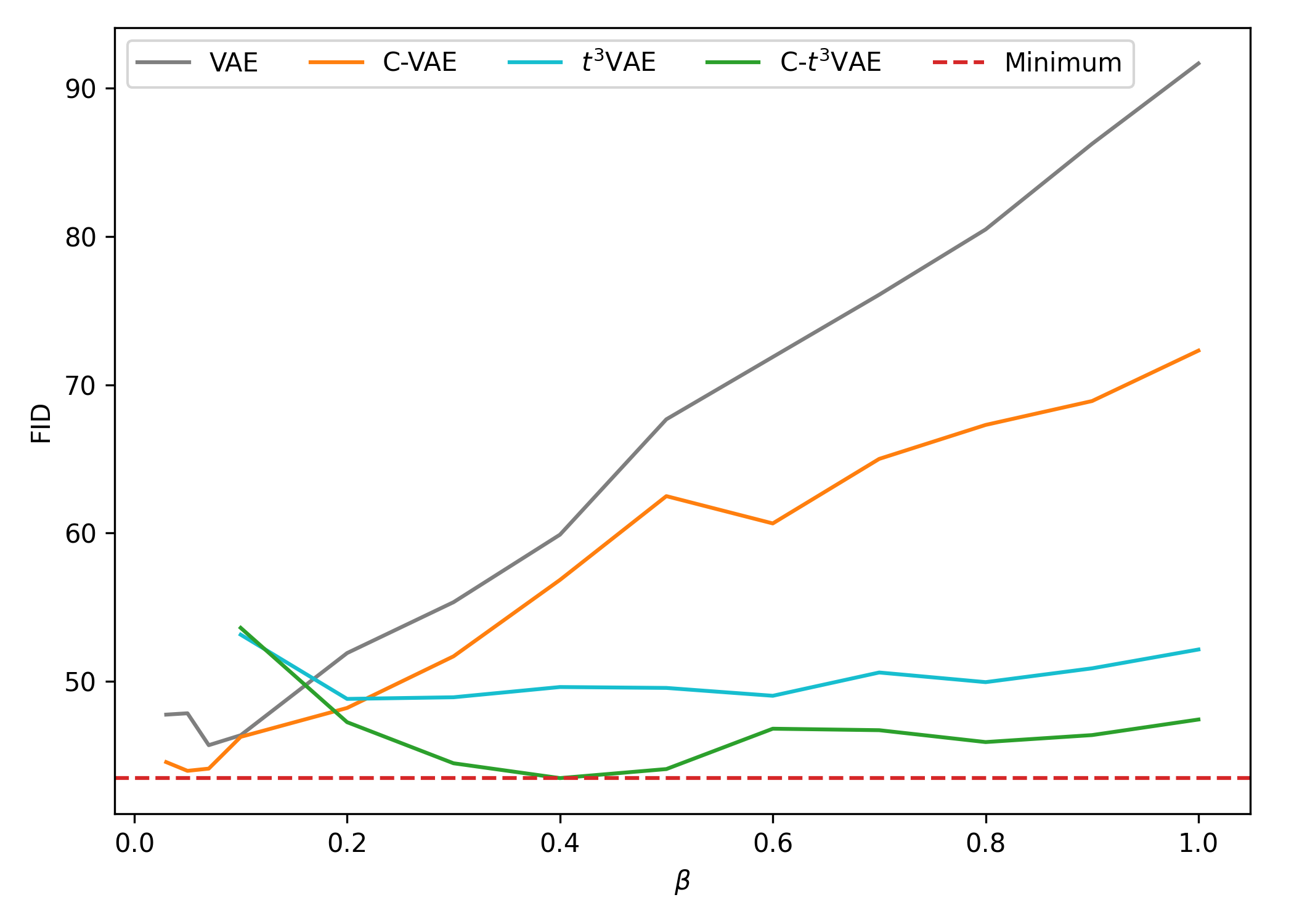}
			\caption{SVHN-LT $\rho$=10}
		\end{subfigure}
		\quad 
		\begin{subfigure}[b]{0.44\columnwidth}
			\centering
			\includegraphics[width=\linewidth]{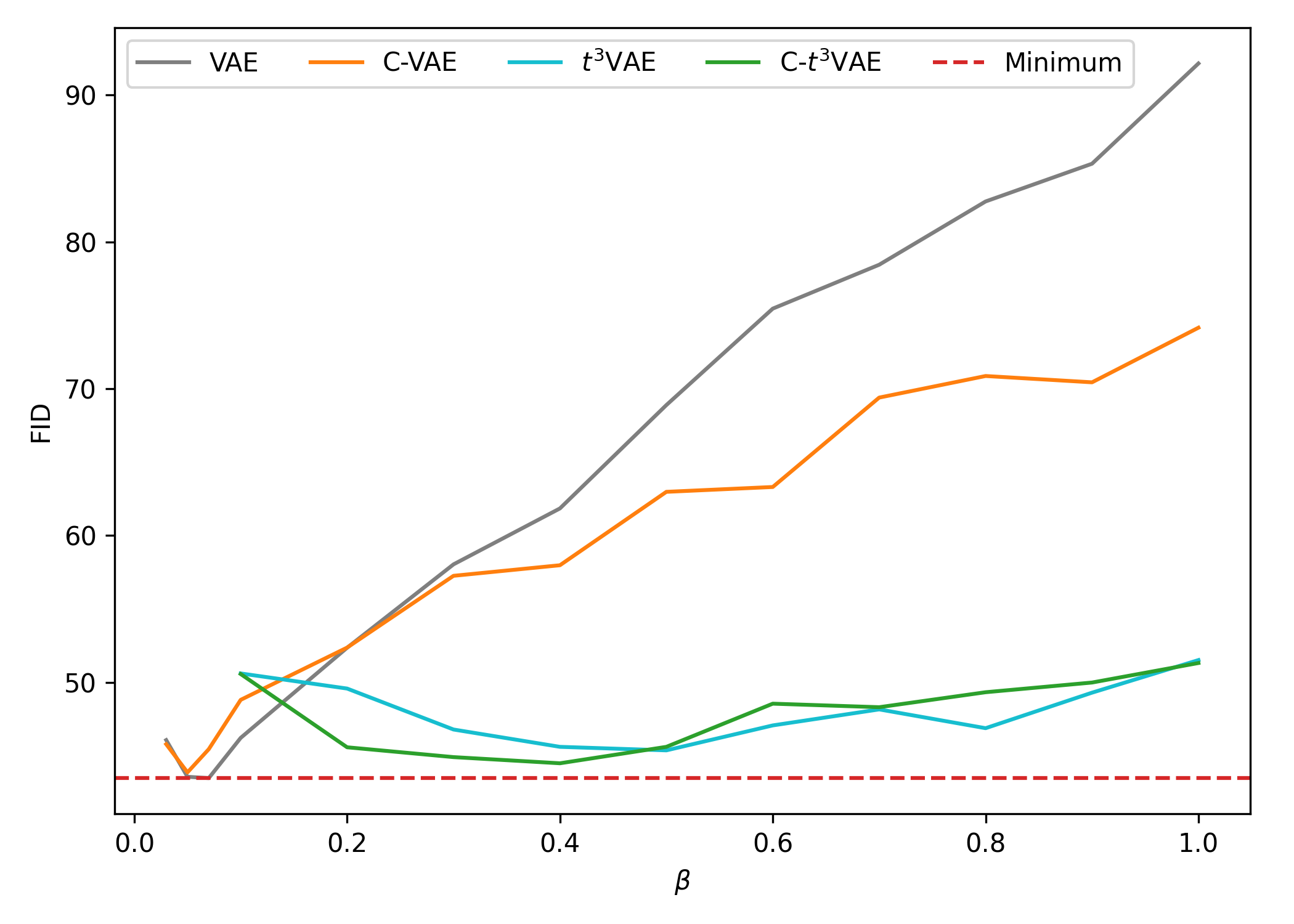}
			\caption{SVHN-LT $\rho$=1}
		\end{subfigure}
		
		\caption{Variability of the FID as a function of the $\beta$ hyperparameter for the VAE, C-VAE, $t^3$VAE and C-$t^3$VAE on the SVHN-LT dataset.}
		\label{beta study svhn}
	\end{figure}

	As shown in Figure \ref{beta study svhn}, the optimal $\beta$ values for Student's t models lies in the range $\beta \in [0.4, 0.6]$ whereas it lies in the $\beta \in [0.05, 0.07]$ range for Gaussian-based models. This is because the regularization term in the $\gamma$-power divergence loss is ten times larger than the KL divergence. Figure \ref{beta study svhn} also shows that FID performance is highly sensitive to $\beta$ in the Gaussian setting, requiring careful tuning which is not the case for Student's t based models. Finally, C-$t^3$VAE achieves the best FID surpassing the $t^3$VAE and the C-VAE for all imbalance settings.
	
	\subsubsection{On the CIFAR100-LT dataset} 
\begin{figure}[H]
		\centering
		\begin{subfigure}[b]{0.44\columnwidth}
			\centering
			\includegraphics[width=\linewidth]{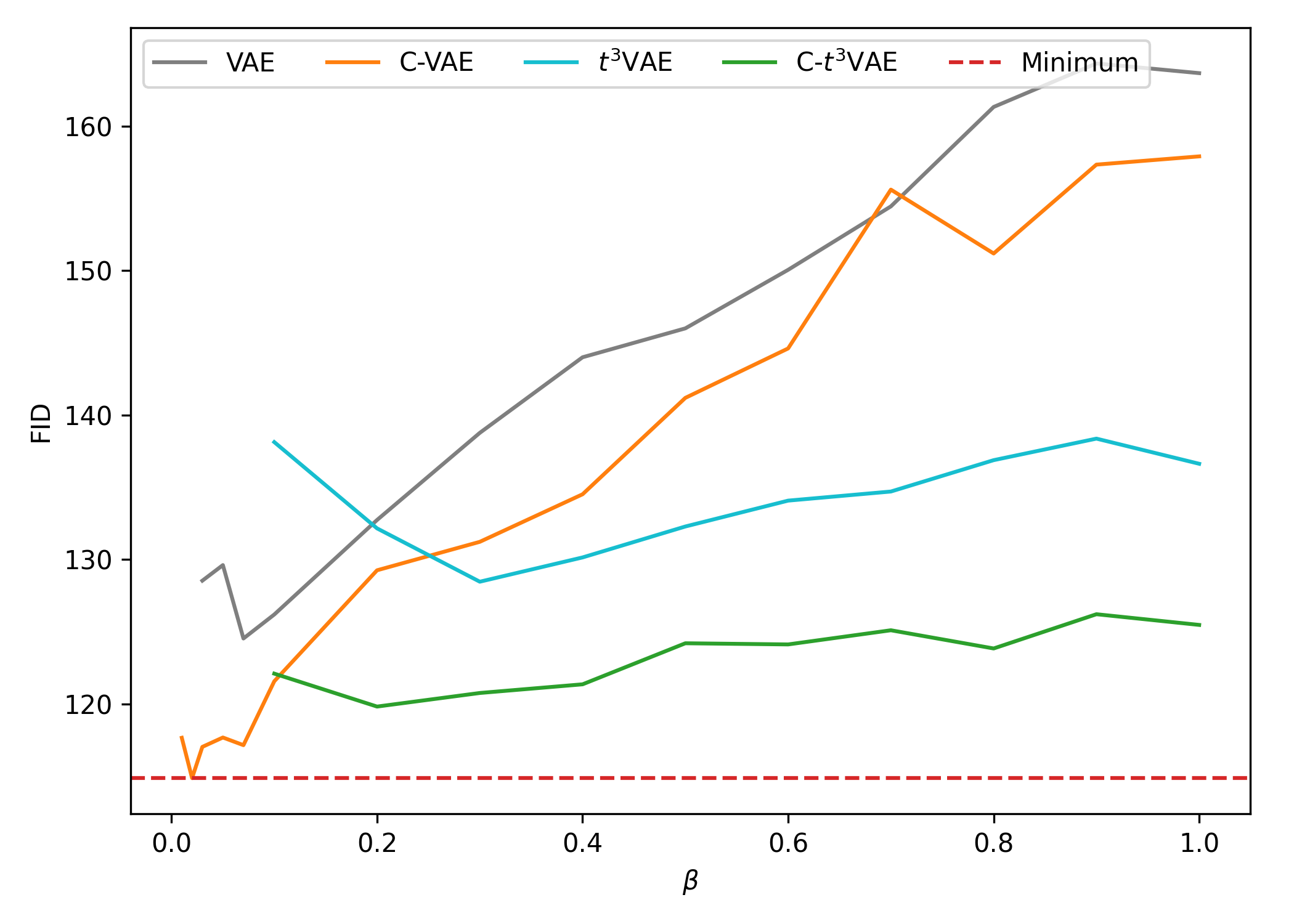}
			\caption{CIFAR100-LT $\rho$=100}
		\end{subfigure}
		\hfill 
		\begin{subfigure}[b]{0.44\columnwidth}
			\centering
			\includegraphics[width=\linewidth]{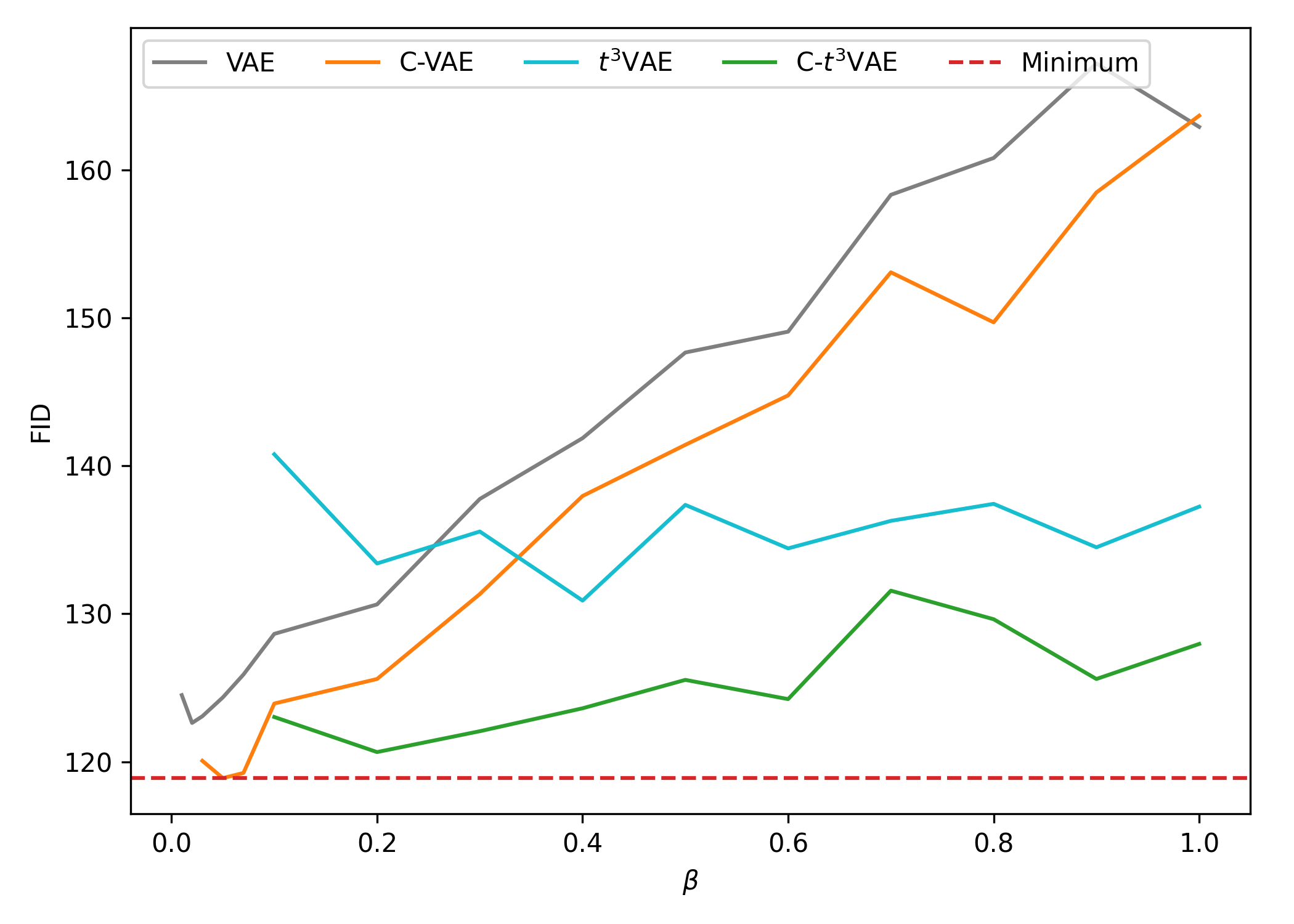}
			\caption{CIFAR100-LT $\rho$=50}
		\end{subfigure}
		
		\begin{subfigure}[b]{0.44\columnwidth}
			\centering
			\includegraphics[width=\linewidth]{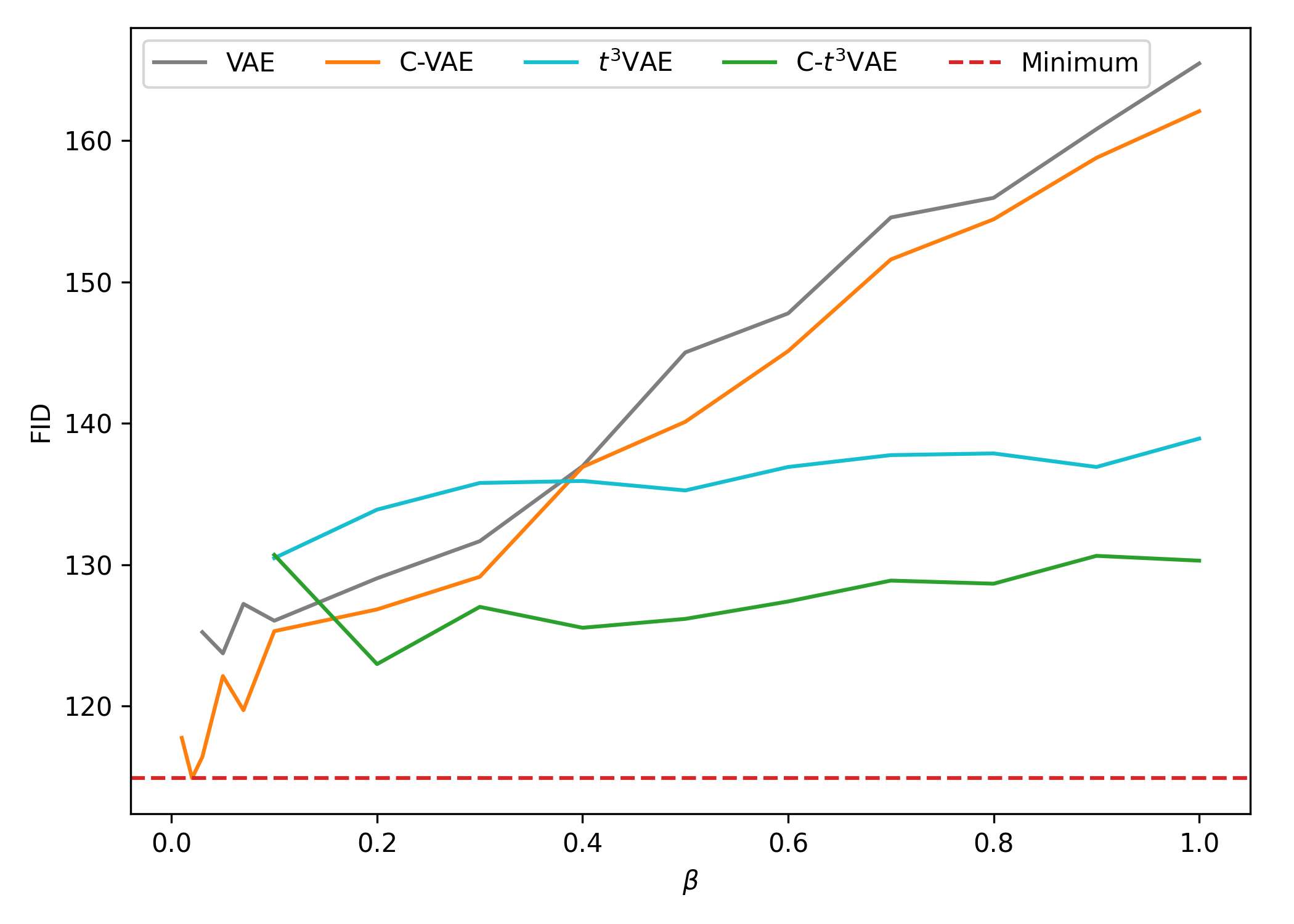}
			\caption{CIFAR100-LT $\rho$=10}
		\end{subfigure}
		\hfill
		\begin{subfigure}[b]{0.44\columnwidth}
			\centering
			\includegraphics[width=\linewidth]{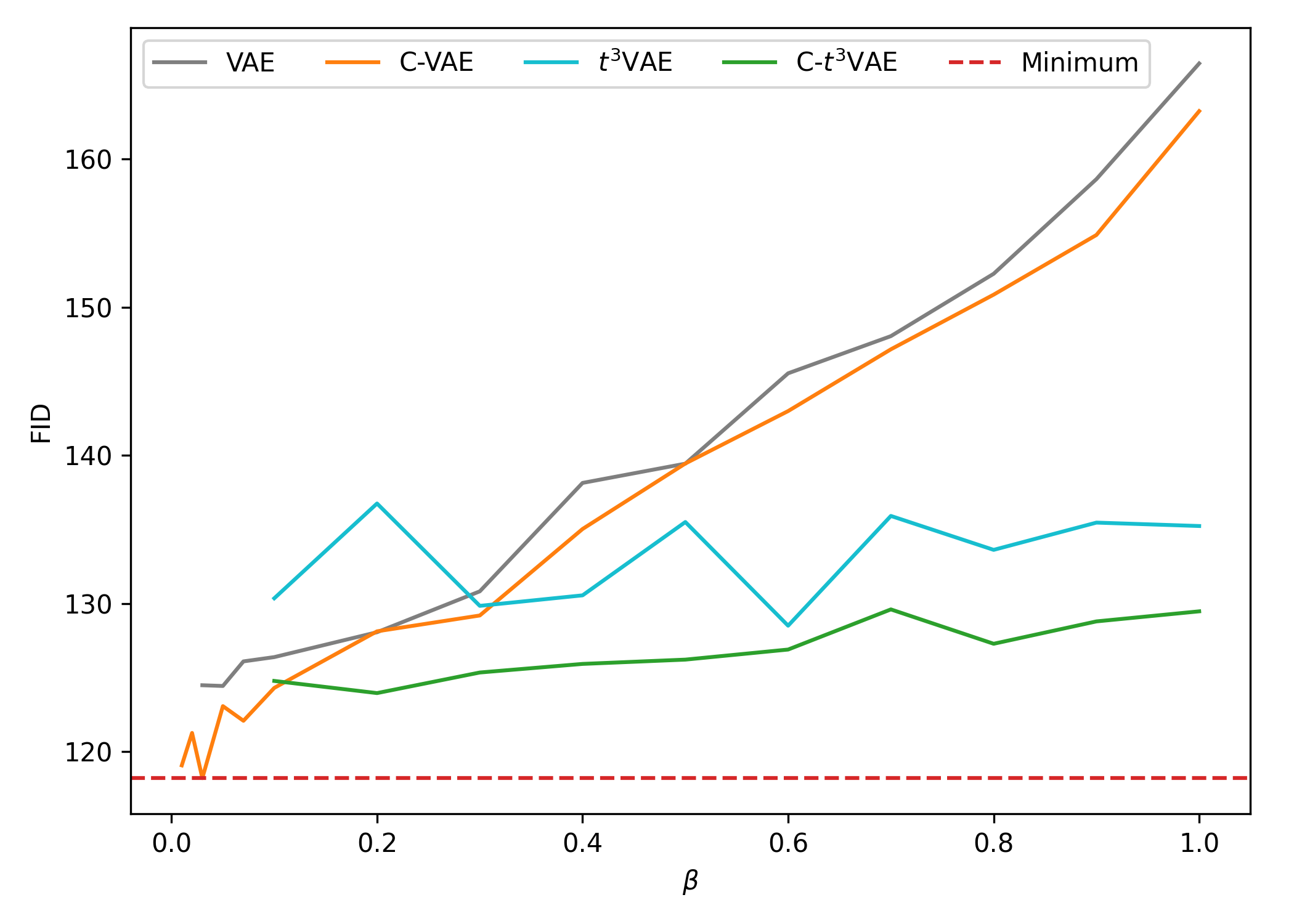}
			\caption{CIFAR100-LT $\rho$=1}
		\end{subfigure}
		
		\caption{Variability of the FID as a function of the $\beta$ hyperparameter for the VAE, C-VAE, $t^3$VAE and C-$t^3$VAE on the CIFAR100-LT dataset.}
		\label{beta study cifar}
	\end{figure}

	From Figure \ref{beta study cifar}, we observe that Student's t models obtain the best performance in terms of FID at $\beta=0.2$ for CIFAR100-LT dataset. However, for the Gaussian-based models, the optimal value is much lower with $\beta \in [0.02, 0.05]$. The reason for this is that on this dataset too the KL regularization term is ten times smaller than the regularization terms present in the $\gamma$-power divergence loss. Additionally, we notice that C-VAE performs slightly better, likely due to the complexity of the dataset preventing full convergence to the imposed latent distribution. We further investigate this hypothesis in the $\tau$ analysis below. 
	
	\subsubsection{On the CelebA dataset} 
\begin{figure}[H]
		\centering
		\begin{subfigure}[b]{0.44\columnwidth}
			\centering
			\includegraphics[width=\linewidth]{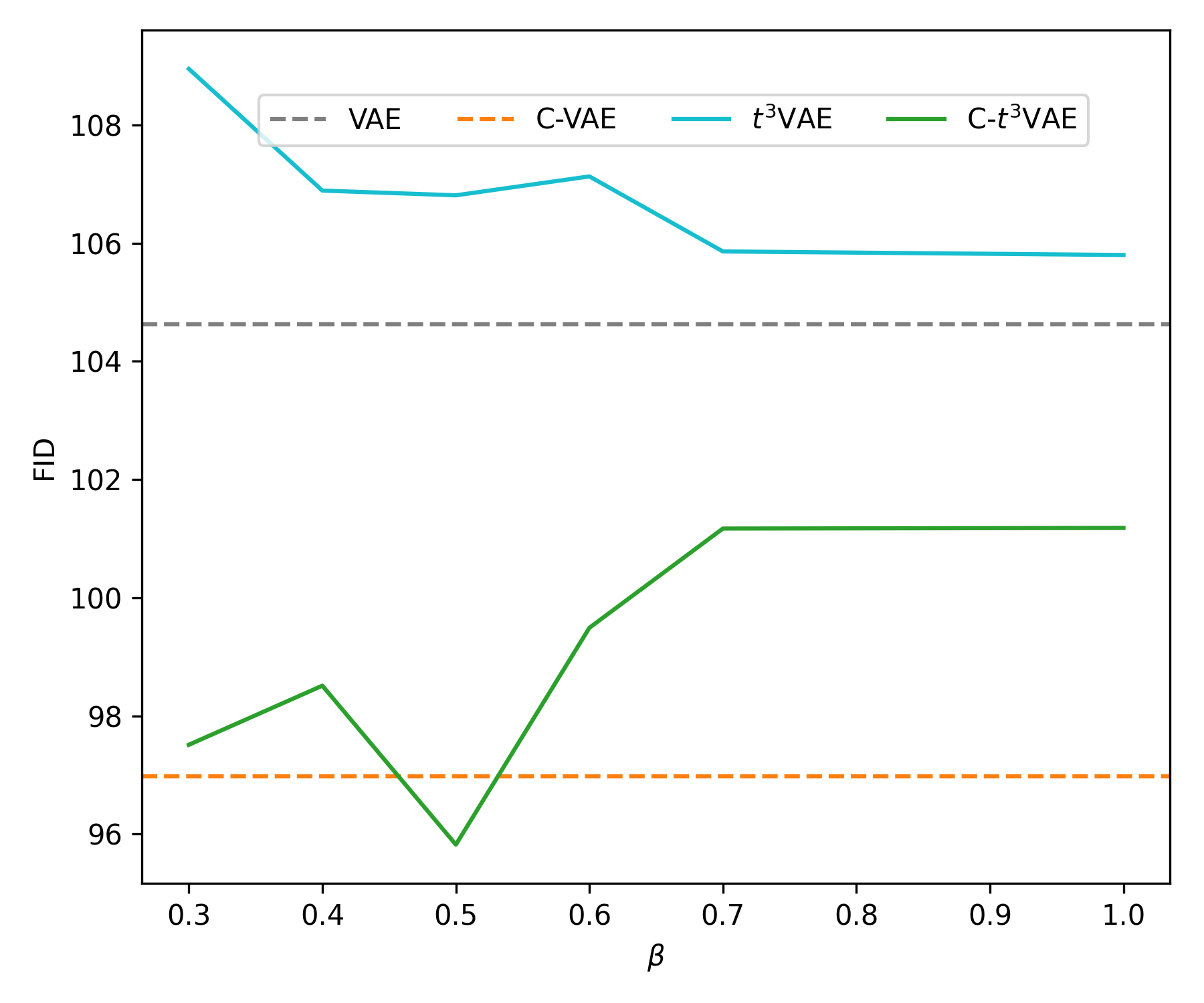}
			\caption{CelebA - Mustache}
		\end{subfigure}
		\hfill
		\begin{subfigure}[b]{0.44\columnwidth}
			\centering
			\includegraphics[width=\linewidth]{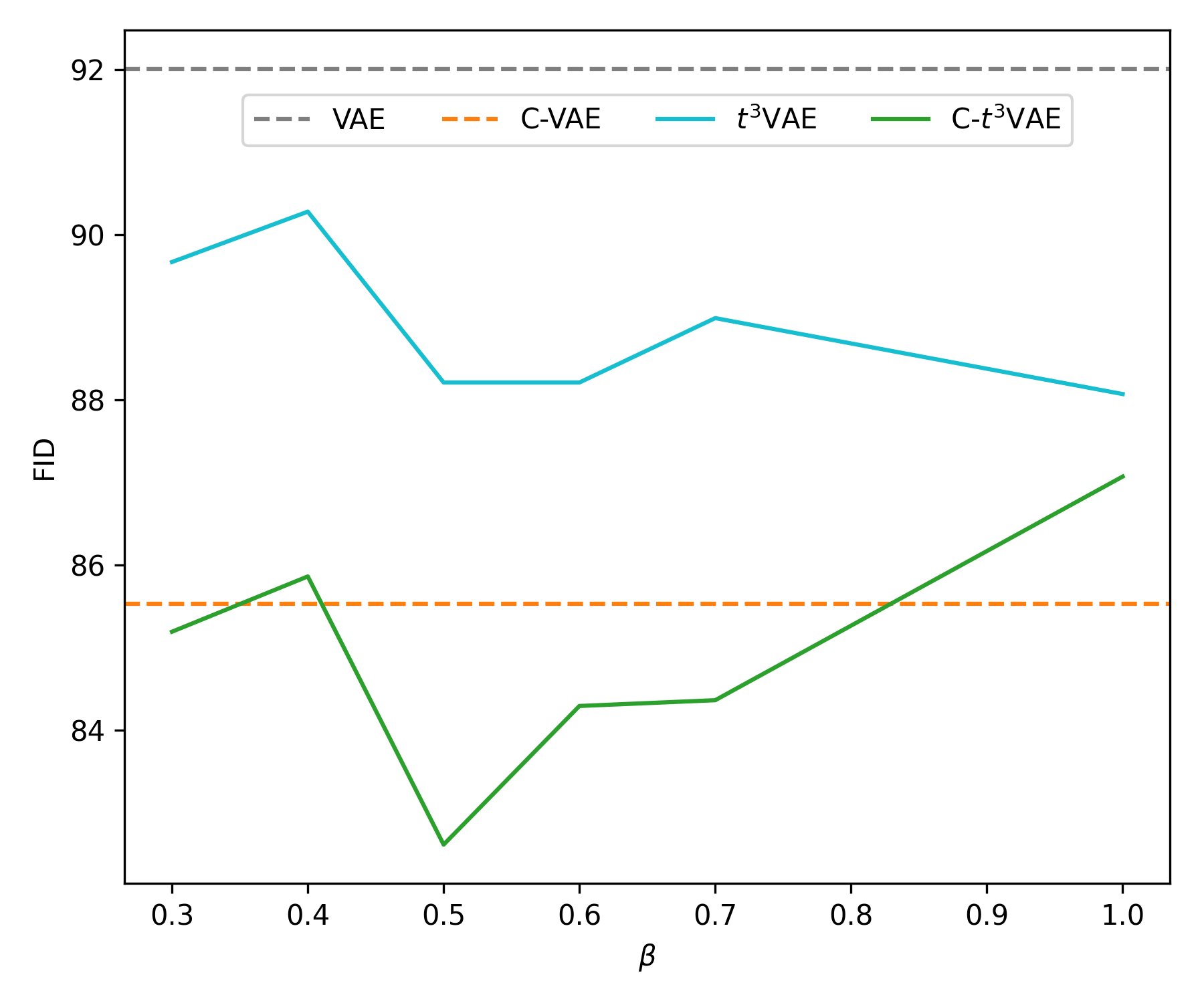}
			\caption{CelebA - Young}
		\end{subfigure}
		
		\begin{subfigure}[b]{0.44\columnwidth}
			\centering
			\includegraphics[width=\linewidth]{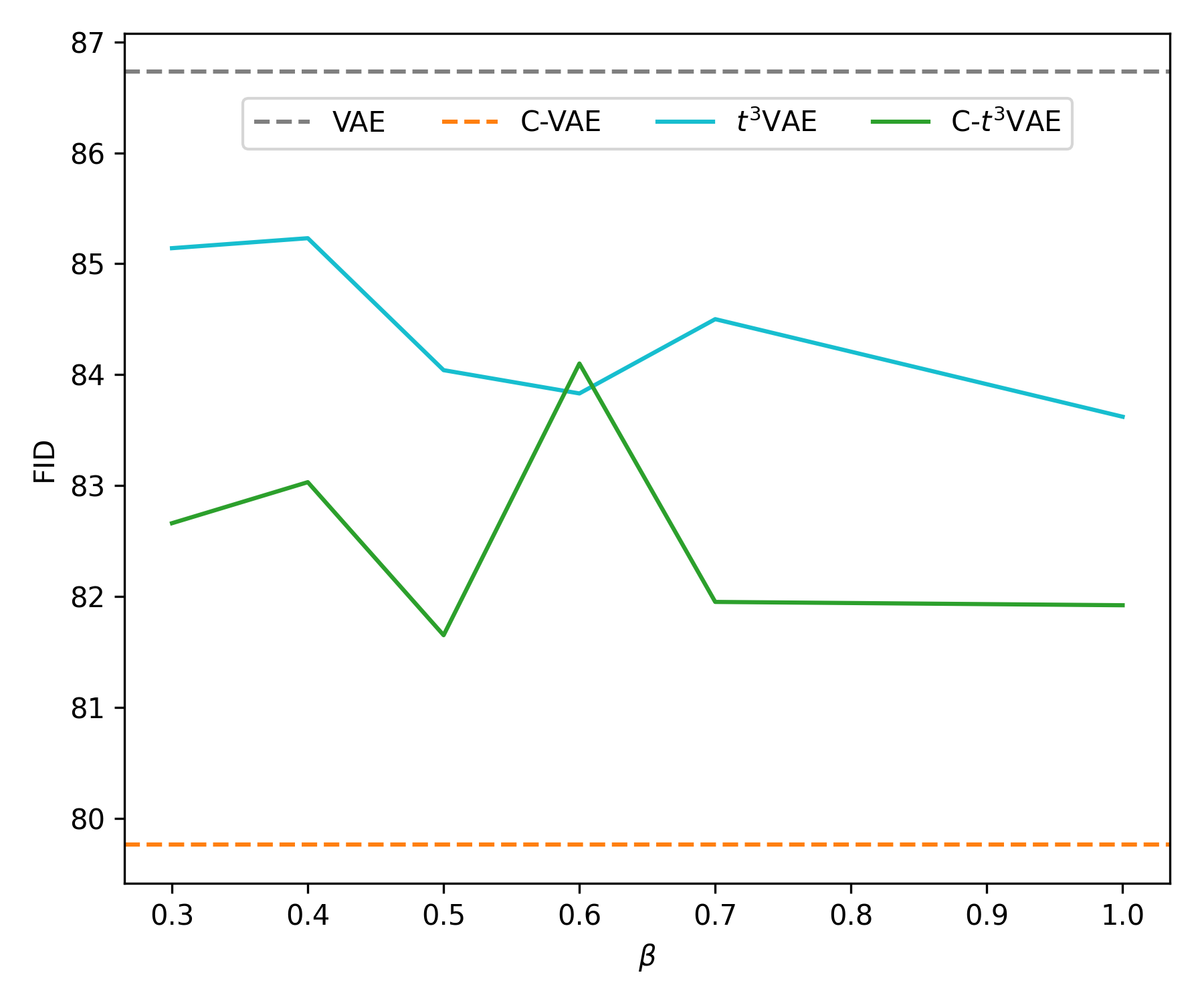}
			\caption{CelebA - Male}
		\end{subfigure}
		\hfill
		\begin{subfigure}[b]{0.44\columnwidth}
			\centering
			\includegraphics[width=\linewidth]{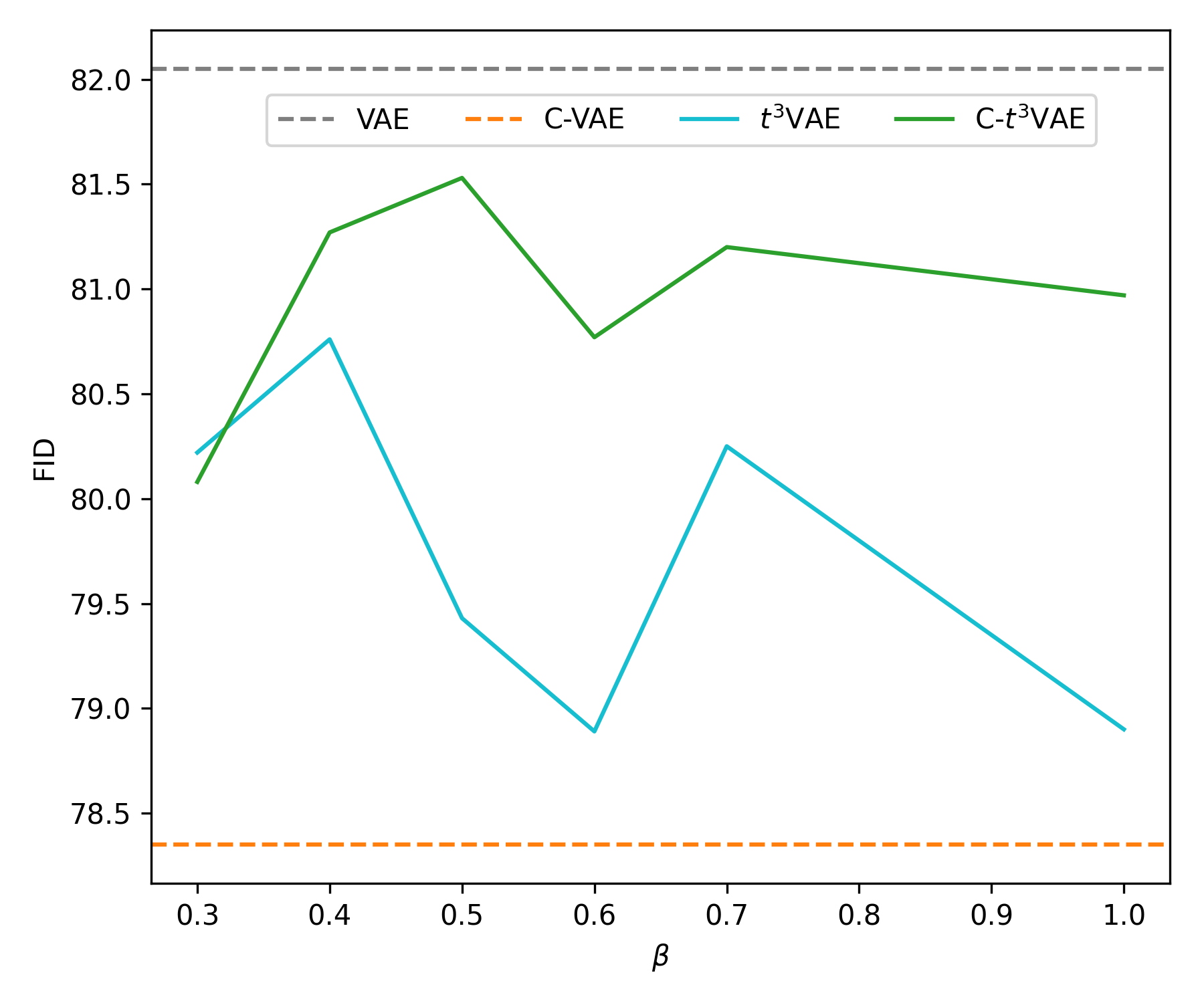}
			\caption{CelebA - Smiling}
		\end{subfigure}
		
		\caption{Variability of the FID as a function of the $\beta$ hyperparameter for the $t^3$VAE and C-$t^3$VAE on the CelebA dataset. The horizontal lines for the VAE and C-VAE models are for the best performing model between $\beta=0.1$ and $\beta=1$.}
		\label{beta study celeba}
		\vspace{-.3cm}
	\end{figure}

	For CelebA, we optimize $\beta$ exclusively for Student's $t$ models, setting $\beta = 0.1$ for Gaussian variants. Table \ref{quantitaive svhn cifar} indicates that $\beta$ has minimal impact on CelebA's FID, unlike the trends seen in SVHN-LT and CIFAR100-LT. Therefore, extensive tuning for Gaussian models is omitted. Figure~\ref{beta study celeba} suggests that Student's t models share this robustness to $\beta$ variations, likely driven by the dataset's limited intra-class variability.

	\subsection{$\nu$ Optimization}
	Table~\ref{nu study} presents a sensitivity analysis of the degrees of freedom parameter $\nu$ for the C-$t^3$VAE on \mbox{SVHN-LT} and CIFAR100-LT, using the optimal $\beta$ from the previous study. Consistent with prior work~\citep{t3vae}, we find that $\nu$ = 10 yields robust performance on average, though slight gains can be achieved by fine-tuning within the range $[2.5, 20]$. Ultimately, however, the generative FID remains relatively insensitive to variations in $\nu$, corroborating the findings of \citep{t3vae} regarding reconstruction quality.
	\begin{table}[H]
		\centering
		\resizebox{0.65\columnwidth}{!}{%
			\begin{tabular}{l|cccc||cccc}
				\toprule
				& \multicolumn{4}{c||}{SVHN-LT} & \multicolumn{4}{c}{CIFAR100-LT}\\
				$\nu$ & 100 & 50 & 10 & 1  & 100 & 50 & 10 & 1  \\\midrule
				2.1 & 45.50 & 44.51 & 42.96 & 46.23 & 121.28 & 122.03 & 121.93 & 123.41  \\
				2.5 & 45.76 & 43.96 & 45.81 & 45.40 & 119.15 & 120.19 & \textbf{120.10} & 124.83  \\
				5  & 44.89 & \textbf{42.60} & 45.03  & 46.33 & 120.52 & 123.21 & 124.29 & 123.71  \\
				10 & 44.59 & 44.37 & 43.48 & \textbf{44.49} & \textbf{119.83} & 120.65 & 122.96 & \textbf{123.95}  \\
				20 & \textbf{44.02} & 43.89 & \textbf{42.01} & 44.75 & 121.48 & \textbf{118.41} & 124.58 & 126.13  \\
				50 & 48.03 & 46.39 & 43.59 & 45.57 & \textbf{119.58} & 126.36 & 124.38 & 127.48  \\
				100 & 45.97 & 44.63 & 43.74 & 47.52 & 123.26 & 122.90 & 127.42 & 125.67  \\\bottomrule
			\end{tabular}
		}
		\caption{Variability of the FID as a function of the standard deviation $\nu$ for the \mbox{C-$t^3$VAE} model.}
		\label{nu study}
		\vspace{-.5cm}
	\end{table}
	
	\subsection{$\tau$ Optimization}
	\label{tau tunning}
\begin{figure}[H]
		\centering
		\begin{subfigure}[b]{0.28\columnwidth}
			\centering
			\includegraphics[width=\linewidth]{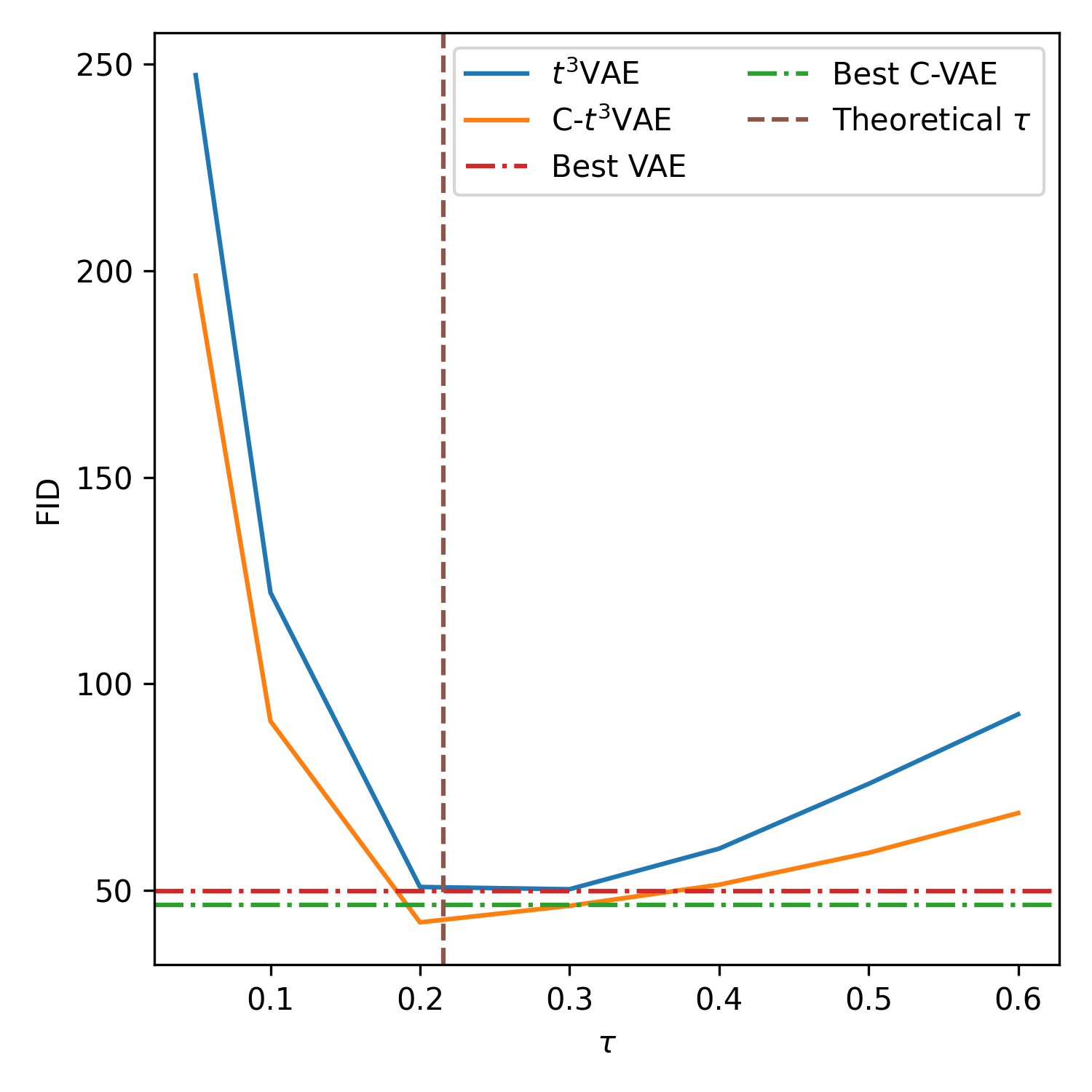}
			\caption{SVHN-LT $\rho$=50}
		\end{subfigure}
		\hfill
		\begin{subfigure}[b]{0.28\columnwidth}
			\centering
			\includegraphics[width=\linewidth]{figures/svhn/tau_study_fid_0_1}
			\caption{SVHN-LT $\rho$=10}
		\end{subfigure}
		\hfill
		\begin{subfigure}[b]{0.28\columnwidth}
			\centering
			\includegraphics[width=\linewidth]{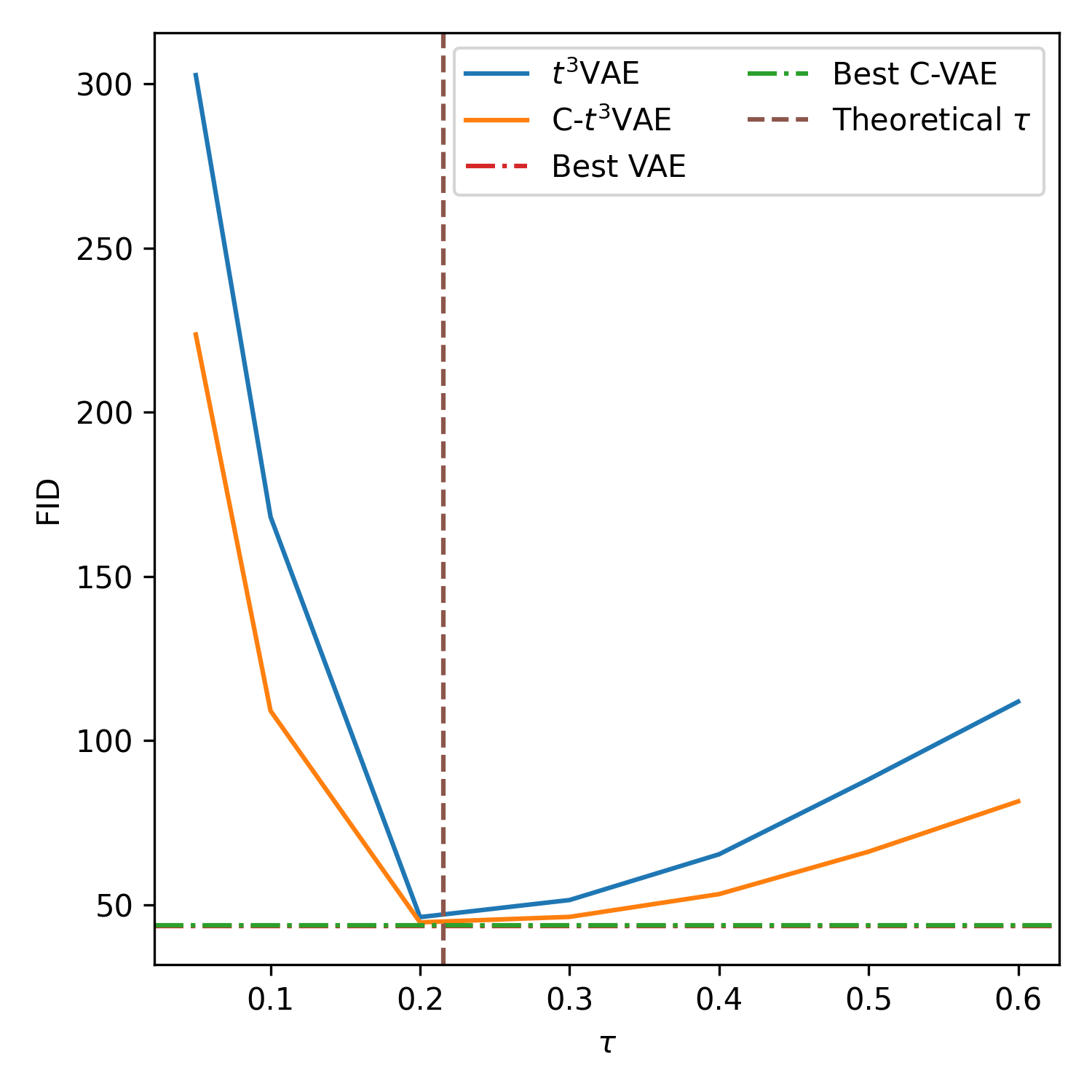}
			\caption{SVHN-LT $\rho$=1}
		\end{subfigure}
		
		\vspace{0.3cm} 
		
		\begin{subfigure}[b]{0.28\columnwidth}
			\centering
			\includegraphics[width=\linewidth]{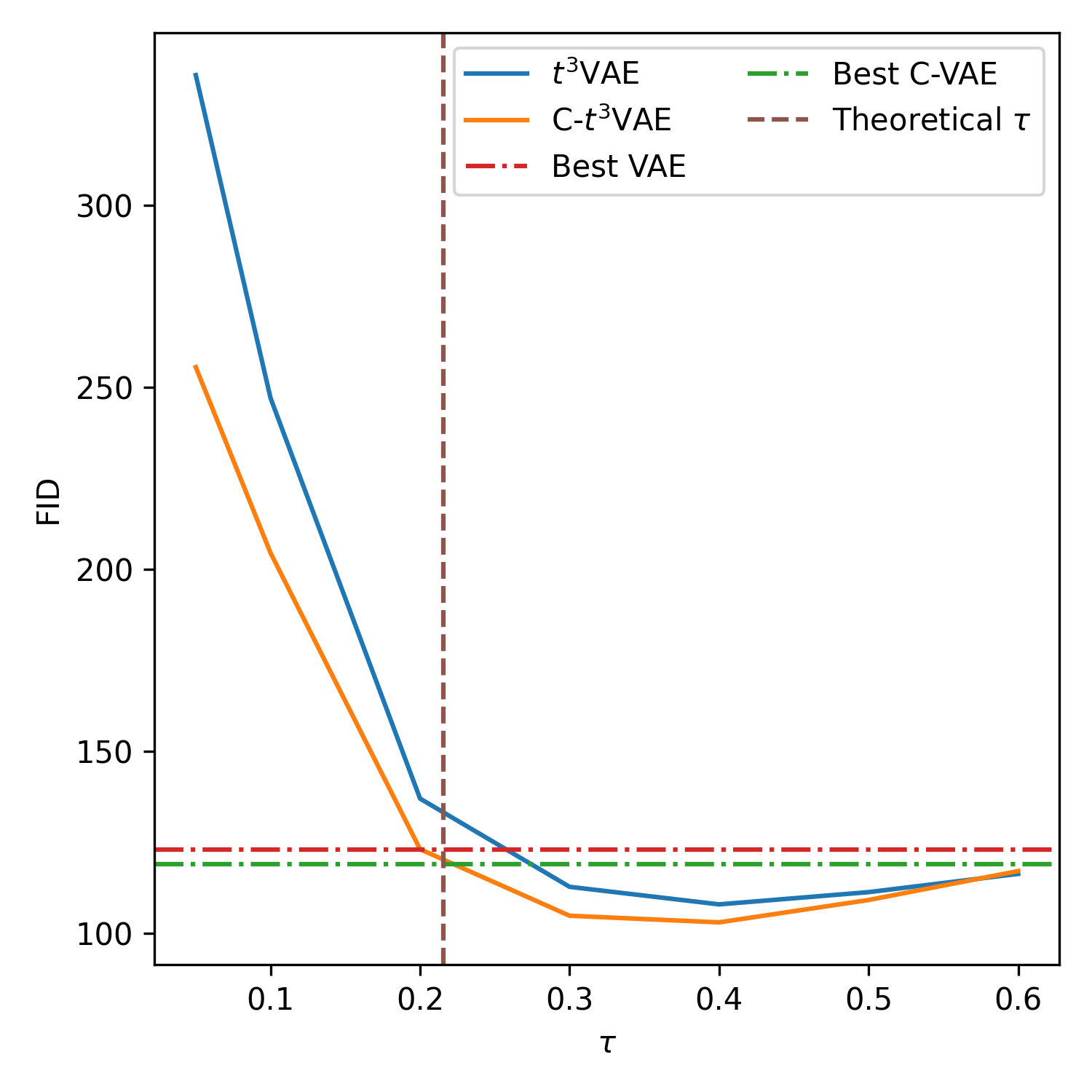}
			\caption{CIFAR100-LT $\rho$=50}
		\end{subfigure}
		\hfill
		\begin{subfigure}[b]{0.28\columnwidth}
			\centering
			\includegraphics[width=\linewidth]{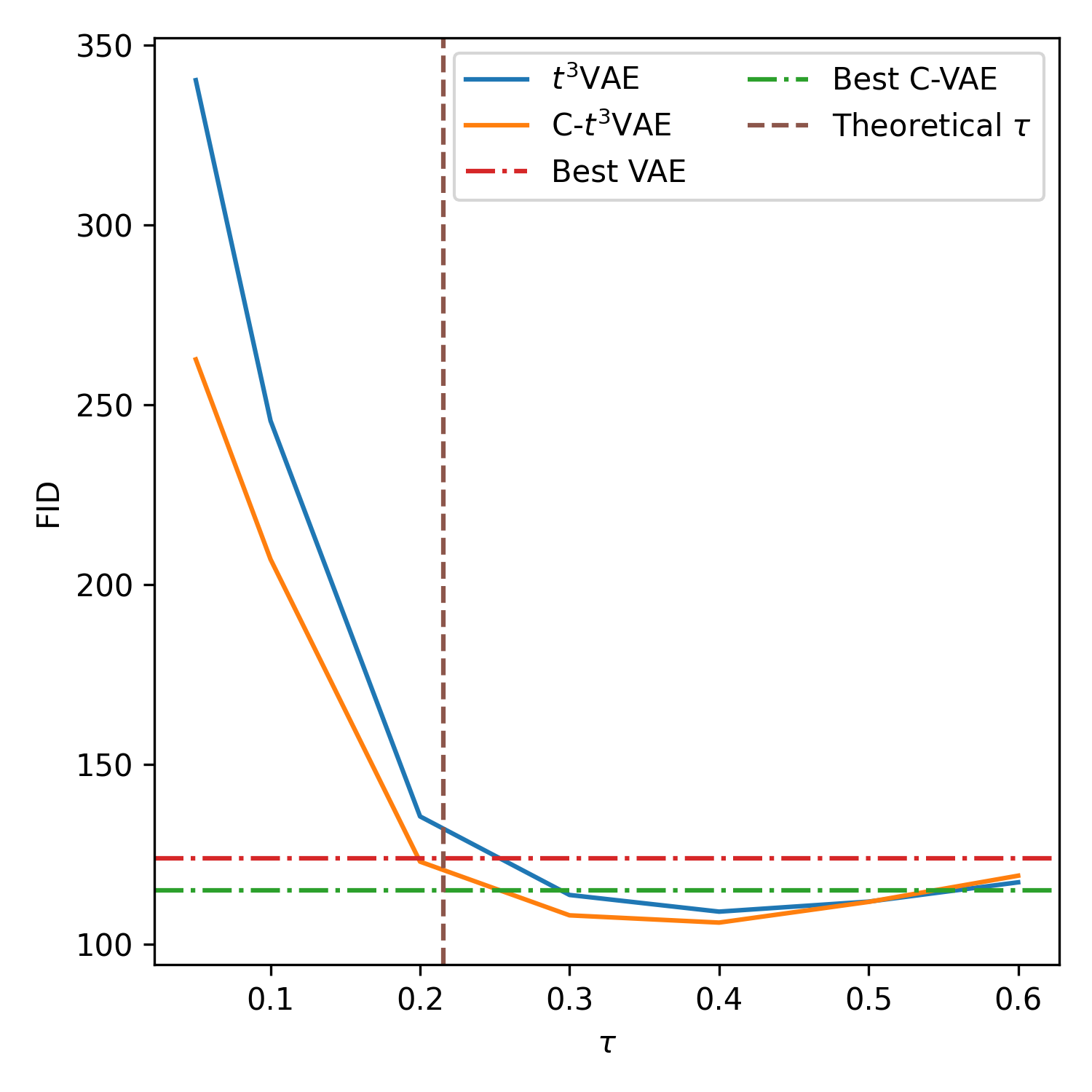}
			\caption{CIFAR100-LT $\rho$=10}
		\end{subfigure}
		\hfill
		\begin{subfigure}[b]{0.28\columnwidth}
			\centering
			\includegraphics[width=\linewidth]{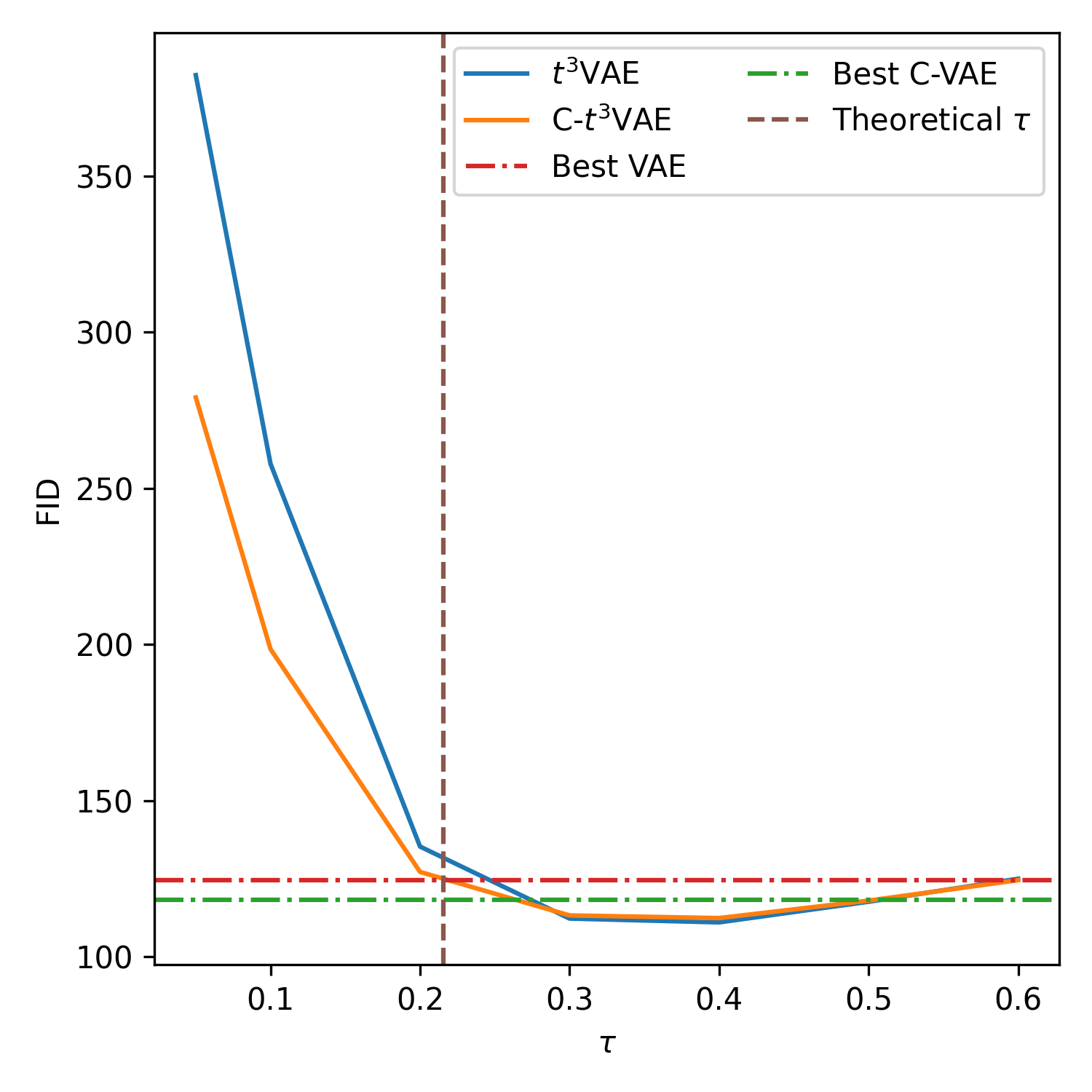}
			\caption{CIFAR100-LT $\rho$=1}
		\end{subfigure}
		
%
		\begin{subfigure}[b]{0.28\columnwidth}
			\centering
			\includegraphics[width=\linewidth]{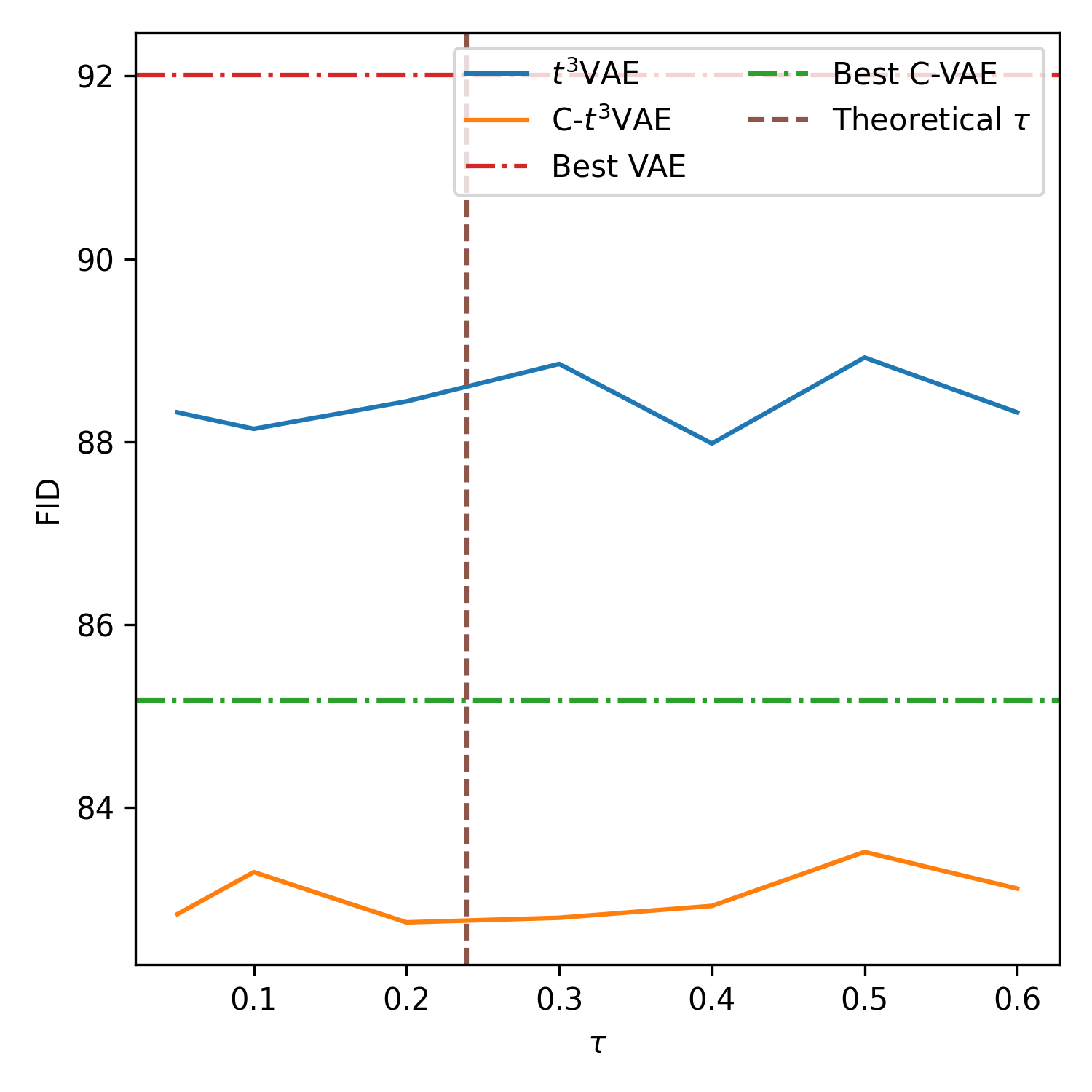}
			\caption{CelebA Young}
		\end{subfigure}
		\hfill
		\begin{subfigure}[b]{0.28\columnwidth}
			\centering
			\includegraphics[width=\linewidth]{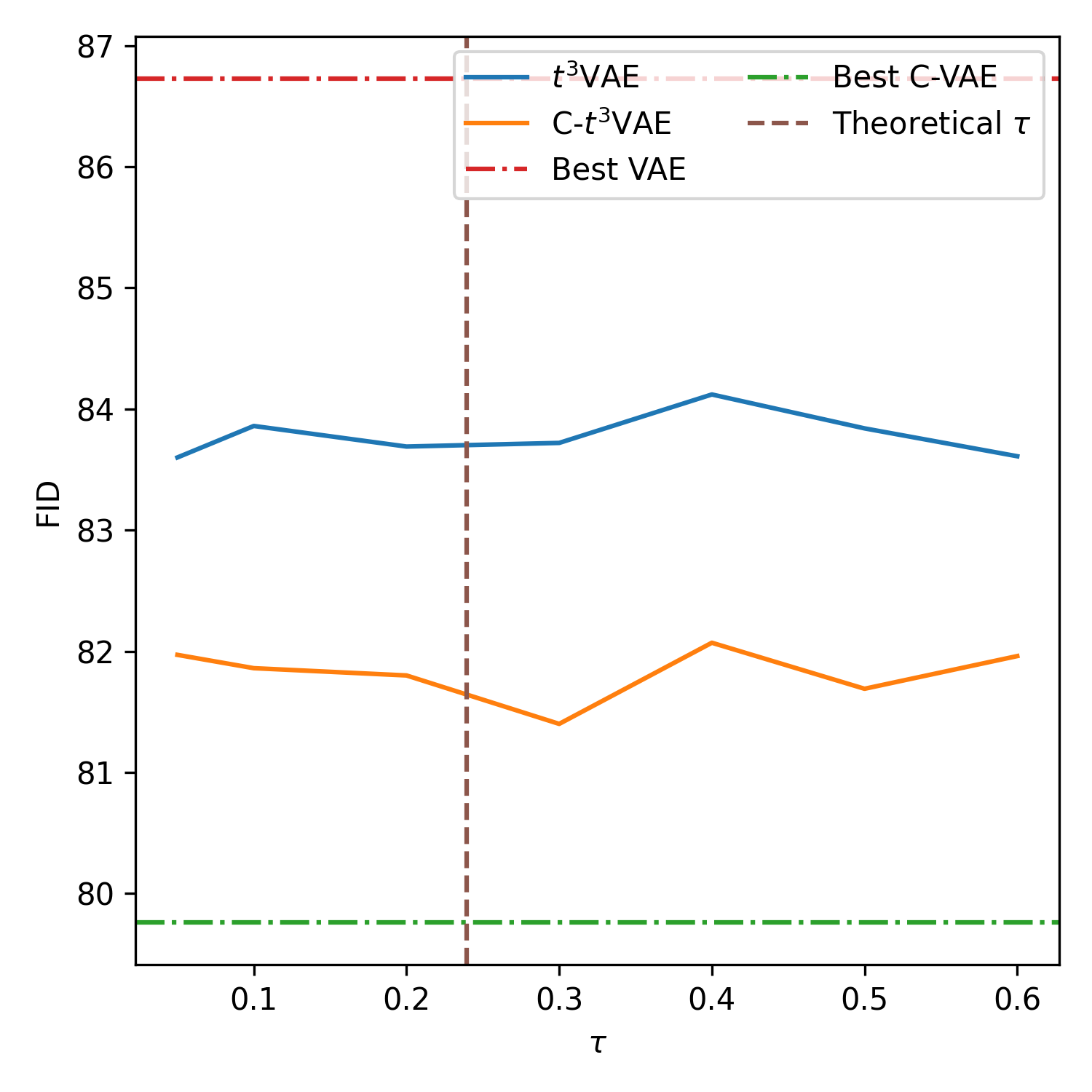}
			\caption{CelebA Male}
		\end{subfigure}
		\hfill
		\begin{subfigure}[b]{0.28\columnwidth}
			\centering
			\includegraphics[width=\linewidth]{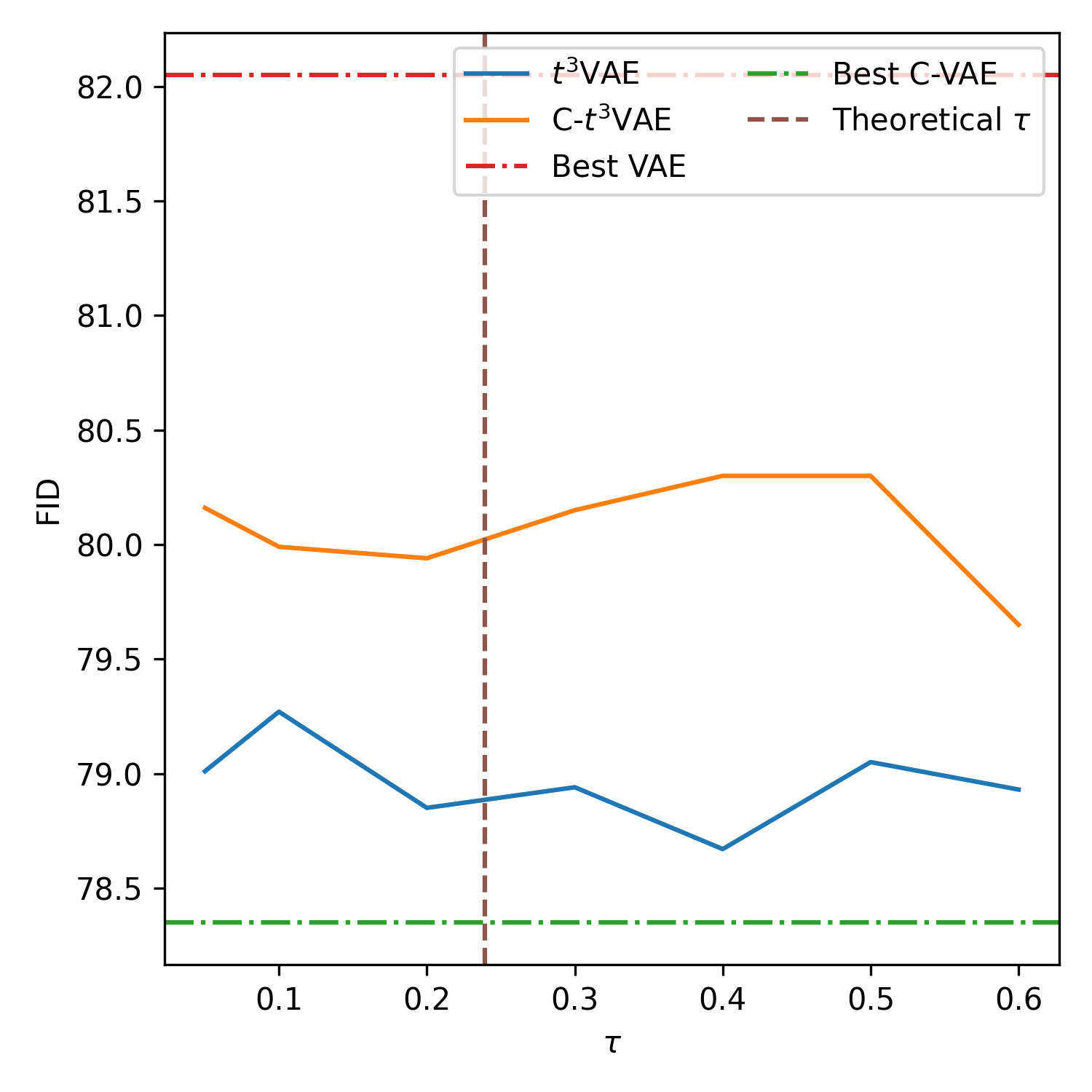}
			\caption{CelebA Smiling}
		\end{subfigure}	
		
		\caption{Variability of the FID as a function of the standard deviation $\tau^2$ for the $t^3$VAE and \mbox{C-$t^3$VAE}. In horizontal dashed lines is the FID value of the best performing VAE and C-VAE on each dataset. In vertical dashed lines is the theoretically identified value of $\tau$.}
		\label{tau study appendix}
	\end{figure}

	In this section, we evaluate the effect of the $\tau$ parameter on the SVHN-LT, CIFAR100-LT and CelebA datasets for all imbalance ratios while setting $\beta$ and $\nu$ to their previously optimized values.
	As shown in Figure~\ref{tau study appendix}, the optimal $\tau$ for SVHN-LT aligns closely with our theoretical prediction. In contrast, CIFAR100-LT consistently benefits from a larger $\tau = 0.4$, yielding improved FID across all imbalance settings and outperforming C-VAE. On CelebA, $\tau$ has minimal impact and the most likely value is $\tau \approx 0.3$. The analytically derived $\tau$ provides a principled initialization grounded in divergence geometry, while dataset-specific deviations reflect encoder--decoder capacity limits rather than theoretical inconsistency, highlighting the interaction between prior heaviness and representation complexity.
	
	\section{Per-Class Evaluation}
	\label{per class eval}
	In this section, we assess the conditional models' per-class Recall, Precision, and F1 metrics under all imbalance settings and for all tested datasets after optimization of all hyper-parameters.
	
	From the following figures in Table \ref{precision recall f1 shvn} and \ref{precision recall f1 cifar100}, we see that the C-$t^3$VAE consistently improves Recall and mode coverage in highly imbalanced settings with $\rho=100$ and $\rho=50$. This comes at a minor Precision cost but results in significantly better F1 scores across most classes. However, on balanced or mildly imbalanced datasets, its performance remains competitive with Gaussian-based models. This observation is valid for both the SVHN-LT and CIFAR100-LT but is more pronounced on the later.
	
	
	\begin{longtable}{M{.05\textwidth} | P{.25\textwidth} P{.25\textwidth} P{.25\textwidth}}
		\caption{Per-class generative metrics on SVHN-LT after optimization of $\beta$, $\nu$ and $\tau$ hyper-parameters.}
		\label{precision recall f1 shvn}
		\\
		\toprule
		$\rho$ & Recall & Precision & F1 score\\\midrule
		$100$ &
		\includegraphics[height=3.5cm]{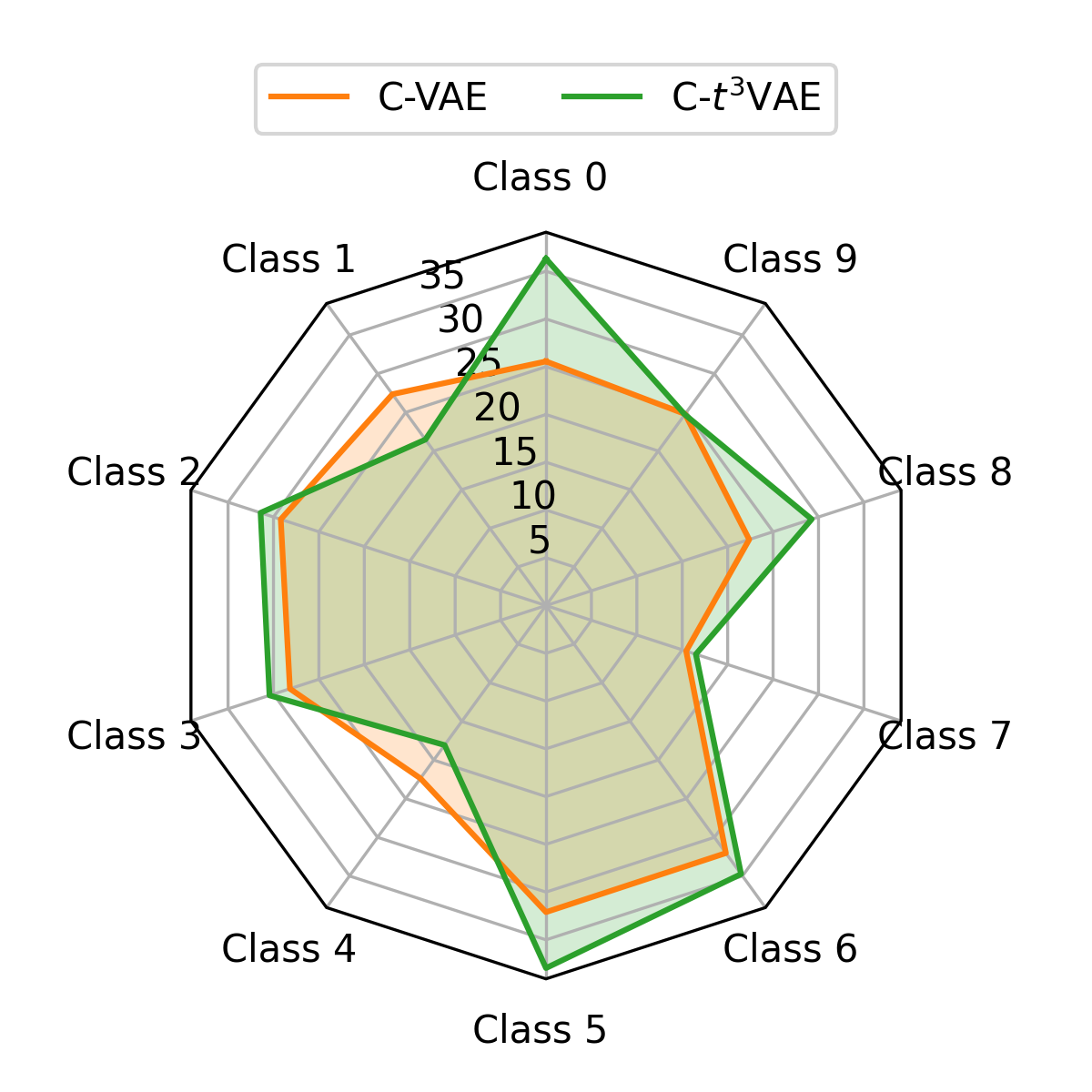} &
		\includegraphics[height=3.5cm]{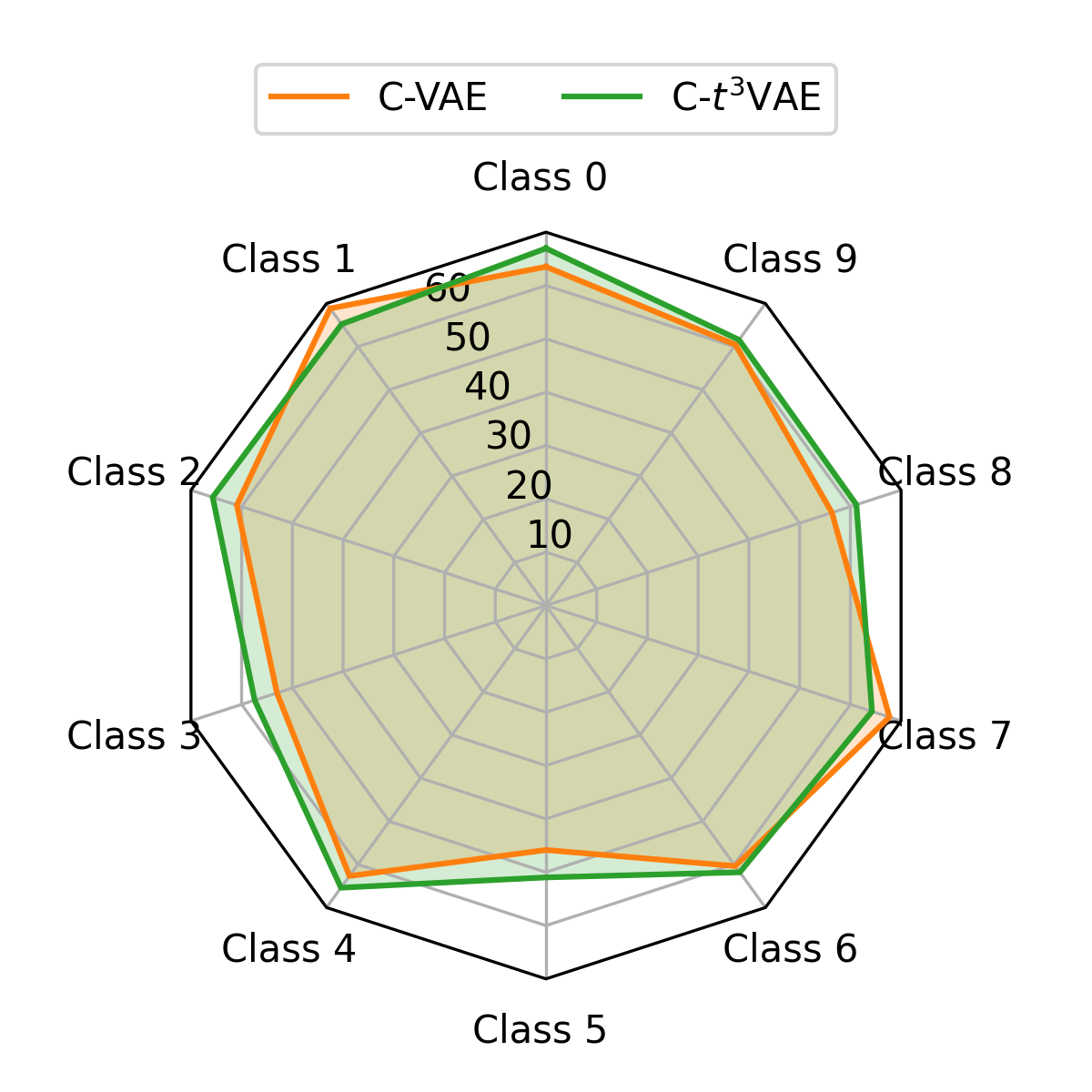}& 
		\includegraphics[height=3.5cm]{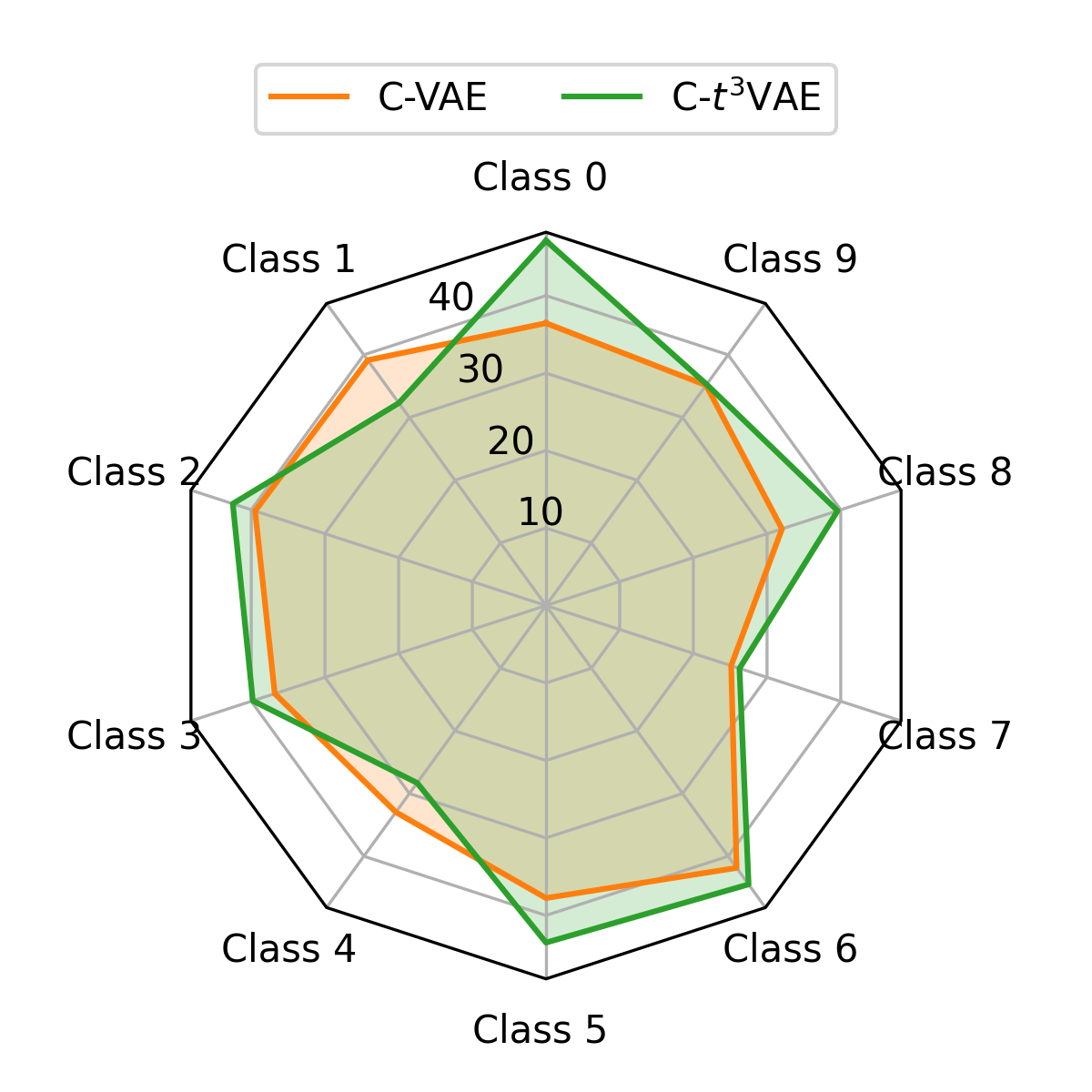}\\\midrule
		$50$ &
		\includegraphics[height=3.5cm]{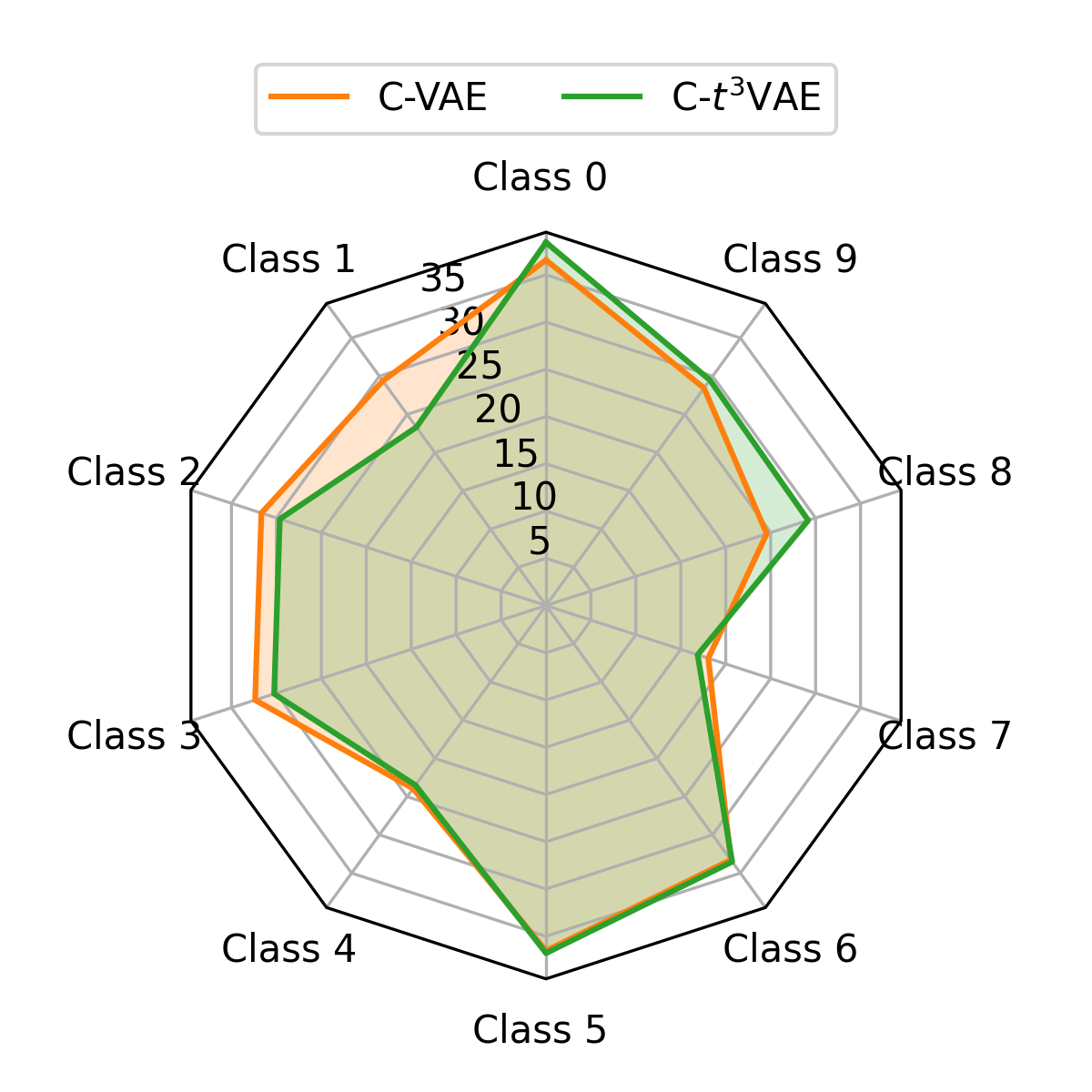} &
		\includegraphics[height=3.5cm]{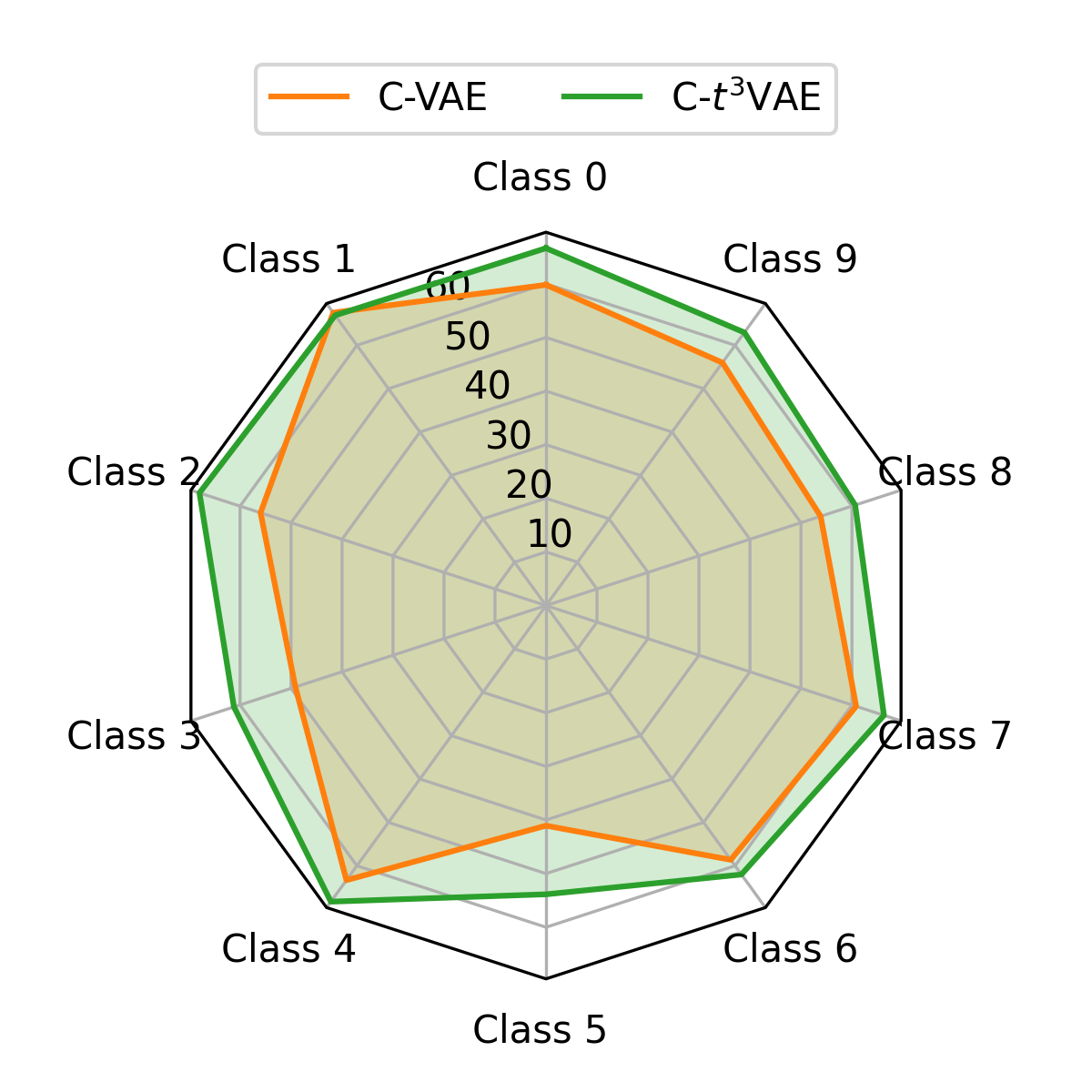}& 
		\includegraphics[height=3.5cm]{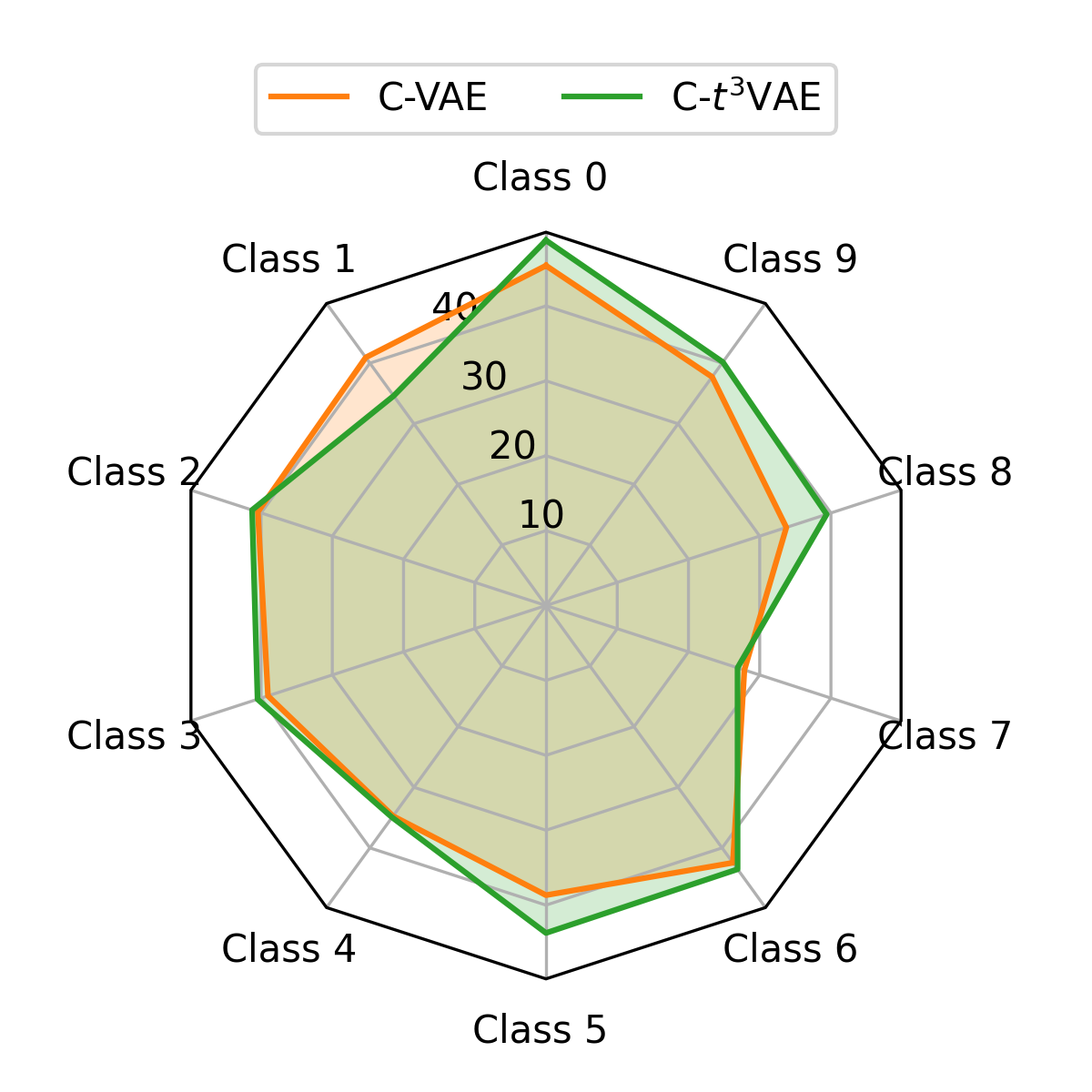}\\\midrule
		$10$ &
		\includegraphics[height=3.5cm]{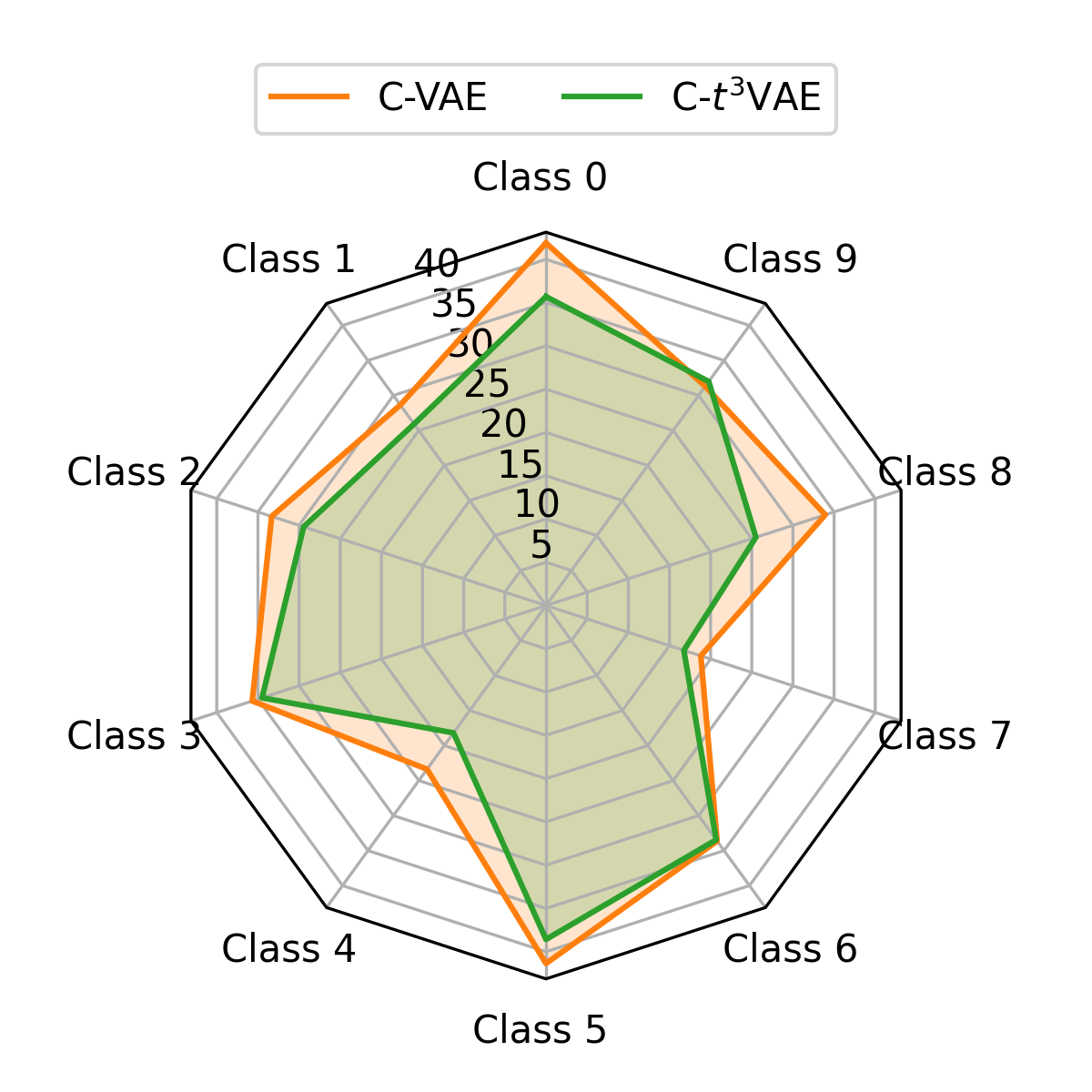} &
		\includegraphics[height=3.5cm]{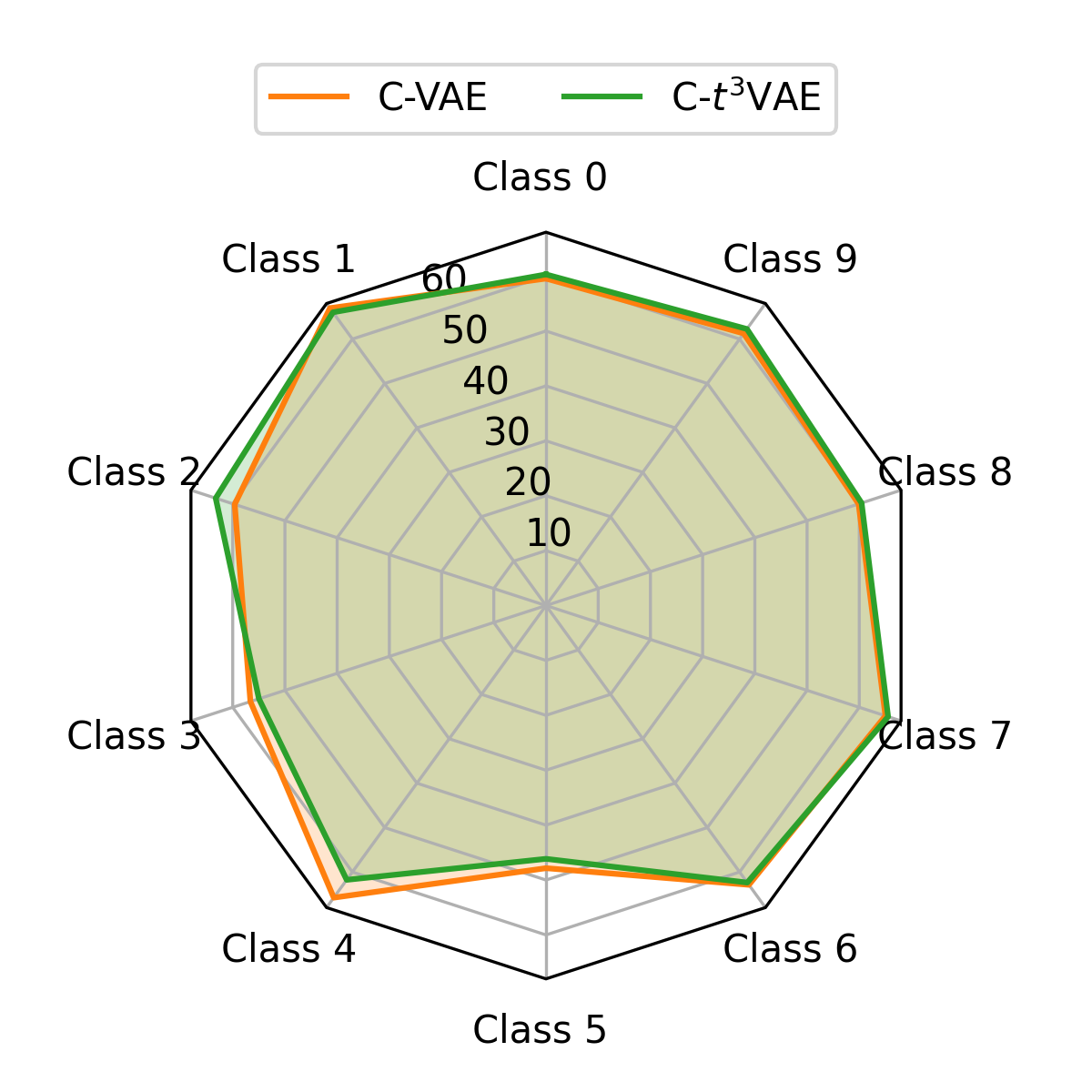}& 
		\includegraphics[height=3.5cm]{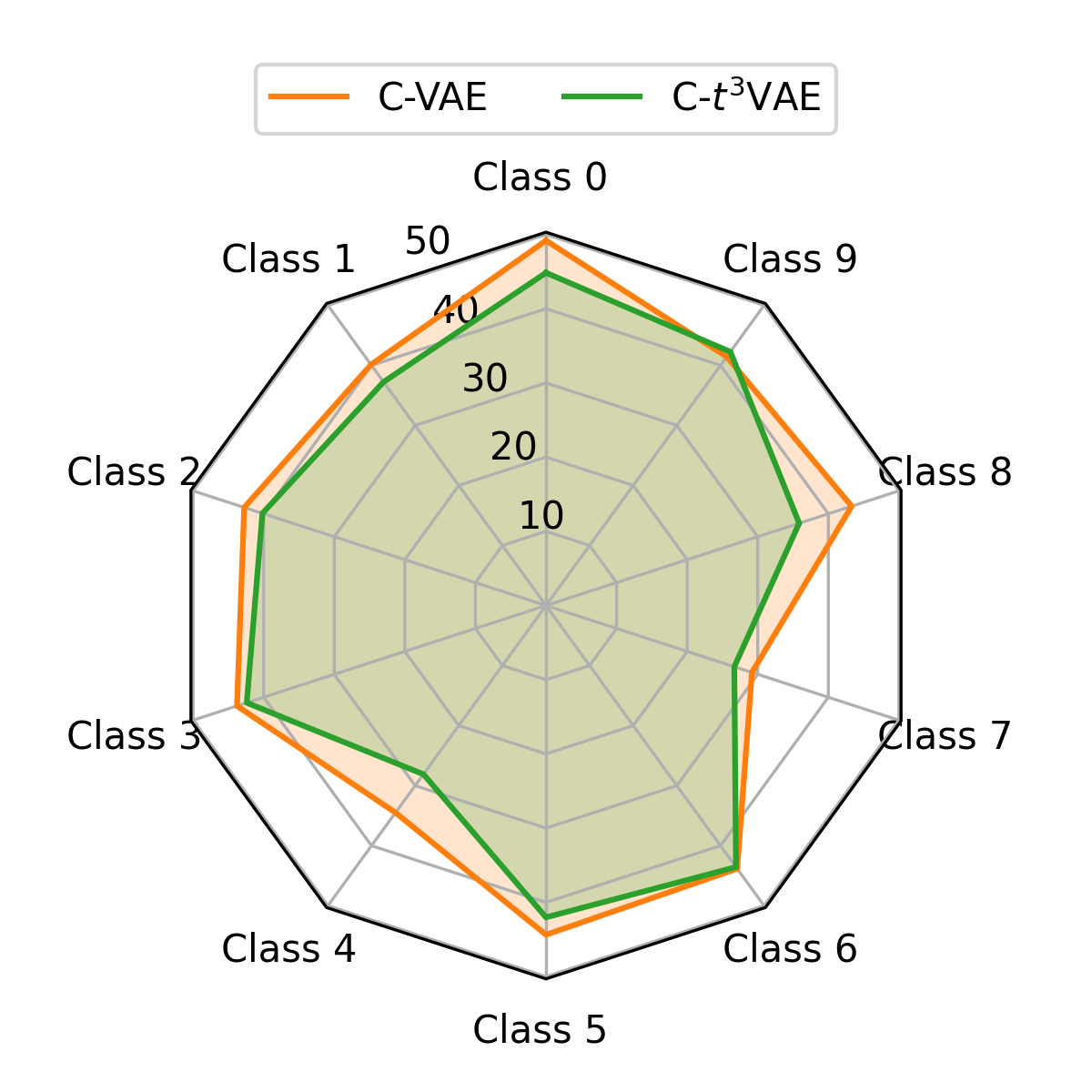}\\\midrule
		$1$ &
		\includegraphics[height=3.5cm]{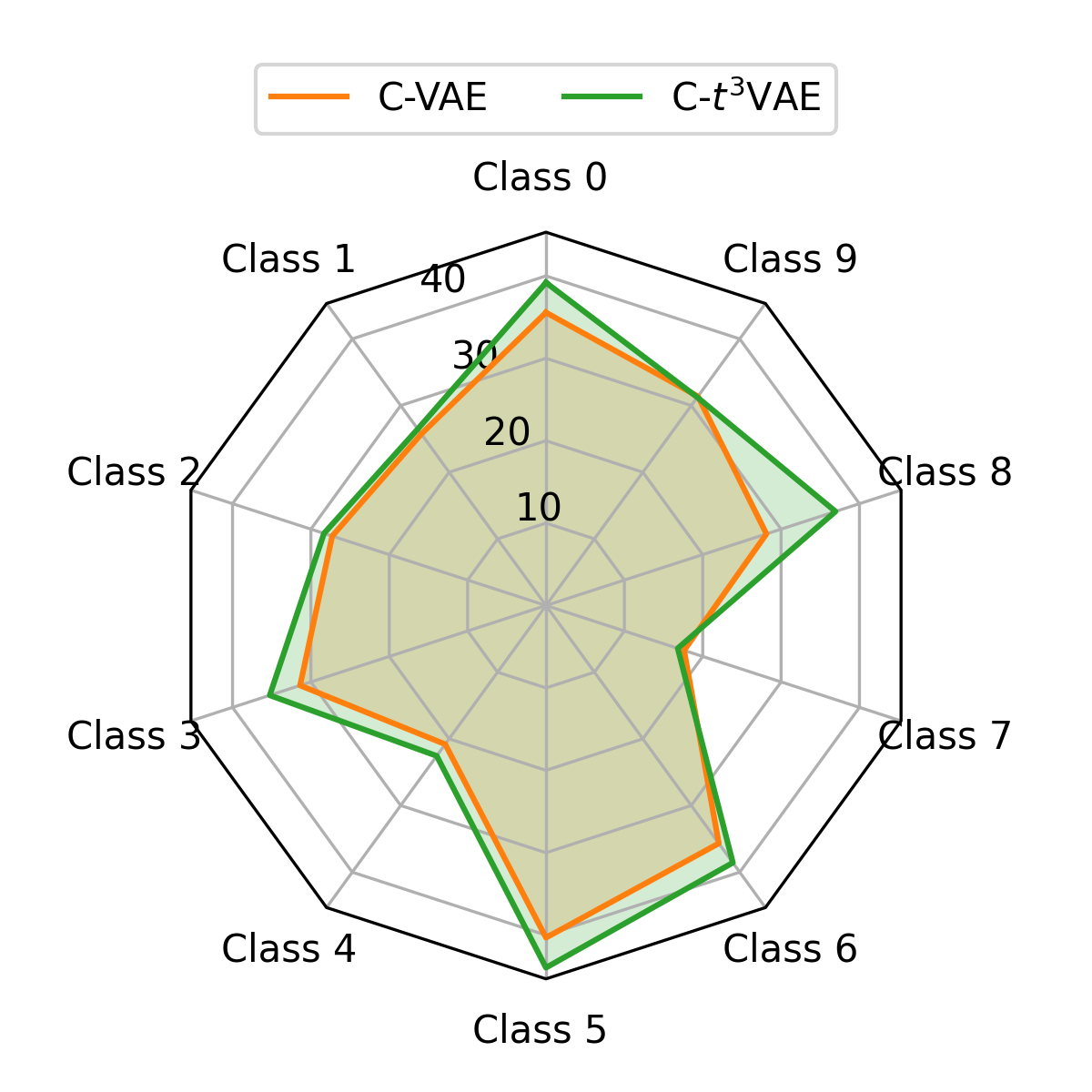} &
		\includegraphics[height=3.5cm]{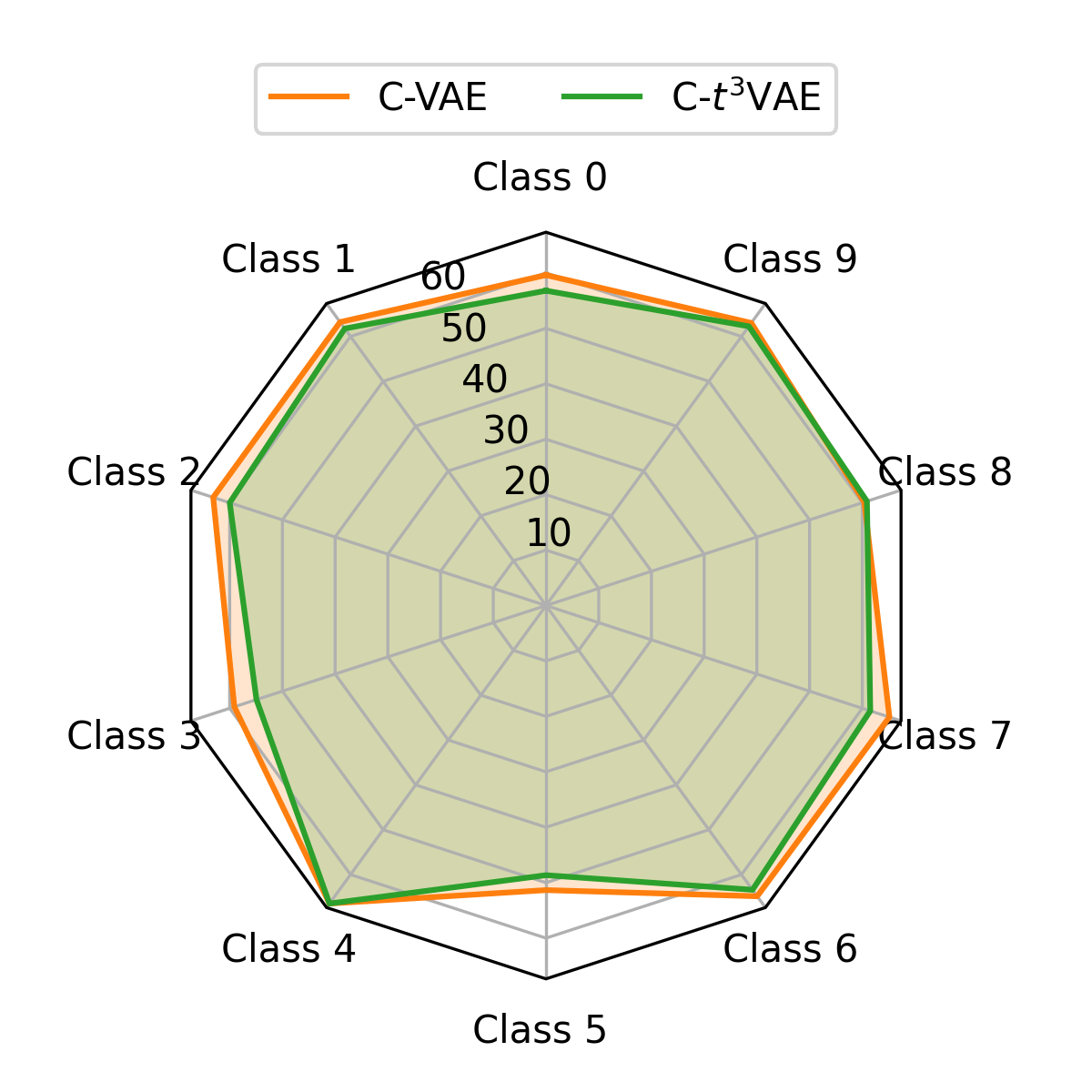}& 
		\includegraphics[height=3.5cm]{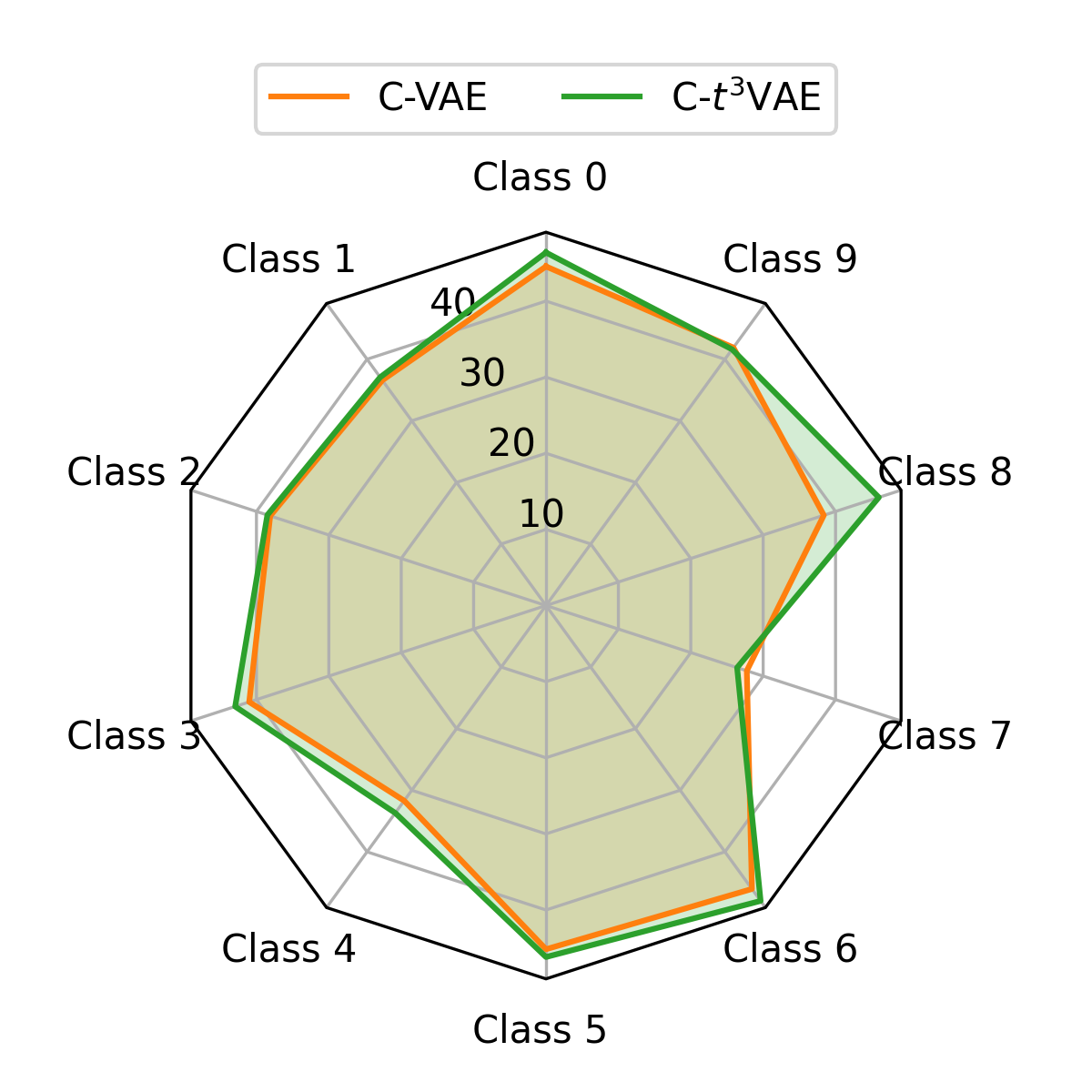}\\\bottomrule
	\end{longtable}
	
	\begin{longtable}{M{.05\textwidth} | P{.25\textwidth} P{.25\textwidth} P{.25\textwidth}}
		\caption{Per-class generative metrics on CIFAR100-LT after optimization of $\beta$, $\nu$ and $\tau$ hyper-parameters, we focus on the top 5 head and tail classes.}
		\label{precision recall f1 cifar100}
		\\
		\toprule
		$\rho$ & Recall & Precision & F1 score\\\midrule
		$100$ &
		\includegraphics[height=3.5cm]{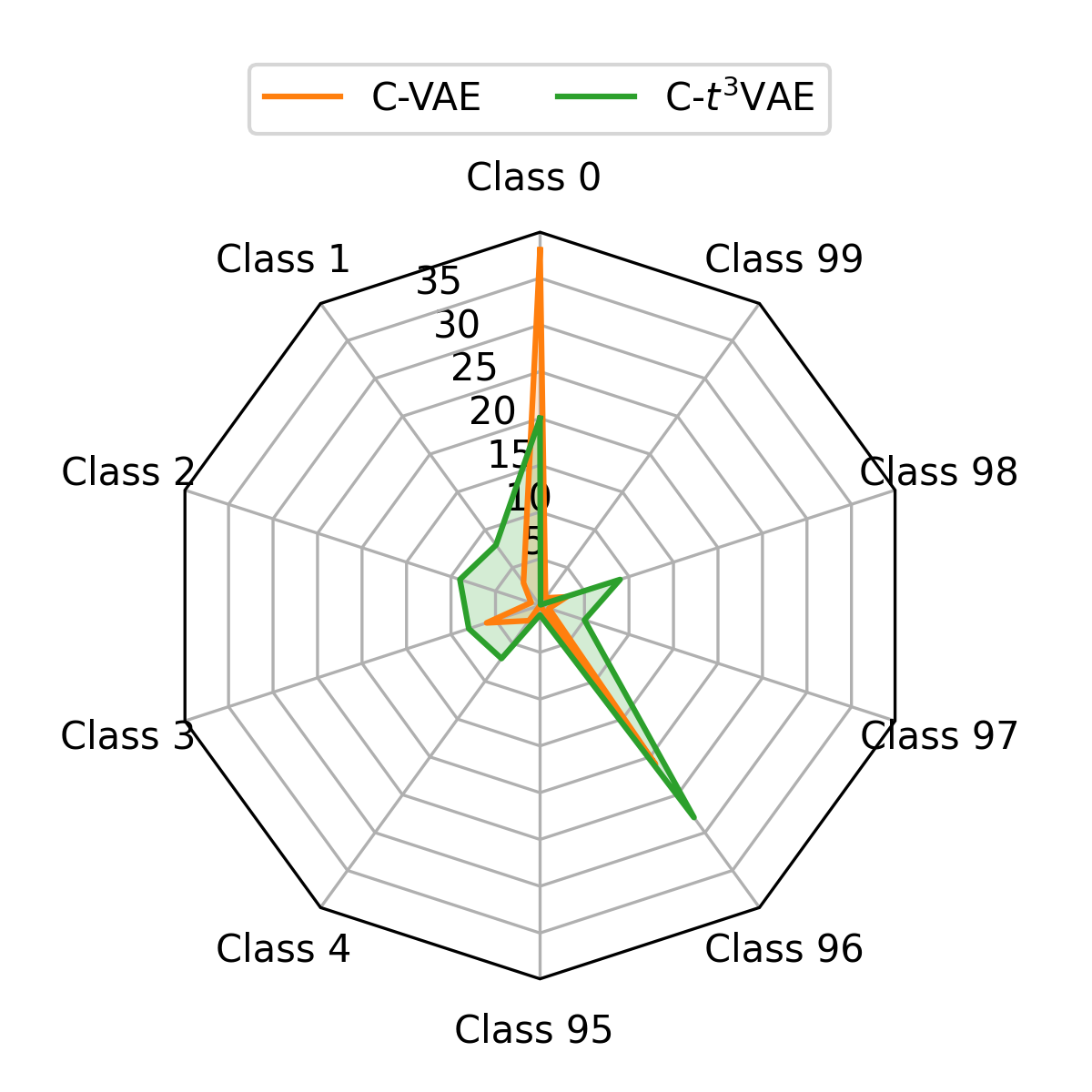} &
		\includegraphics[height=3.5cm]{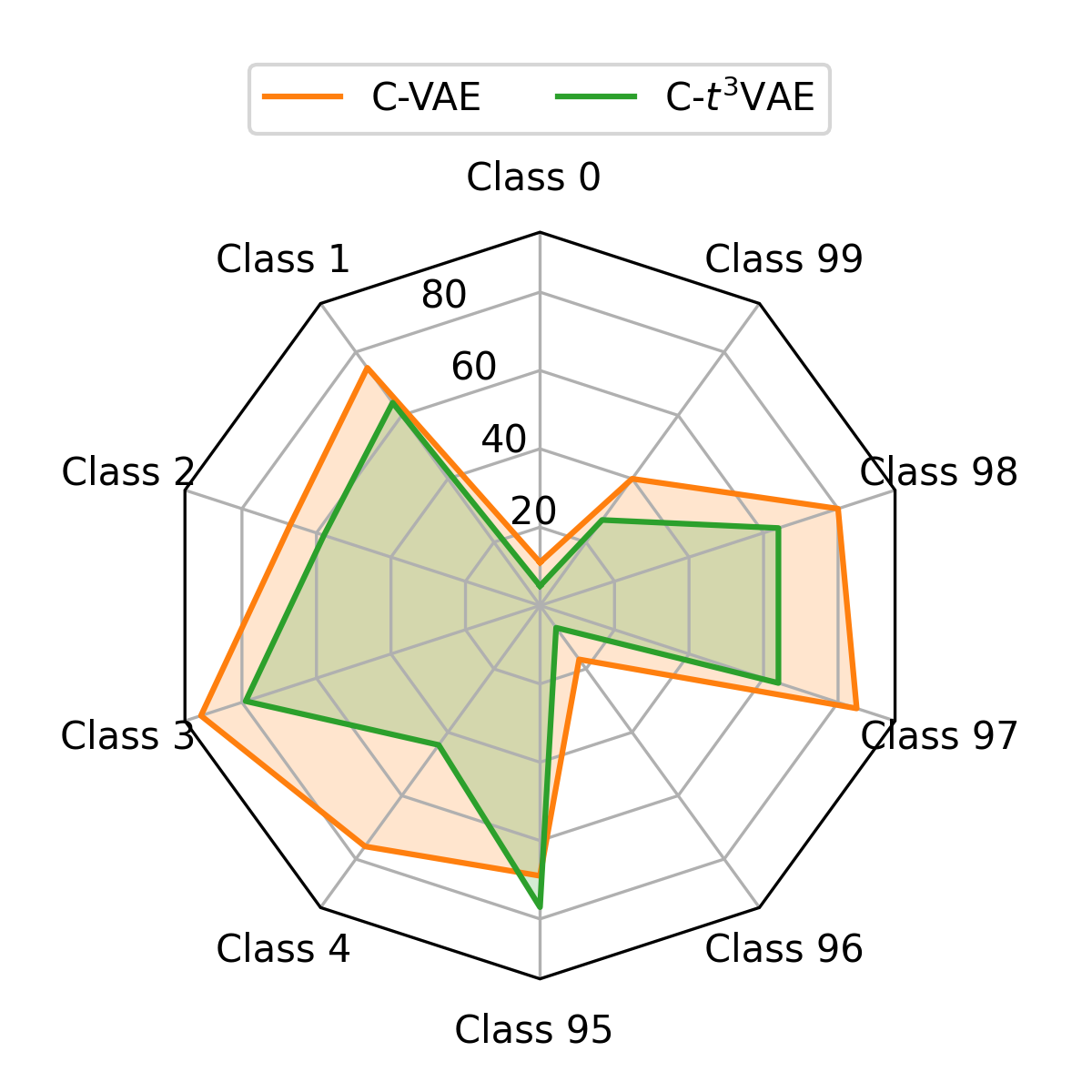}& 
		\includegraphics[height=3.5cm]{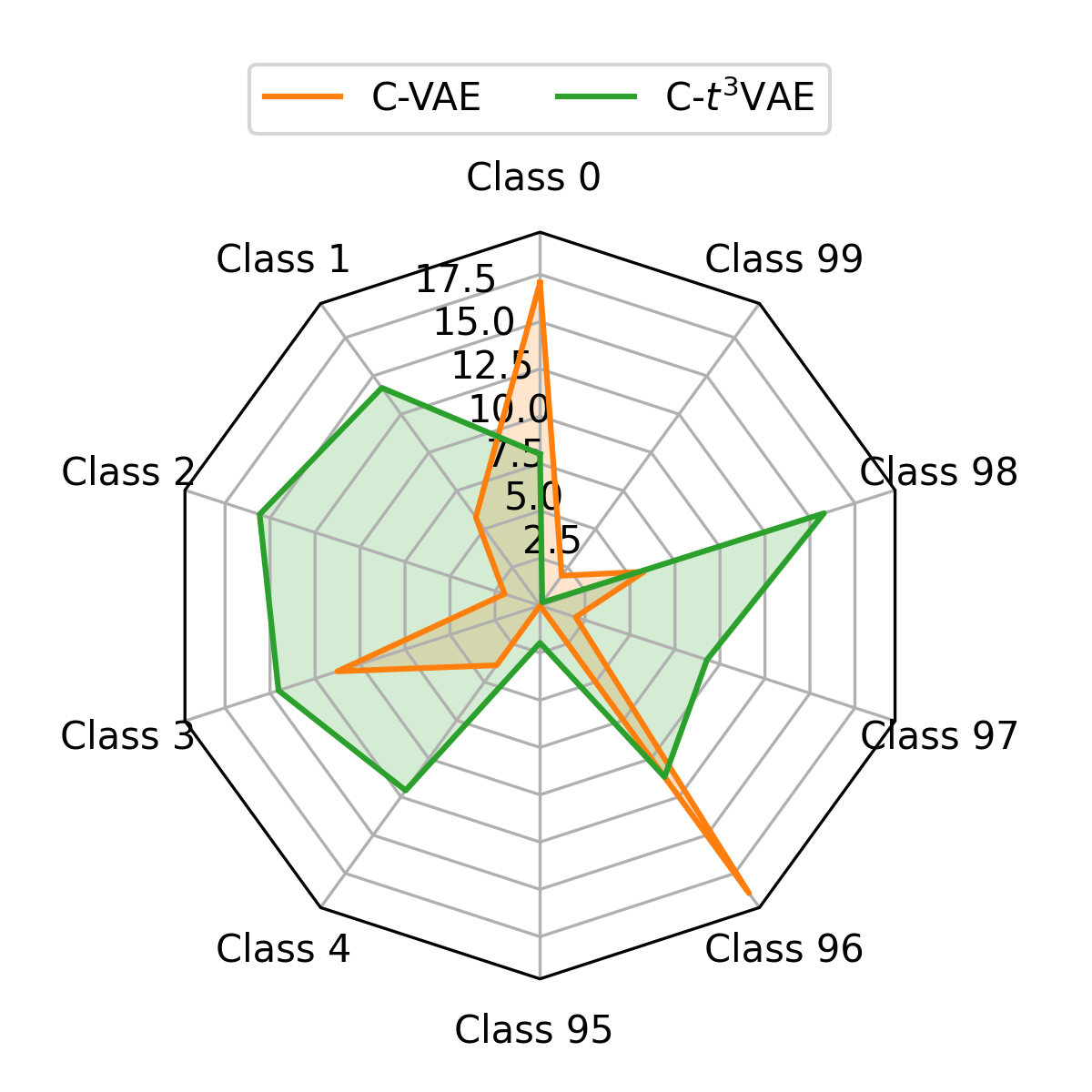}\\\midrule
		$50$ &
		\includegraphics[height=3.5cm]{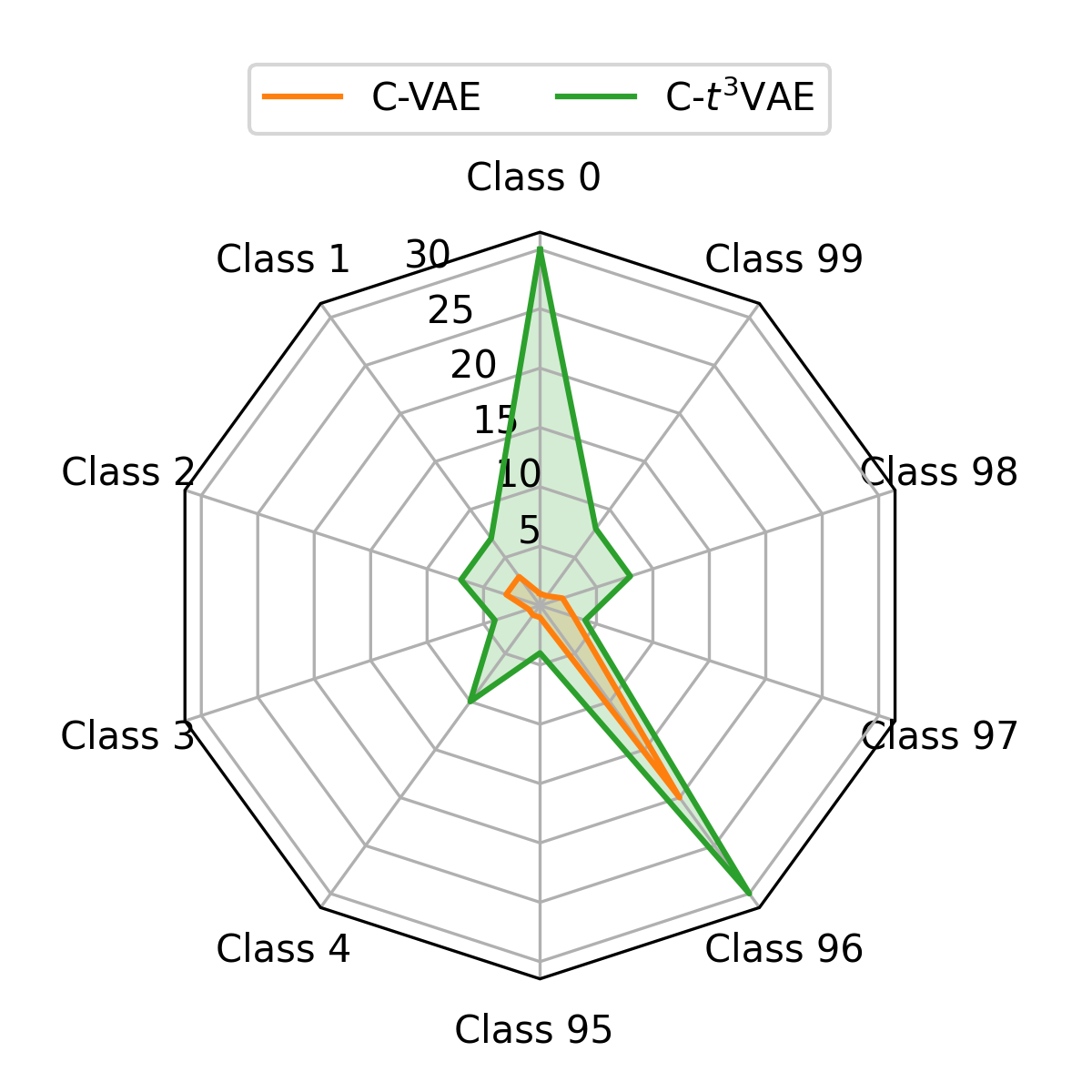} &
		\includegraphics[height=3.5cm]{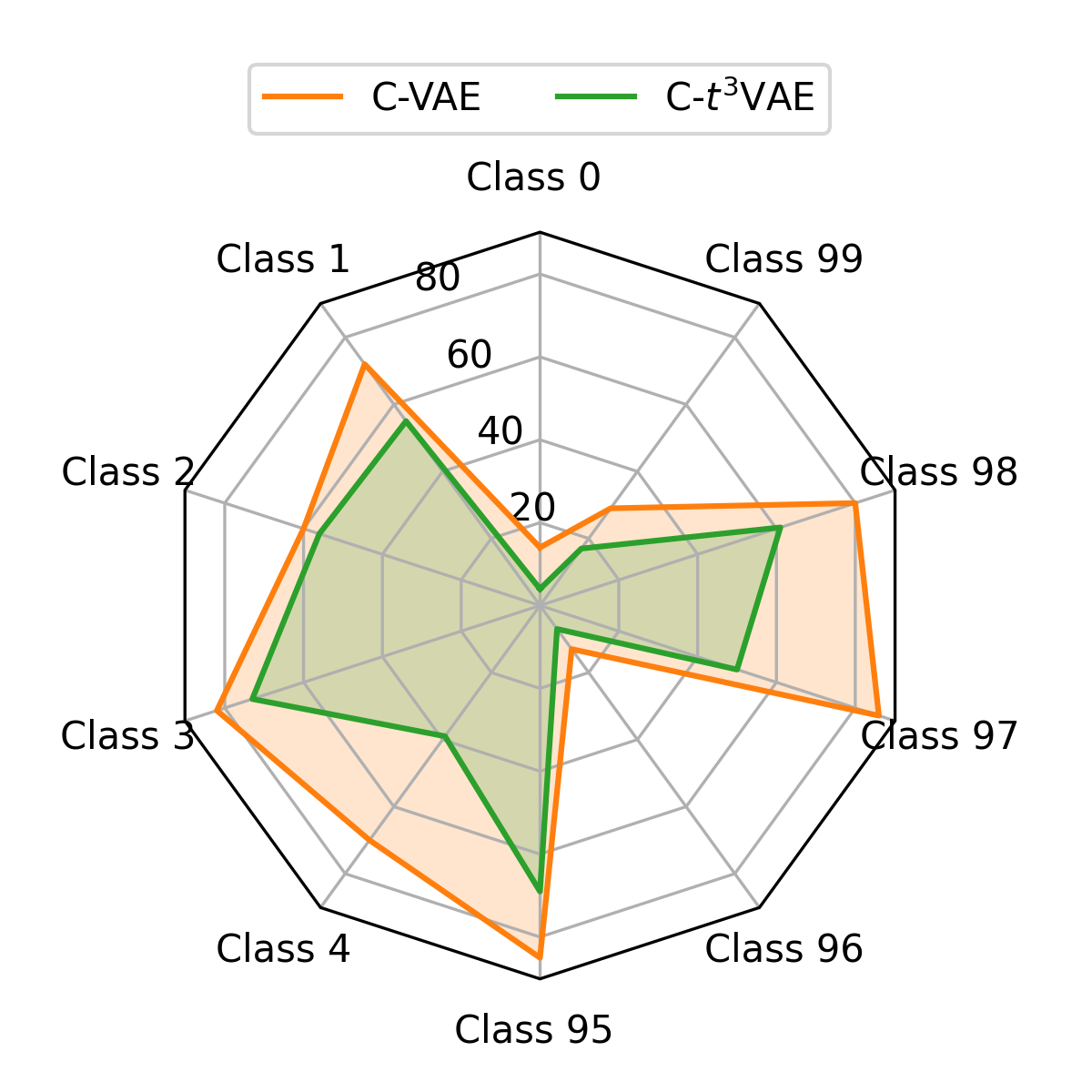}& 
		\includegraphics[height=3.5cm]{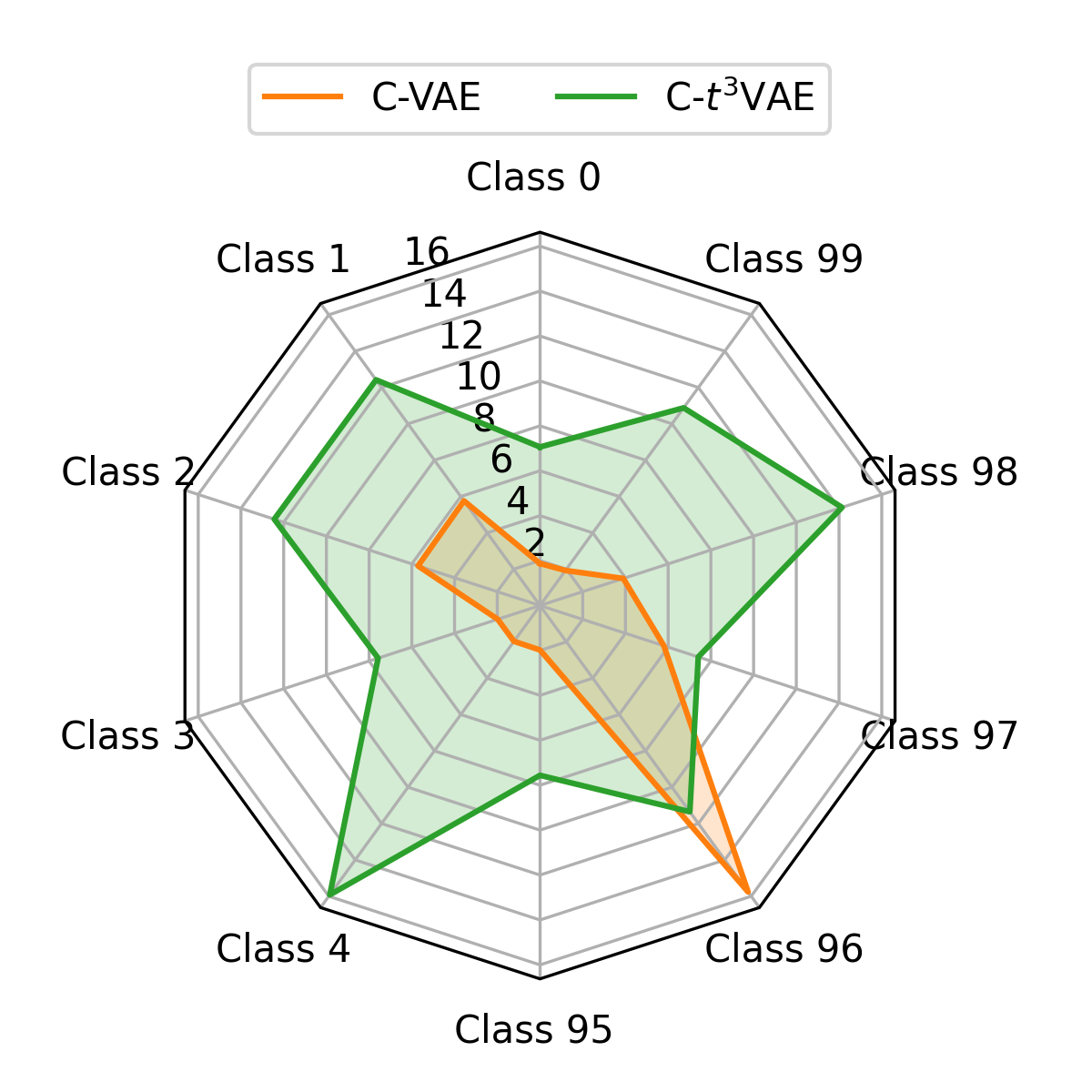}\\\midrule
		$10$ &
		\includegraphics[height=3.5cm]{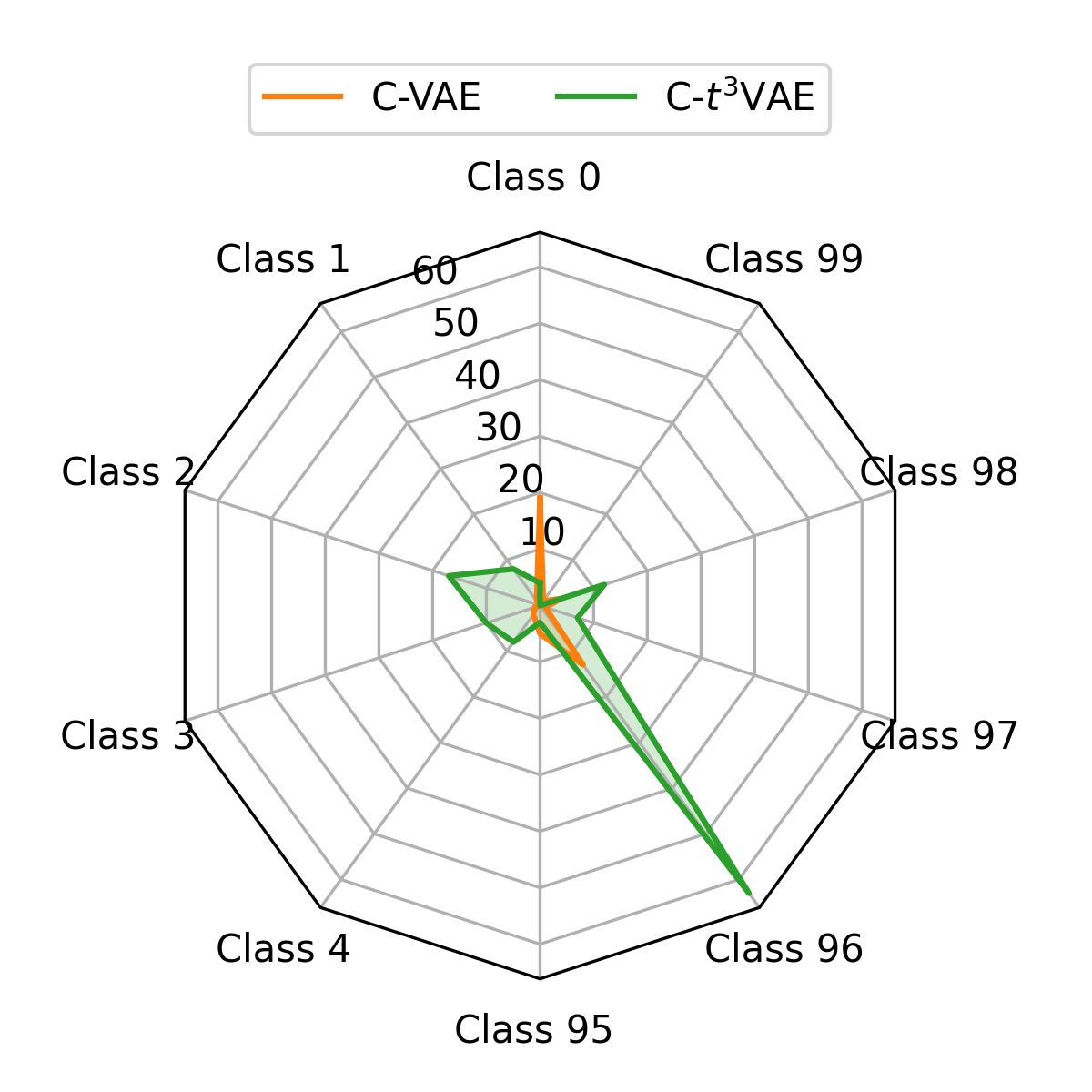} &
		\includegraphics[height=3.5cm]{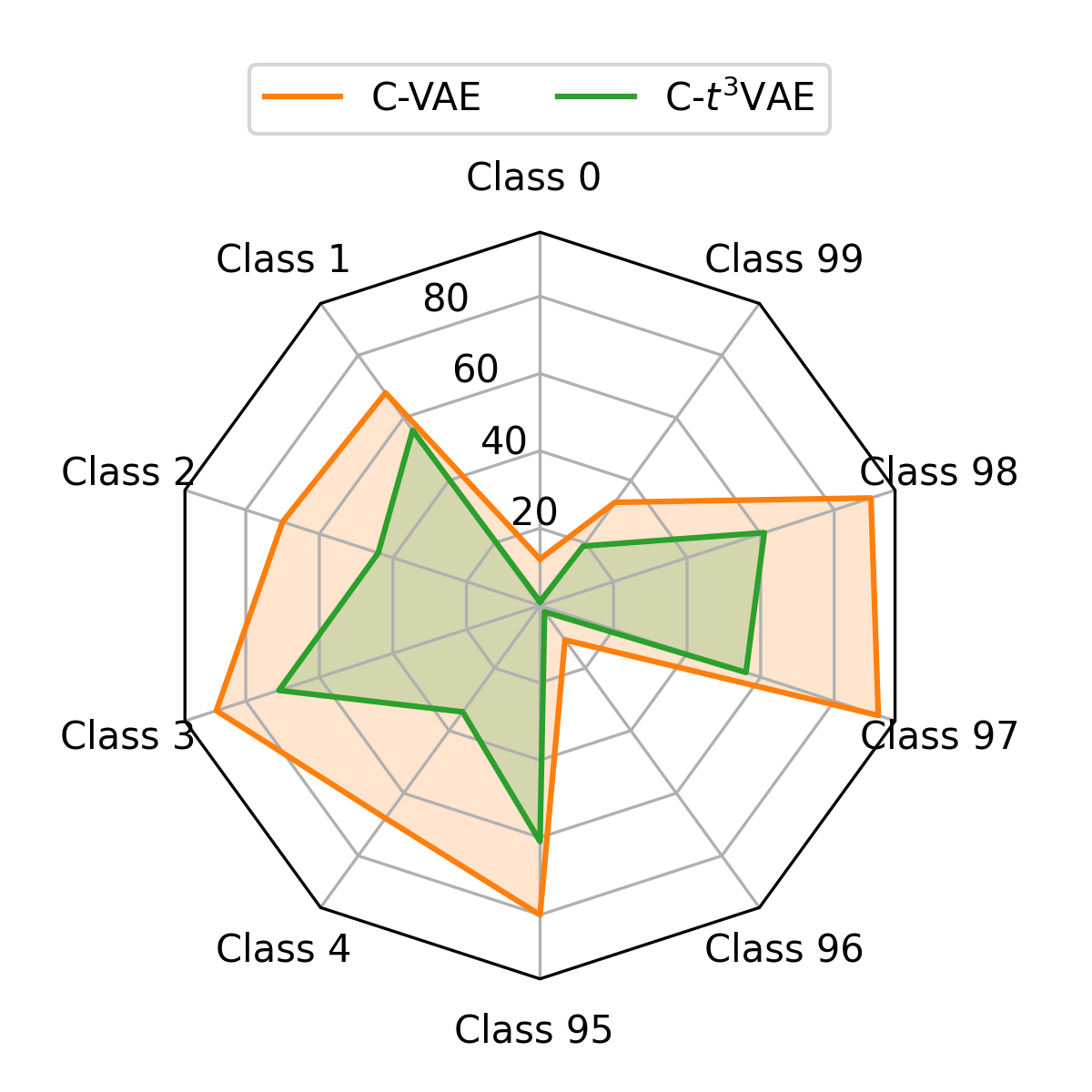}& 
		\includegraphics[height=3.5cm]{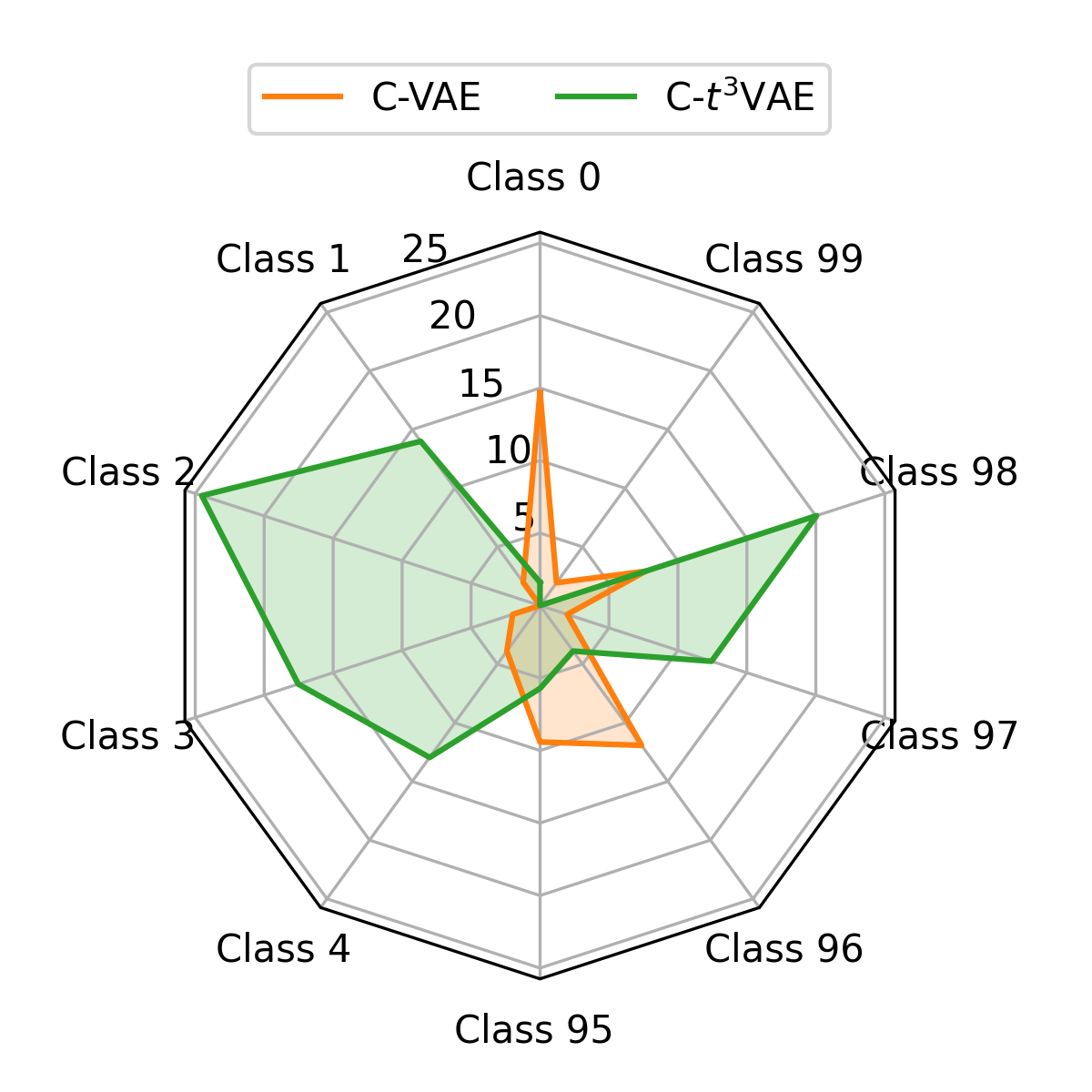}\\\midrule
		$1$ &
		\includegraphics[height=3.5cm]{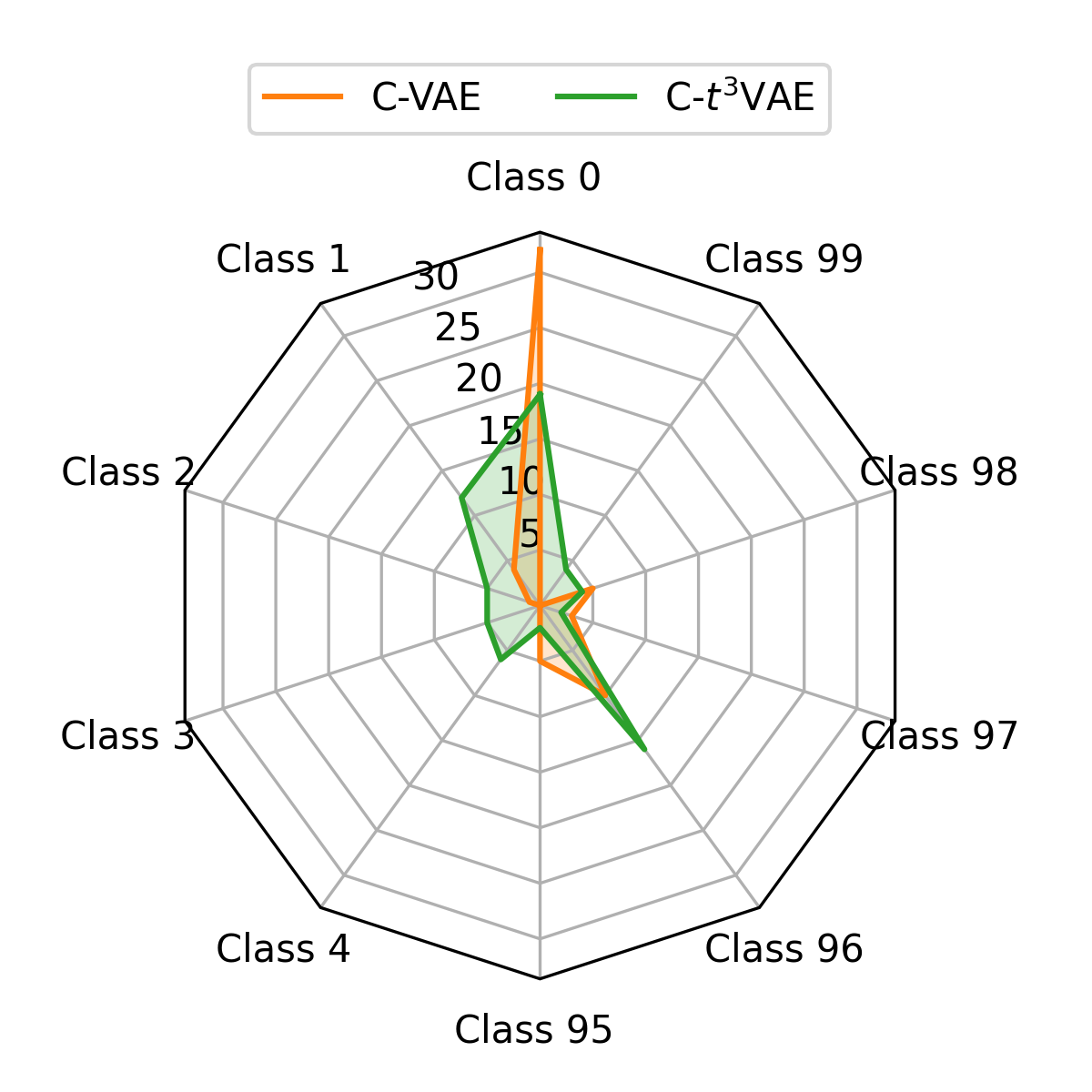} &
		\includegraphics[height=3.5cm]{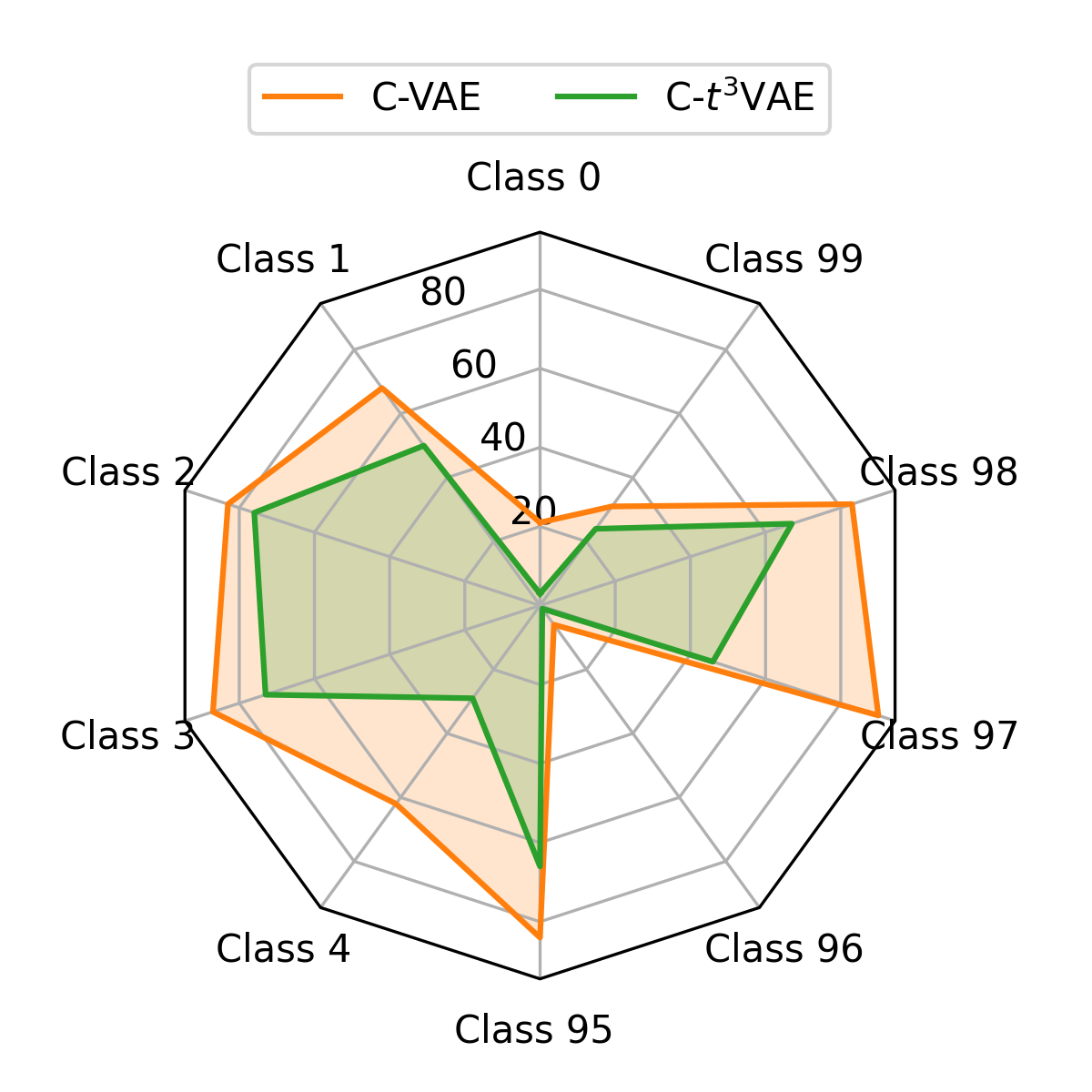}& 
		\includegraphics[height=3.5cm]{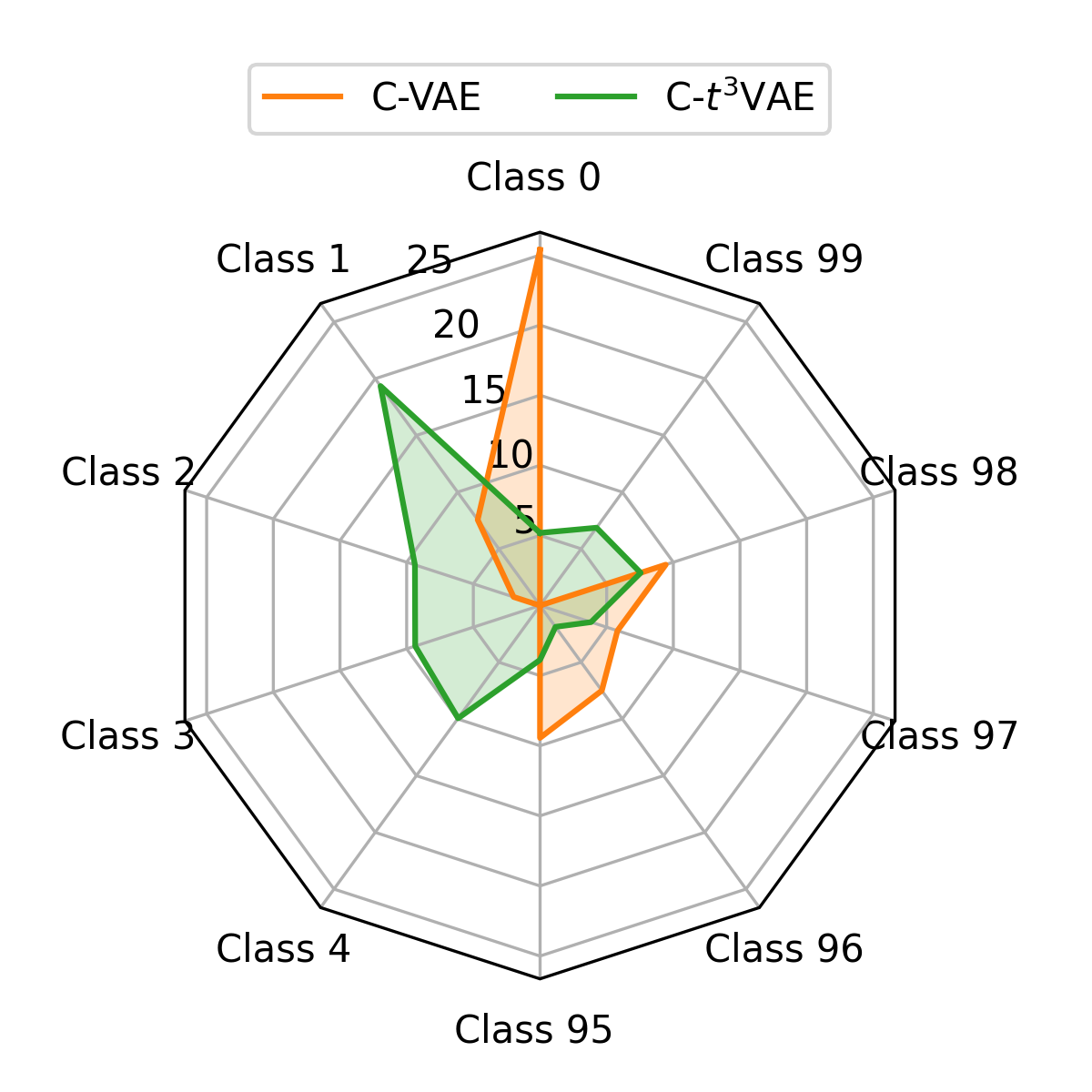}\\\bottomrule
	\end{longtable}

\end{document}